\documentclass[twoside]{article}
\usepackage{ecj,palatino,epsfig,latexsym,natbib}

\usepackage{amsmath,mathtools}
\usepackage{algorithm}
\usepackage[noend]{algpseudocode}
\usepackage{verbatim}
\usepackage[inline]{enumitem}
\usepackage{soul}
\usepackage{forest}

\usepackage[normalem]{ulem}

\usepackage{placeins} 

\usepackage{hyperref}
\usepackage{booktabs}
\usepackage{amssymb}
\usepackage{makecell}
\usepackage{multirow}
\usepackage{graphicx}
\usepackage{rotating}
\usepackage[detect-all]{siunitx}

\usepackage{colortbl}

\definecolor{Gray}{gray}{0.90}
\definecolor{LightCyan}{rgb}{0.88,1,1}

\definecolor{cbPurple}{RGB}{182,109,255}
\definecolor{cbTeal}{RGB}{0,146,146}
\definecolor{cbCyan}{RGB}{109,182,255}
\definecolor{cbBlue}{RGB}{0,109,219}
\definecolor{cbScarlet}{RGB}{146,0,0}
\definecolor{cbPink}{RGB}{255,109,182}
\definecolor{cbViolet}{RGB}{73,0,146}
\definecolor{cbOrange}{RGB}{219,209,0}
\definecolor{cbBrown}{RGB}{146,73,0}

\newcolumntype{a}{>{\columncolor{LightCyan}}c}
\newcolumntype{b}{>{\columncolor{white}}c}

\newsavebox{\boxedalignbox}
\newenvironment{boxedalign*}
  {\begin{equation*}\begin{lrbox}{\boxedalignbox}$\begin{aligned}}
  {\end{aligned}$\end{lrbox}\fbox{\usebox{\boxedalignbox}}\end{equation*}}

\usepackage{tikz}
\usetikzlibrary{shapes.geometric, arrows}
\tikzstyle{squarered} = [rectangle, minimum width=3cm, minimum height=1cm, text centered, draw=black, fill=red!10]
\tikzstyle{squaregreen} = [rectangle, minimum width=3cm, minimum height=1cm, text centered, draw=black, fill=green!10]
\tikzstyle{squarewhite} = [rectangle, minimum width=3cm, minimum height=1cm, text centered, draw=black, fill=white!10]
\tikzstyle{squareblue} = [rectangle, minimum width=3cm, minimum height=1cm, text centered, draw=black, fill=blue!10]
\tikzstyle{diamondgreen} = [diamond, minimum width=3cm, minimum height=1cm, text centered, draw=black, fill=green!10]
\tikzstyle{diamondblue} = [diamond, minimum width=3cm, minimum height=1cm, text centered, draw=black, fill=blue!10]
\tikzstyle{diamondred} = [diamond, minimum width=3cm, minimum height=1cm, text centered, draw=black, fill=red!10]
\tikzstyle{diamondorange} = [diamond, minimum width=3cm, minimum height=1cm, text centered, draw=black, fill=orange!10]
\tikzstyle{circlewhite} = [circle, minimum width=1cm, minimum height=1cm, text centered, draw=black]
\tikzstyle{circleyellow} = [circle, minimum width=1cm, minimum height=1cm, text centered, draw=black, fill=yellow!10]
\tikzstyle{squareryellow} = [rectangle, rounded corners, minimum width=3cm, minimum height=1cm, text centered, draw=black, fill=yellow!10]
\tikzstyle{invisible} = [rectangle, minimum width=0.5cm, minimum height=0.5cm, text centered]
\tikzstyle{arrow} = [thick,->,>=stealth]
\tikzstyle{darrow} = [dashed,->,>=stealth]
\tikzstyle{line} = [thick,-,>=stealth]

\newlist{inparaenum}{enumerate*}{1}
\setlist[inparaenum]{label=(\roman*)}

\def\CC{{C\nolinebreak[4]\hspace{-.05em}\raisebox{.4ex}{\tiny\bf ++}}}

\usepackage{xspace}

\DeclareRobustCommand{\btC}[1]{{\color{#1} $\blacktriangle$}\xspace}
\DeclareRobustCommand{\wtC}[1]{\raisebox{\depth}{\color{#1} \rotatebox{180}{$\blacktriangle$}}\xspace}

\DeclareRobustCommand{\ssW}{{\color{white} $\blacktriangle$}\xspace}

\usepackage{forest}
\usepackage{xcolor,colortbl}

\usepackage[english]{babel}

\algrenewcommand\alglinenumber[1]{{\sffamily\footnotesize#1}}

\begin{document}

\ecjHeader{x}{x}{xxx-xxx}{201X}{Improving Model-based GP for Symbolic Regression}{M. Virgolin, T. Alderliesten, C. Witteveen, P.A.N. Bosman}
\title{\bf Improving Model-based Genetic Programming for Symbolic Regression of Small Expressions}

\author{\name{\bf M.\ Virgolin} \hfill \addr{marco.virgolin@cwi.nl}\\ 
        \addr{Life Science and Health group, Centrum Wiskunde \& Informatica, Amsterdam, 1098 XG, the Netherlands.}
\AND
      \name{\bf T.\ Alderliesten} \hfill \addr{t.alderliesten@lumc.nl}\\ 
        \addr{Department of Radiation Oncology, Amsterdam UMC, University of Amsterdam, Amsterdam, 1105 AZ,  the Netherlands.}\\
        	\addr{Department of Radiation Oncology, Leiden University Medical Center, Leiden, 2333 ZA, the Netherlands.}
\AND
      \name{\bf C.\ Witteveen} \hfill \addr{c.witteveen@tudelft.nl}\\ 
        \addr{Algorithmics Group, Delft University of Technology, Delft, 2628 XE, the Netherlands.}
\AND
      \name{\bf P.\ A.\ N.\ Bosman} \hfill \addr{peter.bosman@cwi.nl}\\ 
        \addr{Life Science and Health group, Centrum Wiskunde \& Informatica, Amsterdam, 1098 XG, the Netherlands.}\\
        	\addr{Algorithmics Group, Delft University of Technology, Delft, 2628 XE, the Netherlands.}
}

\maketitle

\begin{abstract}

The Gene-pool Optimal Mixing Evolutionary Algorithm (GOMEA) is a model-based EA framework that has been shown to perform well in several domains, including Genetic Programming (GP). Differently from traditional EAs where variation acts blindly, GOMEA learns a model of interdependencies within the genotype, i.e., the linkage, to estimate what patterns to propagate. 
In this article, we study the role of Linkage Learning (LL) performed by GOMEA in Symbolic Regression (SR). We show that the non-uniformity in the distribution of the genotype in GP populations negatively biases LL, and propose a method to correct for this. We also propose approaches to improve LL when ephemeral random constants are used. Furthermore, we adapt a scheme of interleaving runs to alleviate the burden of tuning the population size, a crucial parameter for LL, to SR.
We run experiments on 10 real-world datasets, enforcing a strict limitation on solution size, to enable interpretability. We find that the new LL method outperforms the standard one, and that GOMEA outperforms both traditional and semantic GP. 
We also find that the small solutions evolved by GOMEA are competitive with tuned decision trees, making GOMEA a promising new approach to SR. 

\end{abstract}


\section{Introduction}

Symbolic Regression (SR) is the task of finding a function that explains hidden relationships in data, without prior knowledge on the form of such function. Genetic Programming (GP)~\citep{koza1992gp} is particularly suited for SR, as it can generate solutions of arbitrary form using basic functional components. 

Much work has been done in GP for SR, proposing novel algorithms~\citep{krawiec2016behavioral,zhong2018multifactorial,de2014kaizen}, hybrids~\citep{vzegklitz2017symbolic,icke2013improving}, and other forms of enhancement~\citep{keijzer2003improving,chen2015generalisation}. What is recently receiving a lot of attention is the use of so-called \emph{semantic-aware} operators, which enhance the variation process of GP by considering intermediate solution outputs~\citep{pawlak2015semantic,chen2018improving,moraglio2012geometric}.
The use of semantic-aware operators has proven to enable the discovery of very accurate solutions, but often at the cost of complexity: solution size can range from hundreds to billions of components~\citep{pawlak2015semantic,martins2018solving}. These solutions are consequently impossible to interpret, a fact that complicates or even prohibits the use of GP in many real-world applications because many practitioners desire to understand what a solution means before trusting its use~\citep{lipton2018mythos,guidotti2018survey}. The use of GP to discover uninterpretable solutions can even be considered to be questionable in many domains, as many alternative machine learning algorithms exist that can produce competitive solutions much faster~\citep{orzechowski2018where}.

We therefore focus on SR when GP is \emph{explicitly constrained} to generate small-sized solutions, i.e. mathematical expressions consisting of a small number of basic functional components, to increase the level of interpretability.
With size limitation, finding accurate solutions is particularly hard. It is not without reason that many effective algorithms work instead by growing solution size, e.g., by iteratively stacking components~\citep{moraglio2012geometric,chen2016xgboost}.

A recurring hypothesis in GP literature is that the evolutionary search can be made effective if \emph{salient patterns}, occurring in the representation of solutions (i.e., the genotype), are identified and preserved during variation~\citep{poli2008field}. It is worth studying if this holds for SR, to find accurate small solutions.

The hypothesis that salient patterns in the genotype can be found and exploited is what motivates the design of Model-Based Evolutionary Algorithms (MBEAs). Among them, the Gene-pool Optimal Mixing Evolutionary Algorithm (GOMEA) is recent EA that has proven to perform competitively in different domains: discrete optimization~\citep{thierens2011optimal,luong2014multi}, real-valued optimization~\citep{bouter2017exploiting}, but also grammatical evolution~\citep{medvet2018gomge}, and, the focus of this article, GP~\citep{virgolin2017scalable,virgolin2018symbolic}. GOMEA embodies within each generation a model-learning phase, where \emph{linkage}, i.e. the inter-dependency within parts of the genotype, is modeled. During variation, the linkage information is used to propagate genotype patterns and avoid their disruption.

The aim of this article is to understand the role of linkage learning when dealing with SR, and consequently improve the GP variant of GOMEA (GP-GOMEA), to find small and accurate SR solutions for realistic problems. We present three main contributions. First, we propose an improved linkage learning approach, that, differently from the original one, is unbiased with respect to the way the population is initialized. Second, we analyze how linkage learning is influenced by the presence of many different constant values, sampled by Ephemeral Random Constant (ERC) nodes~\citep{poli2008field}, and explore strategies to handle them. Third, we introduce improvements upon GP-GOMEA's Interleaved Multistart Scheme (IMS), a scheme of multiple evolutionary runs of increasing evolutionary budget that executes them in an interleaved fashion, to better deal with SR and learning tasks in general. 

The structure of this article is as follows. In Section~\ref{sec:related} we briefly discuss related work on MBEAs for GP. In Section~\ref{sec:gp-gomea}, we explain how GP-GOMEA and linkage learning work. 
Before proceeding with the description of the new contributions and experiments, Section~\ref{sec:expsettings} shows general parameter settings and datasets that will be used along the article. 
Next, we proceed by interleaving our findings on current limitations of GP-GOMEA followed by proposals to overcome such limitations, and respective experiments. In other words, we describe how we improve linkage learning one step at a time. 
In particular, Section~\ref{sec:improved-linkage-learning} presents current limitations of linkage learning, and describes how we improve linkage learning. Strategies to learn linkage efficiently and effectively when ERCs are used are described in Section~\ref{sec:linkage-erc}. 
We propose a new IMS for SR in Section~\ref{sec:ims}, and use it in Section~\ref{sec:benchmarking} to benchmark GP-GOMEA with competing algorithms: traditional GP, GP using a state-of-the-art semantic-aware operator, and the very popular decision tree for regression~\citep{breiman1984classification}. 
Lastly, we discuss our findings and draw conclusions in Section~\ref{sec:discussion}.

\section{Related work}\label{sec:related}
We differentiate today's MBEAs into two classes: Estimation-of-Distribution Algorithms (EDA), and Linkage-based Mixing EAs (LMEA).
EDAs work by iteratively updating a probabilistic model of good solutions, and sampling new solutions from that model. LMEAs attempt to capture linkage, i.e., inter-dependencies between parts of the genotype, and proceed by variating solutions with mechanisms to avoid the disruption of patterns with strong linkage.

Several EDAs for GP have been proposed so far. \citep{hauschild2011introduction} and \citep{kim2014probabilistic} are relatively recent surveys on the matter.
Two categories of EDAs for GP have mostly emerged in the years: one where the shape of solutions adheres to some template to be able to estimate probabilities of what functions and terminals appear in what locations (called \emph{prototype tree} for tree-based GP) \citep{salustowicz1997probabilistic,sastry2003probabilistic,yanai2003estimation,hemberg2012investigation}, and one where the probabilistic model is used to sample grammars of rules which, in turn, determine how solutions are generated~\citep{shan2004grammar,bosman2004learning,wong2014grammar,sotto2017probabilistic}.
Research on EDAs for GP appears to be limited. The review of \citep{kim2014probabilistic} admits, quoting, that
``\emph{Unfortunately, the latter research [EDAs for GP] has been sporadically carried out, and reported in several different research streams, limiting substantial communication and discussion}''. 

Concerning symbolic regression, we crucially found no works where it is attempted on realistic datasets (we searched among the work reported by the surveys and other recent work cited here). Many contributions on EDAs for GP have been validated on hard problems of artificial nature instead, such as \emph{Royal Tree} and \emph{Deceptive Max} \citep{hasegawa2009latent}.
Some real-world problems have been explored, but concerning only a limited number of variables \citep{tanev2007genetic,li2010genetic}. When considering symbolic regression, at most synthetic functions or small physical equations 
with only few ($\leq 5$) variables have been considered (e.g., by \citep{ratle2001avoiding,sotto2017probabilistic}). 

The study of LMEAs has emerged the first decade of the millennium in the field of binary optimization, where it remains mostly explored to date~\citep{chen2007survey,thierens2013hierarchical,goldman2014parameter,hsu2015optimization}.
Concerning GP, GOMEA is the first state-of-the-art LMEA ever brought to GP~\citep{virgolin2017scalable}. 

GP-GOMEA was first introduced in~\citep{virgolin2017scalable}, to tackle classic yet artificial benchmark problems of GP (including some of the ones mentioned before), where the optimum is known. The IMS, largely inspired on the work by~\citep{harik1999parameterless}, was also proposed, to relieve the user from the need of tuning the population size. Population sizing is particularly crucial for MBEAs in general: the population needs to be big enough for probability or linkage models to be reliable, yet small enough to allow efficient search~\citep{harik1999gambler}.

GP-GOMEA has also seen a first adaptation to SR, to find small and accurate solutions for a clinical problem where interpretability is important~\citep{virgolin2018symbolic}. There, GP-GOMEA was engineered for the particular problem, and no analysis of what linkage learning brings to SR was performed. Also, instead of using the IMS, a fixed population size was used. This is because the IMS was originally designed by~\citep{virgolin2017scalable} to enable benchmark problems to be solved to optimality. No concern on generalization of solutions to unseen test cases was incorporated.

As to combining LMEAs with grammatical evolution, \citep{medvet2018gomge} also employed GOMEA, to attempt to learn and exploit linkage when dealing with different types of pre-defined grammars. In that work, only one synthetic function was considered for symbolic regression, among other four benchmark problems.

There is a need of assessing whether MBEAs can bring an advantage to real-world symbolic regression problems. This work attempts to do this, by exploring possible limitations of GP-GOMEA and ways to overcome them, and validating experiments upon realistic datasets with dozens of features and thousands of observations.

\section{Gene-pool Optimal Mixing Evolutionary Algorithm for GP}\label{sec:gp-gomea}
Three main concepts are at the base of (GP-)GOMEA: solution representation (genotype), linkage learning, and linkage-based variation. 
These components are arranged in a common outline that encompasses all algorithms of the GOMEA family.

Algorithm~\ref{alg:gomea-outline} shows the outline of GOMEA. As most EAs, GOMEA starts by initializing a population $\mathcal{P}$, given the desired population size $n^\text{pop}$. The generational loop is then started and continues until a termination criterion is met, e.g., a limit on the number of generations or evaluations, or a maximum time. Lines 4 to 8 represent a generation. First, the linkage model is learned, which is called Family of Subsets (FOS) (explained in Sec.~\ref{sec:linkage-learning}). Second, each solution $\mathcal{P}_i$ is used to generate an offspring solution $\mathcal{O}_i$ by the variation operator Gene-pool Optimal Mixing (GOM). Last, the offspring replace the parent population. Note the lack of a separate selection operator. This is because GOM performs variation and selection at the same time (see Sec~\ref{sec:gom}).

For GP-GOMEA, an extra parameter is needed, the tree height (or, equivalently, tree depth) $h$. This is necessary to determine the representation of solutions, as described in the following Section~\ref{sec:gomea-representation}.

\begin{algorithm}
\caption{Outline of GOMEA}\label{alg:gomea-outline}
\begin{algorithmic}[1]
\Procedure{\texttt{runGOMEA}}{$n^\text{pop}$}
\State $\mathcal{P} \gets $\texttt{initializePopulation($n^\text{pop}$)}
\While {\texttt{terminationCriteriaNotMet()}}
	\State $F \gets $\texttt{learnFOS($\mathcal{P}$)}
	\State $\mathcal{O} \gets \emptyset$
	\For {$i \in \{1, \dots, n^\text{pop}\}$}
		\State $\mathcal{O}_i \gets $\texttt{GOM($\mathcal{P}_i,\mathcal{P},F$)}
	\EndFor
	\State $\mathcal{P} \gets \mathcal{O}$
\EndWhile
\EndProcedure
\end{algorithmic}
\end{algorithm}

\subsection{Solution representation in GP-GOMEA}\label{sec:gomea-representation}
GP-GOMEA uses a modification of the tree-based representation~\citep{koza1992gp} which is similar to the one used by~\citep{salustowicz1997probabilistic}. While typical GP trees can have any shape, GP-GOMEA uses a fixed template, that allows linkage learning and linkage-based variation to be performed in a similar fashion as for other, fixed string-length versions of GOMEA.

\begin{figure}
\centering
\scalebox{0.7}{
\begin{tikzpicture}[node distance=1.25cm]


\node (off1) [circlewhite] { $\times$ };
\node (off2) [circlewhite, below of=off1, xshift=-2.4cm] { $+$ };
\node (off3) [circlewhite, below of=off1, xshift=+2.4cm] { $\exp$ };
\node (off4) [circlewhite, below of=off2, xshift=-1.2cm] { $x$ };
\node (off5) [circlewhite, below of=off2, xshift=+1.2cm] { $\exp $ };
\node (off6) [circlewhite, below of=off3, xshift=-1.2cm] { $x$ };
\node (off7) [circlewhite, below of=off3, xshift=+1.2cm, fill=lightgray] { $/$ };
\node (off8) [circlewhite, below of=off4, xshift=-.6cm, fill=lightgray] { $x$ };
\node (off9) [circlewhite, below of=off4, xshift=+.6cm, fill=lightgray] { $z$ };
\node (off10) [circlewhite, below of=off5, xshift=-.6cm, ] { $y$ };
\node (off11) [circlewhite, below of=off5, xshift=+.6cm, fill=lightgray] { $x$ };

\node (off12) [circlewhite, below of=off6, xshift=-.6cm, fill=lightgray] { $z$ };
\node (off13) [circlewhite, below of=off6, xshift=+.6cm, fill=lightgray] { $z$ };
\node (off14) [circlewhite, below of=off7, xshift=-.6cm, fill=lightgray] { $x$ };
\node (off15) [circlewhite, below of=off7, xshift=+.6cm, fill=lightgray] { $x$ };

\draw [line] (off1)--(off2);
\draw [line] (off1)--(off3);
\draw [line] (off2)--(off4);
\draw [line] (off2)--(off5);
\draw [line] (off3)--(off6);
\draw [line] (off3)--(off7);
 
\draw [line] (off4)--(off8);
\draw [line] (off4)--(off9);
\draw [line] (off5)--(off10);
\draw [line] (off5)--(off11);
\draw [line] (off6)--(off12);
\draw [line] (off6)--(off13);
\draw [line] (off7)--(off14);
\draw [line] (off7)--(off15);

\end{tikzpicture}
}
\vspace{-1.5mm}
\caption{Example of tree for GP-GOMEA with $h=3$ and $r=2$. While 15 nodes are present, the nodes that influence the output are only 7: the gray nodes are introns.}\label{fig:gomea-genotype}
\vspace{-3mm}
\end{figure}
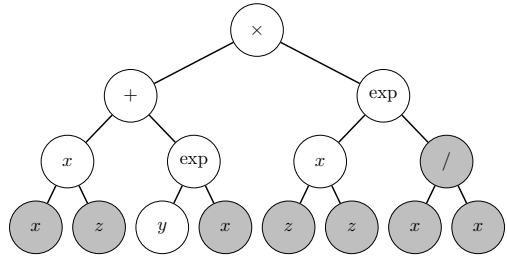

All solutions are generated as \emph{perfect} $r$-ary trees of height $h$, i.e., such that all non-leaf nodes have exactly $r$ children, and leaves are all at maximum depth $h$, with $r$ being the maximum number of inputs accepted by the functions (arity) provided in the function set (e.g., for $\{ +, -, \times \}$, $r=2$), and $h$ chosen by the user. Note that, for any node that is not at maximum depth, $r$ child nodes are appended anyway: no matter if the node is a terminal, or if it is a function requiring less than $r$ inputs (in this case, the leftmost nodes are used as inputs). Some nodes are thus \emph{introns}, i.e., they are not executed to compute the output of the tree. It follows that while trees are \emph{syntactically} redundant, they are not necessarily \emph{semantically} so. All trees of GP-GOMEA have the same number of nodes, equal to $\ell = \sum_{i=0}^h r^i = \frac{r^{h+1}-1}{r-1}$.
Figure~\ref{fig:gomea-genotype} shows a tree used by GP-GOMEA.

\subsection{Linkage learning}\label{sec:linkage-learning}
The linkage model used by GOMEA algorithms is called the Family of Subsets (FOS), and is a set of sets:
\vspace{-1mm}
\begin{equation*}
F = \{ F_1, \dots, F_{|F|} \}, F_i \subseteq \{ 1, \dots, l \}.
\vspace{-1mm}
\end{equation*}
Each $F_i$ (called FOS subset) contains indices representing locations in the genotype. For GP-GOMEA, these indices represent node locations. It is sufficient to choose a parsing order to identify the same node locations in all trees, since trees share the same shape.

In GOMEA, linkage learning corresponds to building a FOS. Different types of FOS exist in literature, however, the one recommended as default is the \emph{Linkage Tree} (LT), by, e.g.,~\citep{thierens2013hierarchical,virgolin2017scalable}. The LT captures linkage on hierarchical levels. An LT is learned every generation, from the population. To assess whether linkage learning plays a key role, i.e. whether it is better than randomly choosing linkage relations, we also consider the Random Tree (RT)~\citep{virgolin2017scalable}.

\subsubsection{Linkage Tree}\label{sec:linkage-tree}
The LT arranges the FOS subsets in a binary tree structure representing hierarchical levels of linkage strength among genotype locations. The LT is built bottom-up, i.e., from the leaves to the root.
The bottom level of the LT, i.e., the leaves, assume that all genotype locations are independent (no linkage), and is realized by instantiating FOS subsets to singletons, each containing a genotype location $i, \forall i \in \{ 1, \dots, \ell \}$.

To build the next levels, mutual information is used as a proxy for linkage strength. Mutual information is a sensible choice to represent linkage strength because it expresses, considering e.g. the pair $(i,j)$ of genotype locations as random variables, the amount of information gained on $i$ given observations on $j$ (and vice versa). In this light, the population can be considered as a set of realizations of the genotype. In particular, the realizations of each genotype location $i$ are what \emph{symbols} appear at location $i$ in the population. In a binary genetic algorithm, symbols are either `$0$' or `$1$', while in GP, symbols correspond to the types of function and terminal nodes, e.g., `$+$',`$-$',`$x_1$',`$x_2$'. In other words, random variables can assume as many values as there are possible symbols in the instruction set\footnote{More symbols can be possible than the number of instructions in case ERCs are used, since instantiating an ERC in a solution results in a constant being randomly sampled.}.

Now, the next step is to compute the mutual information between each and every pair of locations in the genotype of the entire population.
Mutual information between a pair of locations can be computed after measuring entropy for single locations $\text{H}(i)$, and the joint entropy for locations pairs, $\text{H}(i,j)$ (this aspect will be used in Sec.~\ref{sec:improved-linkage-learning}):
\vspace{-1mm}
\begin{gather}\label{eq:MI-definition}
\begin{split}
 &\text{MI}(i,j) = \text{H}(i) + \text{H}(j) - \text{H}(i,j), \text{where} \\
 \text{H}(i) &= - \sum \text{P}_i \log \text{P}_i, \hspace{0.5cm}
\text{H}(i,j) = - \sum \text{P}_{ij} \log \text{P}_{ij},
\vspace{-1mm}
\end{split}
\end{gather}
and $\text{P}_i$ ($\text{P}_{ij}$) is the (joint) probability distribution over the symbols at location(s) $i$ ($i,j$), which can be estimated by counting occurrences of symbol types in the population genotype. This requires to loop over the entire population, and to use nested loops over location pairs $i \in \{ 1, \dots, \ell \}$ and $j \in \{i, \dots, \ell \}$, leading to a time complexity of $O(n^\text{pop} \ell^2 )$. The contribution to the entropy of null probability cases ($-0 \log 0$) is set to $0$.

Given mutual information between location pairs, we approximate linkage among higher orders of locations using the Unweighted Pair Group Method with Arithmetic Mean (UPGMA)~\citep{gronau2007optimal}. To ease understanding, we now provide an explanation of how UPGMA is used to build the rest of the LT that is primarily meant to be intuitive. In practice, we do not use an implementation that strictly adheres to the following explanation, but we use a more advanced algorithm that achieves the same result while having lower time complexity, called the Reciprocal Nearest Neighbor algorithm (RNN). For details on RNN, see~\citep{gronau2007optimal}. 

UPGMA operates in a recursive, hierarchical fashion. Consider each singleton containing a different genotype location $i$ as a cluster $C_i$, and the mutual information between location pairs as a measure of similarity $S$ between clusters, i.e., $S(C_i, C_j) := \text{MI}(i,j)$.
Let $\mathcal{C}$ be the collection of clusters to be parsed, initially containing all location singletons. Every iteration, firstly a new cluster $C_{i^\star} \cup C_{j^\star}$ is formed by joining the clusters $C_{i^\star}, C_{j^\star}$ that have maximal similarity. Secondly, $C_{i^\star}$ and $C_{j^\star}$ are removed from $\mathcal{C}$, and $C_{i^\star} \cup C_{j^\star}$ is inserted in $\mathcal{C}$. When this happens, a FOS subset is added in the LT that corresponds to (contains the same locations of) $C_{i^\star} \cup C_{j^\star}$, as parent of the subsets that represent $C_{i^\star}$ and $C_{j^\star}$. Thirdly, the similarity between $C_{i^\star} \cup C_{j^\star}$ and every other cluster $C_k$ is computed, with:
\vspace{-1mm}
\begin{equation*}
S(C_k, C_{i^\star} \cup C_{j^\star}) = \frac{|C_{i^\star}|}{|C_{i^\star}| + |C_{j^\star}|} S(C_k, C_i) + \frac{|C_{j^\star}|}{|C_{i^\star}| + |C_{j^\star}|} S(C_k, C_j).
\vspace{-1mm}
\end{equation*} 
Iterations are repeated until no more merging is possible, i.e., $\mathcal{C} = \emptyset$. This necessarily happens in $2\ell - 1$ iterations. Note that the last iterations sets the root of the LT, i.e., the subset that contains all genotype locations: $\{1, \dots, \ell \}$. Note also that the structure of the LT is related to the structure of the tree-like genotype of GP solutions only in the sense that the LT contains $2\ell - 1$ FOS subsets and the genotype has length $\ell$, but it is not a one-to-one match to the structure of the genotype.

With the efficient implementation of UPGMA by RNN, the time complexity to build the LT remains bounded by $O(n^\text{pop} \ell^2)$.

\subsubsection{Random Tree}
While linkage learning assumes an inherent structural inter-dependency to be present within the genotype that can be captured in an LT, such hypothesis may not be true. In such a scenario, using the LT might be not better than building a similar FOS in a completely random fashion. The RT is therefore considered to test this. The RT shares the same tree-like structure of the LT, but is built randomly rather than using mutual information (taking $O(\ell)$). We use the RT as an alternative FOS for GP-GOMEA.

\subsection{Gene-pool Optimal Mixing}\label{sec:gom}
Once the FOS is learned, the variation operator GOM generates the offspring population. GOM varies a given solution $\mathcal{P}_i$ in iterative steps, by overriding the nodes at the locations specified by each $F_j$ in the FOS, with the nodes in the same locations taken from random donors in the population. Selection is performed within GOM in a hill-climbing fashion, i.e., variation attempts that result in worse fitness are undone.

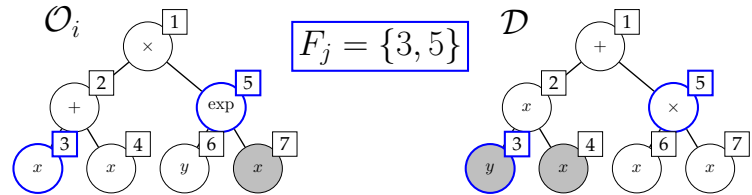
\begin{figure}
\centering
\scalebox{0.65}{
\begin{tikzpicture}[node distance=1.25cm]


\node (off1) [circlewhite] { $\times$ };
\node (off2) [circlewhite, below of=off1, xshift=-1.5cm] { $+$ };
\node (off3) [circlewhite, below of=off1, xshift=+1.5cm, draw=blue, very thick] { $\exp$ };
\node (off4) [circlewhite, below of=off2, xshift=-.75cm, draw=blue, very thick] { $x$ };
\node (off5) [circlewhite, below of=off2, xshift=+.75cm] { $x$ };
\node (off6) [circlewhite, below of=off3, xshift=-.75cm] { $y$ };
\node (off7) [circlewhite, below of=off3, xshift=+.75cm, fill=lightgray] { $x$ };

\draw [line] (off1)--(off2);
\draw [line] (off1)--(off3);
\draw [line] (off2)--(off4);
\draw [line] (off2)--(off5);
\draw [line] (off3)--(off6);
\draw [line] (off3)--(off7);

\node (offlabel) [left of=off1, xshift=-.5cm, yshift=.5cm] {\huge $\mathcal{O}_i$};

\node (offnodelabel1) [squarewhite, right of=off1, xshift=-.7cm, yshift=.5cm, minimum width=.5cm, minimum height=.5cm] {\large 1};
\node (offnodelabel2) [squarewhite, right of=off2, xshift=-.7cm, yshift=.5cm, minimum width=.5cm, minimum height=.5cm] {\large 2};
\node (offnodelabel3) [squarewhite, right of=off3, xshift=-.7cm, yshift=.5cm, minimum width=.5cm, minimum height=.5cm, draw=blue, very thick] {\large 5};
\node (offnodelabel4) [squarewhite, right of=off4, xshift=-.7cm, yshift=.5cm, minimum width=.5cm, minimum height=.5cm, draw=blue, very thick] {\large 3};
\node (offnodelabel5) [squarewhite, right of=off5, xshift=-.7cm, yshift=.5cm, minimum width=.5cm, minimum height=.5cm] {\large 4};
\node (offnodelabel6) [squarewhite, right of=off6, xshift=-.7cm, yshift=.5cm, minimum width=.5cm, minimum height=.5cm] {\large 6};
\node (offnodelabel7) [squarewhite, right of=off7, xshift=-.7cm, yshift=.5cm, minimum width=.5cm, minimum height=.5cm] {\large 7};


\node (don1) [circlewhite, right of=off1, xshift=8cm] { $+$ };
\node (don2) [circlewhite, below of=don1, xshift=-1.5cm] { $x$ };
\node (don3) [circlewhite, below of=don1, xshift=+1.5cm, draw=blue, very thick] { $\times$ };
\node (don4) [circlewhite, below of=don2, xshift=-.75cm, fill=lightgray, draw=blue, very thick] { $y$ };
\node (don5) [circlewhite, below of=don2, xshift=+.75cm, fill=lightgray] { $x$ };
\node (don6) [circlewhite, below of=don3, xshift=-.75cm] { $x$ };
\node (don7) [circlewhite, below of=don3, xshift=+.75cm] { $x$ };

\node (donlabel) [left of=don1, xshift=-.5cm, yshift=.5cm] {\huge $\mathcal{D}$};

\draw [line] (don1)--(don2);
\draw [line] (don1)--(don3);
\draw [line] (don2)--(don4);
\draw [line] (don2)--(don5);
\draw [line] (don3)--(don6);
\draw [line] (don3)--(don7);

\node (donnodelabel1) [squarewhite, right of=don1, xshift=-.7cm, yshift=.5cm, minimum width=.5cm, minimum height=.5cm] {\large 1};
\node (donnodelabel2) [squarewhite, right of=don2, xshift=-.7cm, yshift=.5cm, minimum width=.5cm, minimum height=.5cm] {\large 2};
\node (donnodelabel3) [squarewhite, right of=don3, xshift=-.7cm, yshift=.5cm, minimum width=.5cm, minimum height=.5cm, draw=blue, very thick] {\large 5};
\node (donnodelabel4) [squarewhite, right of=don4, xshift=-.7cm, yshift=.5cm, minimum width=.5cm, minimum height=.5cm, draw=blue, very thick] {\large 3};
\node (donnodelabel5) [squarewhite, right of=don5, xshift=-.7cm, yshift=.5cm, minimum width=.5cm, minimum height=.5cm] {\large 4};
\node (donnodelabel6) [squarewhite, right of=don6, xshift=-.7cm, yshift=.5cm, minimum width=.5cm, minimum height=.5cm] {\large 6};
\node (donnodelabel7) [squarewhite, right of=don7, xshift=-.7cm, yshift=.5cm, minimum width=.5cm, minimum height=.5cm] {\large 7};


\node (Fj) [squarewhite, right of=off1, xshift=3.5cm, draw=blue, very thick] {\huge $F_j = \{ 3, 5 \}$};

\end{tikzpicture}
}
\vspace{-1.5mm}
\caption{Example of variation step performed by GOM for trees with $h=2$. Squares on top of each node indicate the node location according to pre-order traversal (depth-first). GOM replaces the nodes of $\mathcal{O}_i$ of which the location is specified by $F_j$, with the homologous nodes of $\mathcal{D}$ (blue contour).}\label{fig:gomvariationstep}
\vspace{-3mm}
\end{figure}

The pseudo-code presented in Algorithm~\ref{alg:gom} describes GOM in detail. To begin, a backup $\mathcal{B}_i$ of the parent solution $\mathcal{P}_i$ is made, including its fitness, and similarly an offspring solution $\mathcal{O}_i = \mathcal{P}_i$ is created. Next, the FOS $F$ is shuffled randomly: this is to provide different combinations of variation steps along the run and prevent bias. For each set of node locations $F_j$, a random donor $\mathcal{D}$ is then picked from the population, and $\mathcal{O}_i$ is changed by replacing the nodes specified by $F_j$ with the homologous ones from $\mathcal{D}$. This process is exemplified in Fig.~\ref{fig:gomvariationstep}. It is then assessed whether at least one (syntactic) non-intron node of the tree has been changed by variation (indicated by $ \neq^\star$ in line~\ref{line:gom-ineq}). When that is not the case, $\mathcal{O}_i$ will have the same behavior as $\mathcal{B}_i$, thus the fitness is necessarily identical. Otherwise, the new fitness $f_{\mathcal{O}_i}$ is computed: if not worse than the previous one, the change is kept, and the backup is updated, otherwise the change is reversed.

Note that if a change results in $f_{\mathcal{O}_i} = f_{\mathcal{B}_i}$, the change is kept. This allows random walks in the neutral fitness landscape~\citep{ebner2001neutral,sadowski2013usefulness}. 
Note also that differently from traditional subtree crossover and subtree mutation~\citep{koza1992gp}, GOM can change unconnected nodes at the same time, and keeps tree height limited to the initially specified parameter $h$. Finally, GOM \emph{does not consider} any FOS subset that contains all node locations, i.e., $F_j=\{ 1, \dots, \ell \}$, as using such subset would mean to entirely replace $\mathcal{O}_i$ with $\mathcal{D}$.

\begin{algorithm}
\caption{Pseudocode of GOM}\label{alg:gom}
\begin{algorithmic}[1]
\Procedure{\texttt{GOM}}{$\mathcal{P}_i,\mathcal{P},F$}
\State $\mathcal{B}_i \gets \mathcal{P}_i$; $f_{\mathcal{B}_i} \gets f_{\mathcal{P}_i}$; $\mathcal{O}_i \gets \mathcal{P}_i$ 
\State $F \gets $\texttt{randomShuffle($F$)}
\For {$F_j \in F$}
	\State $\mathcal{D} \gets $\texttt{pickRandomDonor($\mathcal{P}$)}
	\State $\mathcal{O}_i \gets $\texttt{overrideNodes($\mathcal{O}_i,\mathcal{D},F_j$)}
	\If {$\mathcal{O}_i \neq^\star \mathcal{B}_i$}\label{line:gom-ineq}
		\State $f_{\mathcal{O}_i} \gets $\texttt{computeFitness($\mathcal{O}_i$)}
		\If {$f_{\mathcal{O}_i} \leq f_{\mathcal{B}_i}$}\label{line:gom-fitness-check} \textit{\hspace{3.3cm} \#Assumption: minimization of} $f$
			\State $\mathcal{B}_i \gets \mathcal{O}_i$; $f_{\mathcal{B}_i} \gets f_{\mathcal{O}_i}$
		\Else
			\State $\mathcal{O}_i \gets \mathcal{B}_i$; $f_{\mathcal{O}_i} \gets f_{\mathcal{B}_i}$
		\EndIf
	\Else
		\State $\mathcal{B}_i \gets \mathcal{O}_i$
	\EndIf
\EndFor
\EndProcedure
\end{algorithmic}
\end{algorithm}

\section{General experimental settings}\label{sec:expsettings}
We now describe the general parameters that will be used in this article. 
Table~\ref{tab:general-parameters} reports the parameter settings which are typically used in the following experiments, unless specified otherwise. The notation $\mathbf{x}$ represents the matrix of feature values. We use the Analytic Quotient (AQ)~\citep{ni2013use} instead of protected division. This is because the AQ is continuous in 0 for the second operand: $x_1 \div_\text{AQ} x_2 \coloneqq x_1 / \sqrt{1 + x^2_2}$. Albeit continuity is not needed by many GP variation operators (including GOM), it is useful at prediction time: \citep{ni2013use} show that using the AQ helps generalization (whereas using protected division does not). However, the AQ may be considered relatively hard to interpret.

As mentioned in the introduction, we focus on the evolution of solutions that are constrained to be small, to \emph{enable} interpretability. We choose $h=4$ because this results in relatively balanced trees with up to $31$ nodes (since $r=2$). We consider this size limitation a critical value: for the given function set, we found solutions to be already borderline interpretable for us (this is discussed further in Sec.~\ref{sec:discussion}). Larger values for $h$ would therefore play against the aim of this study. When benchmarking GP-GOMEA in Sec.~\ref{sec:benchmarking}, we also consider $h=3$ and $h=5$ for completeness.

We consider 10 real-world benchmark datasets from literature~\citep{martins2018solving} that can be found on the UCI repository\footnote{\url{https://archive.ics.uci.edu/ml/index.php}}~\citep{asuncion2007uci} and other sources\footnote{\url{https://goo.gl/tn6Zxv}}. The characteristics of the datasets are summarized in Table~\ref{tab:datasets}. 

We use the linearly-scaled Mean Squared Error (MSE) to measure solution fitness~\citep{keijzer2003improving}, as it can be particularly beneficial when evolving small solutions. This means a fast (cost $O(n)$ with $n$ number of dataset examples) linear regression is applied between the target $y$ and the solution prediction $\tilde{y}$ prior to computing the MSE.
We present our results in terms of variance-Normalized MSE (NMSE), i.e. $100 \times \frac{\text{MSE}(y, \tilde{y})}{var(y)}$, so that results from different datasets are on a similar scale (the $100\times$ factor is added for readability).

To assess statistical significance when comparing the results of multiple executions of two algorithms (or configurations) on a certain dataset, we use the Wilcoxon signed-rank test~\citep{demvsar2006statistical}. This test is set up to compare competing algorithms based on the same prior conditions. In particular, we employ pairs of executions where the dataset is split into identical training, validation, and test sets for both algorithms being tested. This is because the particular split of data determines the fitness function (based on the training set), and the achievable generalization error (for the validation and test sets). We consider a difference to be significant if a smaller $p$-value than $0.05 / \beta$ is found, with $\beta$ the Bonferroni correction coefficient, used to prevent false positives. If more than two algorithms need to be compared, we first perform a Friedman test on mean performance over all datasets~\citep{demvsar2006statistical}.
We use the symbols \btC{black}, \wtC{black} to respectively indicate significant superiority, and inferiority (absence of a symbol means no significant difference). The result \emph{next} to the symbol \btC{black} (\wtC{black}) signifies a result being better (worse) than the result obtained by the algorithm that has the same color of the symbol. Algorithms and/or configurations are color coded in each table reporting results (colors are color-blind safe).

\begin{table}
\centering
\caption{General parameter settings for the experiments}\label{tab:general-parameters}
\scriptsize
\begin{tabular}{lc}
\toprule
Parameter & Setting \\
\midrule
Function set & $\{ +, -, \times,  \div_\text{AQ} \}$\\
Terminal set & $\mathbf{x} \cup \{\text{ERC}\}$\\
ERC bounds & $[ \min \mathbf{x}, \max \mathbf{x}] $\\
Initialization for GP-GOMEA & Half-and-Half as in~\citep{virgolin2018symbolic} \\
Tree height $h$ & 4\\
Train-validation-test split & $ 50$\%--$25$\%--$25$\% \\
Experiment repetitions & $30$\\
\bottomrule
\end{tabular}
\vspace{-3mm}
\end{table}

\begin{table}
\caption{Regression datasets used in this work}\label{tab:datasets}
\centering
\scriptsize
\scalebox{0.9}{
\begin{tabular}{lccc}
\toprule
Name & Abbreviation & \# Features & \# Examples \\
\midrule
Airfoil & Air & 5 & 1503 \\
Boston housing & Bos & 13 & 506 \\
Concrete compres. str. & Con & 8 & 1030\\
Dow chemical & Dow & 57 & 1066\\
Energy cooling & EnC & 8 & 768\\
Energy heating & EnH & 8 & 768\\
Tower & Tow & 25 & 4999\\
Wine red & WiR & 11 & 1599\\
Wine white & WiW & 11 & 4898\\
Yacht hydrodynamics & Yac & 6 & 308\\ 
\bottomrule
\end{tabular}
}
\vspace{-3mm}
\end{table}

\section{Improving linkage learning for GP}\label{sec:improved-linkage-learning}
In previous work on GP-GOMEA, learning the LT was performed the same way it is done for any discrete GOMEA implementation, i.e. by computing the mutual information between pairs of locations $(i,j)$ in the genotype (Eq.~\ref{eq:MI-definition})~\citep{virgolin2017scalable}. However, the distribution of node types is typically not uniform when a GP population is initialized (e.g., function nodes never appear as leaves). In fact, this depends on the cardinality of the function and terminal sets, on the arity of the functions, and on the population initialization method (e.g., \emph{Full}, \emph{Grow}, \emph{Half-and-Half}, \emph{Ramped Half-and-Half}~\citep{luke2001survey}). Note that it does not depend on the particular dataset in consideration (except in that the number of features determines the size of the terminal set). The lack of uniformity in the distribution leads to the emergence of mutual information between particular parts of the genotype. Crucially, this mutual information is natural to the solution representation, the sets of symbols and the initialization process.

\begin{figure}
\centering
\begin{tabular}{c}
$\text{MI}, n^\text{pop}=10^6$\\
\includegraphics[width=.4\linewidth]{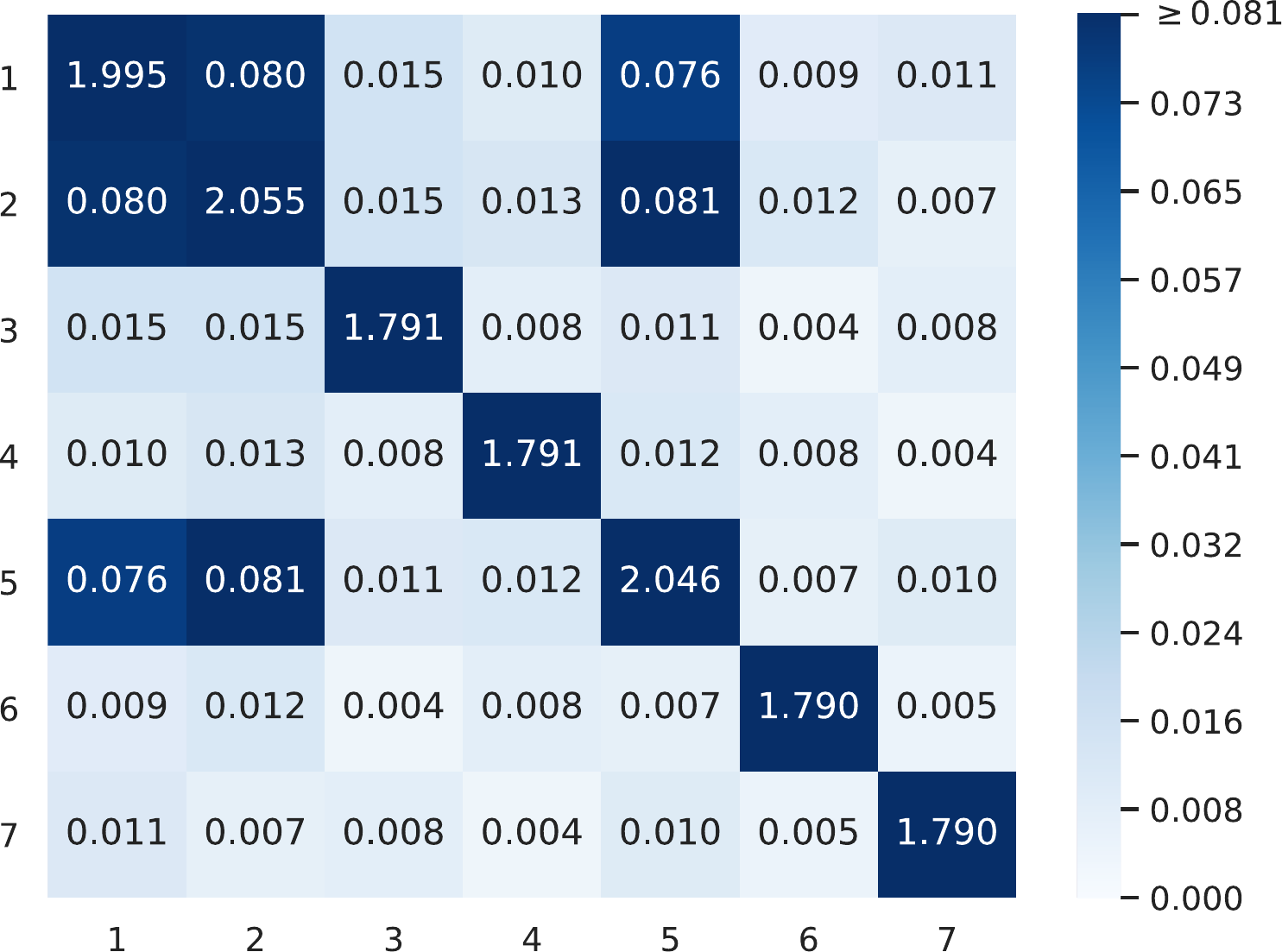}\\
\end{tabular}\vspace{-3mm}
\caption{Mutual information matrix between pairs of locations in the genotype (x and y labels). Darker blue represents higher values. The matrix is computed for an initialized population of size $10^6$. The values suggest the existence of linkage even though no evolution has taken place yet.}\label{fig:mis_1}
\vspace{-3mm}
\end{figure}

If mutual information is used to represent linkage, then linkage will already be observed at initialization. However, it is reasonable to expect no linkage to be present in an initialized population, as evolution did not take place yet.
Figure~\ref{fig:mis_1} shows the mutual information matrix between pairs of node locations in an initial population of $1,000,000$ solutions with maximum height $h=2$, using \emph{Half-and-Half}, a function set of size 4 with maximum number of inputs $r=2$, and a terminal set of size 6 (no ERCs are used). Each tree contains exactly $7$ nodes. We index node locations with pre-order tree traversal, i.e., $1$ is the root, $2$ its first child, $5$ its second child, $3,4$ are (leaves) children of $2$, and $6,7$ are (leaves) children of $5$. Nodes at locations 2 and 5 can be functions only if a function is sampled at node 1. It can be seen that the mutual information matrix of location pairs (correctly) captures the non-uniformity in the initial distribution (i.e., larger mutual information values are present between non-leaf nodes). Using mutual information directly as a proxy for linkage may be undesirable.

\subsection{Biasing mutual information to represent linkage}
We propose to overcome the aforementioned problem by measuring linkage with a modified version of the mutual information, such that no linkage is measured at initialization. Our hypothesis is that, if we apply such a correction so that no patterns are identified at initialization, the truly salient patterns will have a bigger chance of emerging during evolution, and better results will be achieved.

Let us consider the scenario where, at initialization, symbols are uniformly distributed. For example, this typically happens in binary genetic algorithms. The mutual information between pairs of genotype locations that is expected at initialization, i.e., at generation $g=1$ before variation and selection, will then correspond to the identity matrix: $\text{MI}^{g} |_{g=1} = I$ (assuming binary symbols and mutual information in bits as well as a sufficiently large population size). This mutual information matrix is suitable to represent linkage as no linkage should be present at initialization.

We propose to adopt a biased mutual information matrix $\text{MI}_b(i,j)$ to represent the linkage between a pair of genotype locations $(i,j)$, that has the property:
\vspace{-1mm}
\begin{equation*}
\text{MI}^g_b(i,j) |_{g=1} = I, 
\vspace{-1mm}
\end{equation*}
no matter the actual distribution of the initial population.

To this end, we use Eq.~\ref{eq:MI-definition}, i.e., we manipulate the entropy terms, to represent maximal randomness to be present at initialization for each genotype location. In particular, we propose to use biased entropy metrics such that $\text{H}^g_b(i) |_{g=1} = 1$ and $\text{H}^g_b(i,j) |_{g=1} = 2$ (for $i \neq j$) , since
\vspace{-1mm}
\begin{align*}
\text{MI}^g_b(i,j) |_{g=1} &= \left( \text{H}^g_b(i)  + \text{H}^g_b(j) - \text{H}^g_b(i,j)  \right) |_{g=1}\\ 
&= 1 + 1 - 2 = 0 \text{ \hspace{2cm} (for } i \neq j, \text{ else } 1 \text{)}.
\vspace{-1mm}
\end{align*}
We propose to use linear biasing coefficients $\beta_i$ ($\beta_{i,j}$) to have the general biased entropy for any generation $g$ as $\text{H}^g_b(i) = \beta_i \text{H}(i)$ and $\text{H}^g_b(i,j) = \beta_{i,j} \text{H}(i,j)$, with $\beta_i = \left( \text{H}^g_b(i) |_{g=1} \right)^{-1}$ and $\beta_{i,j} = 2 \left( \text{H}^g(i,j) |_{g=1} \right)^{-1}$ to enforce maximal randomness at initialization.

To determine the beta coefficients \emph{exactly} means to know the true distribution inferred by the sampling process used to sample the initial population, and thus the true initial entropy for each genotype location. However, this is generally not trivial to determine for GP, since a number of factors need to be considered. E.g., if the \emph{Ramped Half-and-Half} initialization method is used, what symbol is sampled at a location depends on the chance to use \emph{Full} or \emph{Grow}, the chance to pick the function or the terminal set based on the depth, the size of these sets, and possibly other problem-specific factors. Hence, we propose to simply approximate the $\beta$ coefficients by using the $\text{H}^g(i) |_{g=1}$ measured on the initial population, assuming the population to be large enough.

Summing up, the pairwise linkage estimation we propose to use at generation $g$, for a pair of locations $(i,j)$, will be:
\vspace{-1mm}
\begin{equation}
\text{MI}^g_{\tilde{b}} (i,j) = \beta_i \text{H}^g(i) + \beta_j \text{H}^g(j) - \beta_{i,j} \text{H}^g(i,j). 
\vspace{-1mm}
\end{equation}
The tilde in $\tilde{b}$ is to remark that this is an approximation.

\subsection{Estimation of linkage by MI\textsubscript{$\tilde{b}$}}
As a preliminary step, we observe what linkage values are obtained between pairs of genotype locations by using $\text{MI}_{\tilde{b}}$. For conciseness, in the following we denote $\text{MI}^g_{\tilde{b}} (i,j) |_{g=\Gamma}$ with $\text{MI}^\Gamma_{\tilde{b}} (i,j)$. We show the MI matrix computed at the second generation of a GP-GOMEA run on the dataset Yac ($\text{MI}^{1}_{\tilde{b}} = I$ by construction). We do this for two population sizes, $n^\text{pop} = 10$ and $n^\text{pop}=10^6$. We expect that, the bigger $n^\text{pop}$ is, the closer $\text{MI}^{2}_{\tilde{b}}$ is to $I$.

\begin{figure}
\centering
\setlength{\tabcolsep}{0.5em}
\begin{tabular}{cc}
$\text{MI}^2_{\tilde{b}}, n^\text{pop}=10^1$ & $\text{MI}^2_{\tilde{b}}, n^\text{pop}=10^6$\\
\includegraphics[width=.4\linewidth]{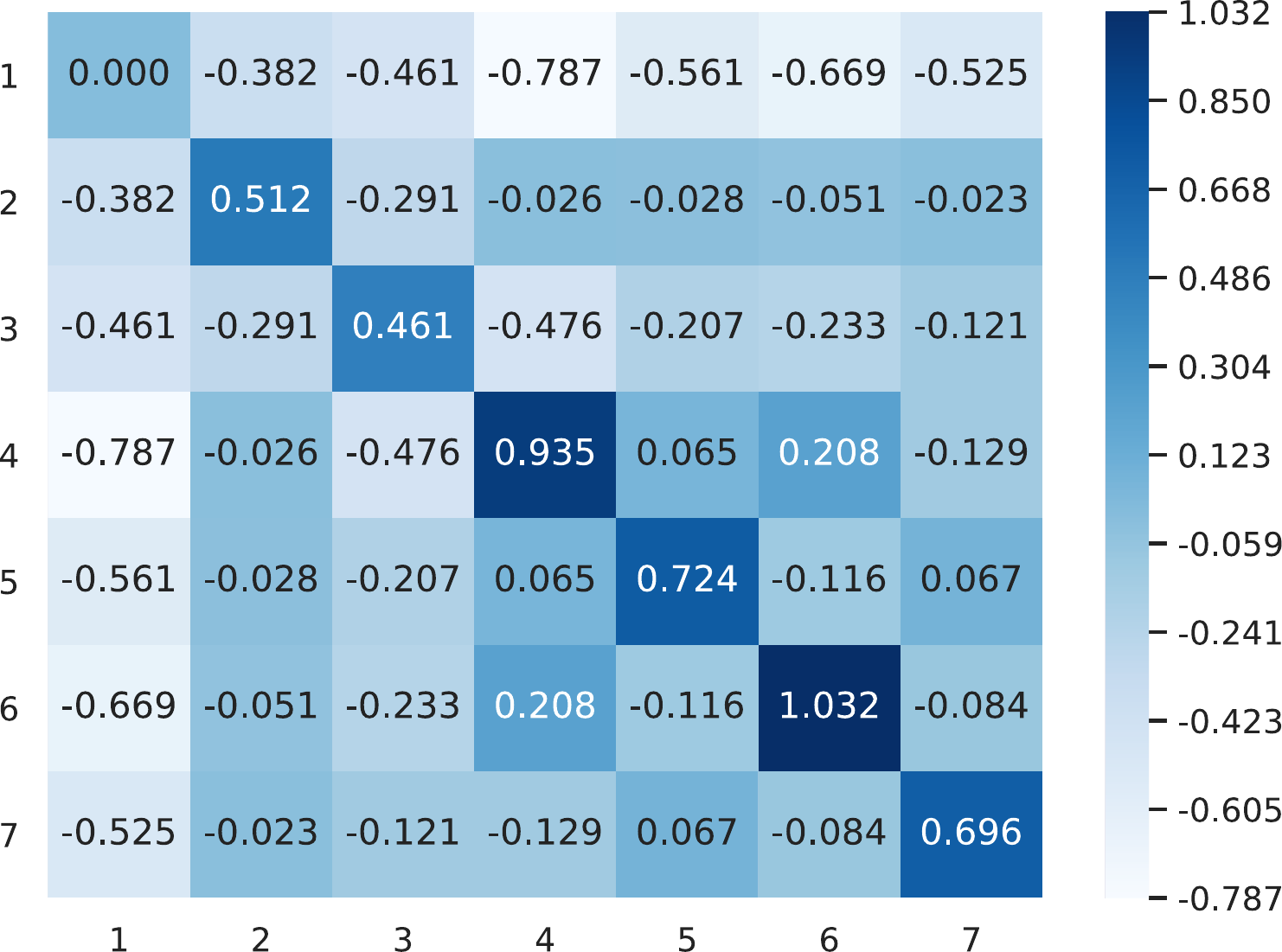}
&
\includegraphics[width=.4\linewidth]{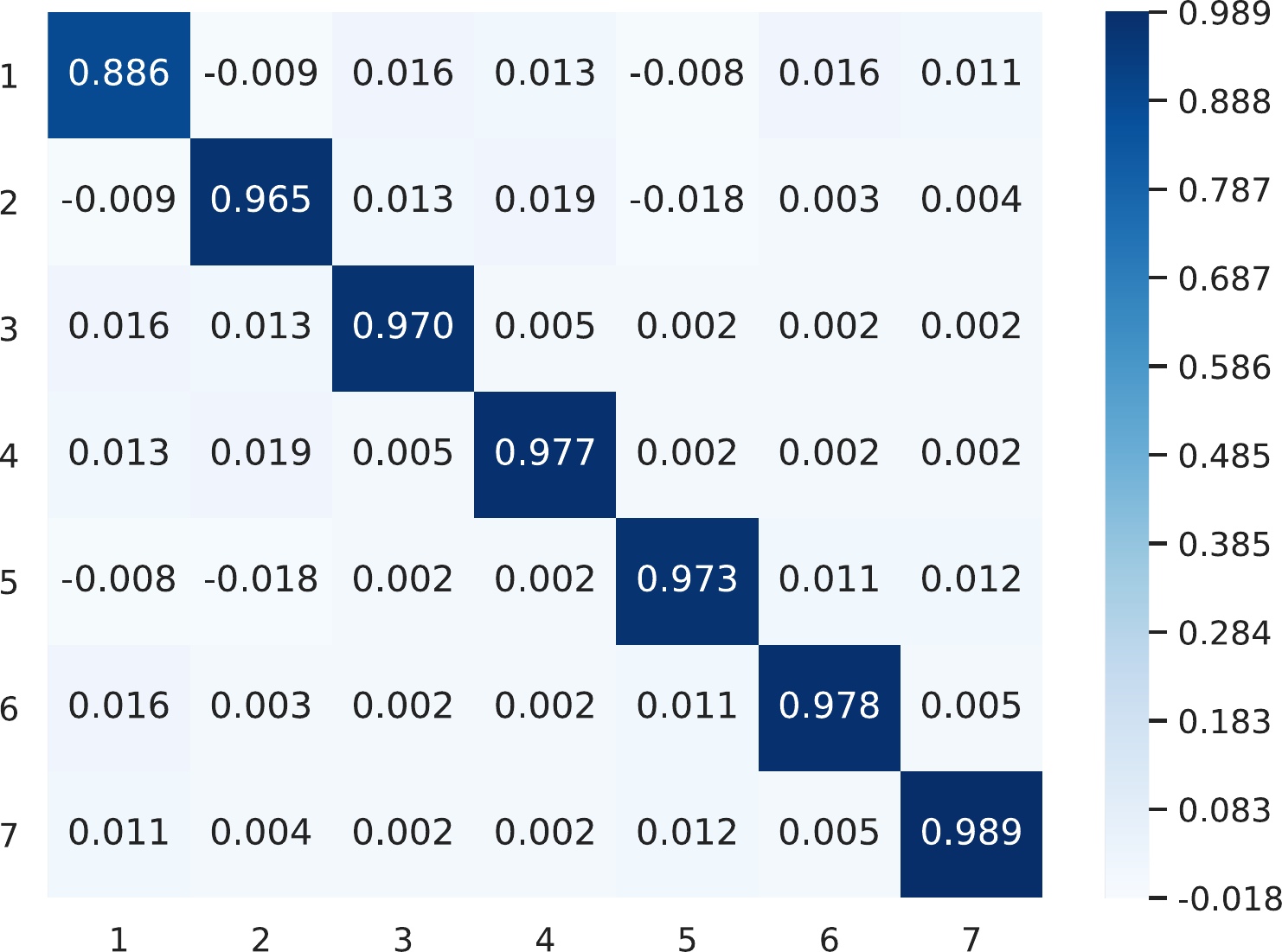}
\end{tabular}
\caption{Mutual information matrices at the second generation using our biasing method to better represent linkage, with population size of $10$ (left), and of $10^6$ (right) for a particular run of GP-GOMEA. The rightmost matrix is closest to the identity $I$.}
\label{fig:mis}\vspace{-3mm}
\end{figure}

We use the parameters of Table~\ref{tab:general-parameters}, a terminal set of size 6 (the features of Yac, no ERC) and $h=2$, i.e. $ \ell = 7$ nodes per tree. Figure~\ref{fig:mis} shows the biased mutual information matrix between location pairs, for the two population sizes. It can be seen that the values can be lower than 0 or bigger than 1. However, while this is particularly marked for $n^\text{pop}=10$, with minimum of -0.787 and maximum of 1.032, it becomes less evident for $n^\text{pop}=10^6$, with minimum of -0.018 and maximum of 0.989. The fact that $\text{MI}^2_{\tilde{b}} \approx I$ for $n^\text{pop}=10^6$ is because, with such a large population size, considerable diversity is still present in the second generation.

\subsection{Experiment: LT--$\text{MI}_{\tilde{b}}$ vs LT--MI vs RT}\label{sec:exp-normalized-mi}
We now test the use of $\text{MI}_{\tilde{b}}$ over the standard $\text{MI}$ for GP-GOMEA with the LT. We denote the two configurations with LT--$\text{MI}_{\tilde{b}}$ and LT--MI. We also consider the RT to see if mutual information drives variation better than random information.

We set the population size to 2000 as a compromise between having enough samples for linkage to be learned, and meeting typical literature values, which range from hundreds to a few thousands. We use the function set of Table~\ref{tab:general-parameters}, and a tree height $h=4$ (thus $\ell=31$). We set a limit of 20 generations, which corresponds to approximately 1200 generations of traditional GP, as each solution is evaluated up to $2\ell-2$ times (size of the LT minus its root and non-meaningful changes, see Sec.~\ref{sec:linkage-learning} and~\ref{sec:gom}).

\begin{table}
\caption{Median NMSE of 30 runs for GP-GOMEA with LT--$\text{MI}_{\tilde{b}}$, LT--MI, and RT.}\label{tab:exp-1-results}
\centering
\scriptsize
\scalebox{1.0}{
\setlength{\tabcolsep}{0.1em}
\begin{tabular}{l S c S c S c c S c S c S c}
\toprule 
 & \multicolumn{6}{c}{Training} & \ssW\ssW & \multicolumn{6}{c}{Test} \\
Dataset & \multicolumn{2}{c}{\color{cbScarlet}LT--$\text{MI}_{\tilde{b}}$} & \multicolumn{2}{c}{\color{cbBlue}LT--MI} & \multicolumn{2}{c}{\color{cbTeal}RT} & 

& \multicolumn{2}{c}{\color{cbScarlet}LT--$\text{MI}_{\tilde{b}}$} & \multicolumn{2}{c}{\color{cbBlue}LT--MI} & \multicolumn{2}{c}{\color{cbTeal}RT} \\
\midrule

Air &
\ssW{}29.9 & \btC{cbBlue}{}\btC{cbTeal}{} & 
\ssW{}31.2 & \wtC{cbScarlet}{}\ssW{} &  
\ssW{}32.7 & \wtC{cbScarlet}{}\ssW{} &
& 
\ssW{}31.8 &\btC{cbBlue}{}\btC{cbTeal}{} & 
\ssW{}34.8 &\wtC{cbScarlet}{}\ssW{}& 
\ssW{}34.0 &\wtC{cbScarlet}{}\ssW{}\\

Bos & 
\ssW{}15.4 & \btC{cbBlue}{}\btC{cbTeal}{} & 
\ssW{}15.4 & \wtC{cbScarlet}{}\btC{cbTeal} & 
\ssW{}17.5  & \wtC{cbScarlet}{}\wtC{cbBlue}{} &
& 
\ssW{}24.0 & \wtC{cbBlue}{}\wtC{cbTeal} & 
\ssW{}23.0 & \btC{cbScarlet}{}\ssW{} & 
\ssW{}22.5 & \btC{cbScarlet}{}\ssW{} \\

Con & 
\ssW{}17.5 &\btC{cbBlue}{}\btC{cbTeal}{} & 
\ssW{}18.5 & \wtC{cbScarlet}{}\btC{cbTeal} & 
\ssW{}19.0 & \wtC{cbScarlet}{}\wtC{cbBlue}{}  & 
& 
\ssW{}18.7 & \btC{cbBlue}{}\btC{cbTeal}{} & 
\ssW{}19.6 & \wtC{cbScarlet}{}\btC{cbTeal} & 
\ssW{}20.1 & \wtC{cbScarlet}{}\wtC{cbBlue} \\

Dow & 
\ssW{}20.9 & \wtC{cbBlue}{}\btC{cbTeal} & 
\ssW{}20.3 & \btC{cbScarlet}{}\btC{cbTeal} & 
\ssW{}24.0 & \wtC{cbScarlet}{}\wtC{cbBlue}{}  &
& 
\ssW{}22.6 & \wtC{cbBlue}{}\btC{cbTeal} & 
\ssW{}21.1 & \btC{cbScarlet}{}\btC{cbTeal} & 
\ssW{}26.0 & \wtC{cbScarlet}{}\wtC{cbBlue} \\

EnC & 
\ssW{}8.42 & \btC{cbBlue}{}\btC{cbTeal}{} & 
\ssW{}9.68 & \wtC{cbScarlet}{}\wtC{cbTeal}{} & 
\ssW{}9.09 & \wtC{cbScarlet}{}\btC{cbBlue}{} &
& 
\ssW{}9.18 & \btC{cbBlue}{}\btC{cbTeal}{} & 
\ssW{}10.7 & \wtC{cbScarlet}{}\wtC{cbTeal}{} & 
\ssW{}10.3 & \wtC{cbScarlet}{}\btC{cbBlue}{}\\

EnH & 
\ssW{}6.24 & \btC{cbBlue}{}\btC{cbTeal} & 
\ssW{}6.44 &  \wtC{cbScarlet}{}\wtC{cbTeal}{} & 
\ssW{}6.40  & \wtC{cbScarlet}{}\btC{cbBlue}{} &
& 
\ssW{}6.50 & \btC{cbBlue}{}\btC{cbTeal}{} & 
\ssW{}7.10 & \wtC{cbScarlet}{}\wtC{cbTeal}{} & 
\ssW{}6.70 & \wtC{cbScarlet}{}\btC{cbBlue}{}\\

Tow & 
\ssW{}12.5 &\wtC{cbBlue}{}\btC{cbTeal} & 
\ssW{}12.5 &\btC{cbScarlet}{}\btC{cbTeal} & 
\ssW{}13.1 & \wtC{cbScarlet}{}\wtC{cbBlue}{}  &
& 
\ssW{}13.0 & \ssW{}\btC{cbTeal}{} & 
\ssW{}12.8 & \ssW{}\btC{cbTeal} & 
\ssW{}13.2 & \wtC{cbScarlet}{}\wtC{cbBlue}\\

WiR & 
\ssW{}60.3 &\btC{cbBlue}{}\btC{cbTeal} & 
\ssW{}60.9 &\wtC{cbScarlet}{}\btC{cbTeal} & 
\ssW{}61.2 &\wtC{cbScarlet}{}\wtC{cbBlue}{}  &
& 
\ssW{}62.5 & \btC{cbBlue}{}\btC{cbTeal}{} & 
\ssW{}63.0 & \wtC{cbScarlet}{}\ssW{} & 
\ssW{}63.1 & \wtC{cbScarlet}{}\ssW{}\\

WiW & 
\ssW{}68.1 &\btC{cbBlue}{}\btC{cbTeal} & 
\ssW{}68.4 & \wtC{cbScarlet}{}\ssW{} & 
\ssW{}68.7 & \wtC{cbScarlet}{}\ssW{} &
& 
\ssW{}69.1 &\btC{cbBlue}{}\btC{cbTeal}{} & 
\ssW{}69.7 &\wtC{cbScarlet}{}\wtC{cbTeal}{} & 
\ssW{}69.5 & \wtC{cbScarlet}{}\btC{cbBlue}{}\\

Yac & 
\ssW{}0.34 &\btC{cbBlue}{}\btC{cbTeal} & 
\ssW{}0.37 & \wtC{cbScarlet}{}\ssW{} & 
\ssW{}0.36 & \wtC{cbScarlet}{}\ssW{}  &
& 
\ssW{}0.58 &\btC{cbBlue}{}\btC{cbTeal}{} & 
\ssW{}0.62 & \wtC{cbScarlet}{}\wtC{cbTeal}{} & 
\ssW{}0.62 & \wtC{cbScarlet}{}\btC{cbBlue}{} \\
\bottomrule
\end{tabular}
}
\vspace{-3mm}
\end{table}

\subsection{Results: LT--$\text{MI}_{\tilde{b}}$ vs LT--MI vs RT}
The training and test NMSE performances are reported in Table~\ref{tab:exp-1-results}. The Friedman test results in significant differences along training and test performance. GP-GOMEA with LT--$\text{MI}_{\tilde{b}}$ is clearly the best performing algorithm, with significantly lower NMSE compared to LT--MI on 8/10 datasets when training, and 7/10 at test time. It is always better than using the RT when training, and in 9/10 cases when testing. The LT--MI is comparable with the RT for these problems.

\begin{figure}
\centering
\setlength{\tabcolsep}{0.05em}
\def\arraystretch{0.8}
\small
\begin{tabular}{cc}
\begin{sideways}\hspace{1.3cm}Training NMSE\end{sideways}& \includegraphics[width=0.50\linewidth]{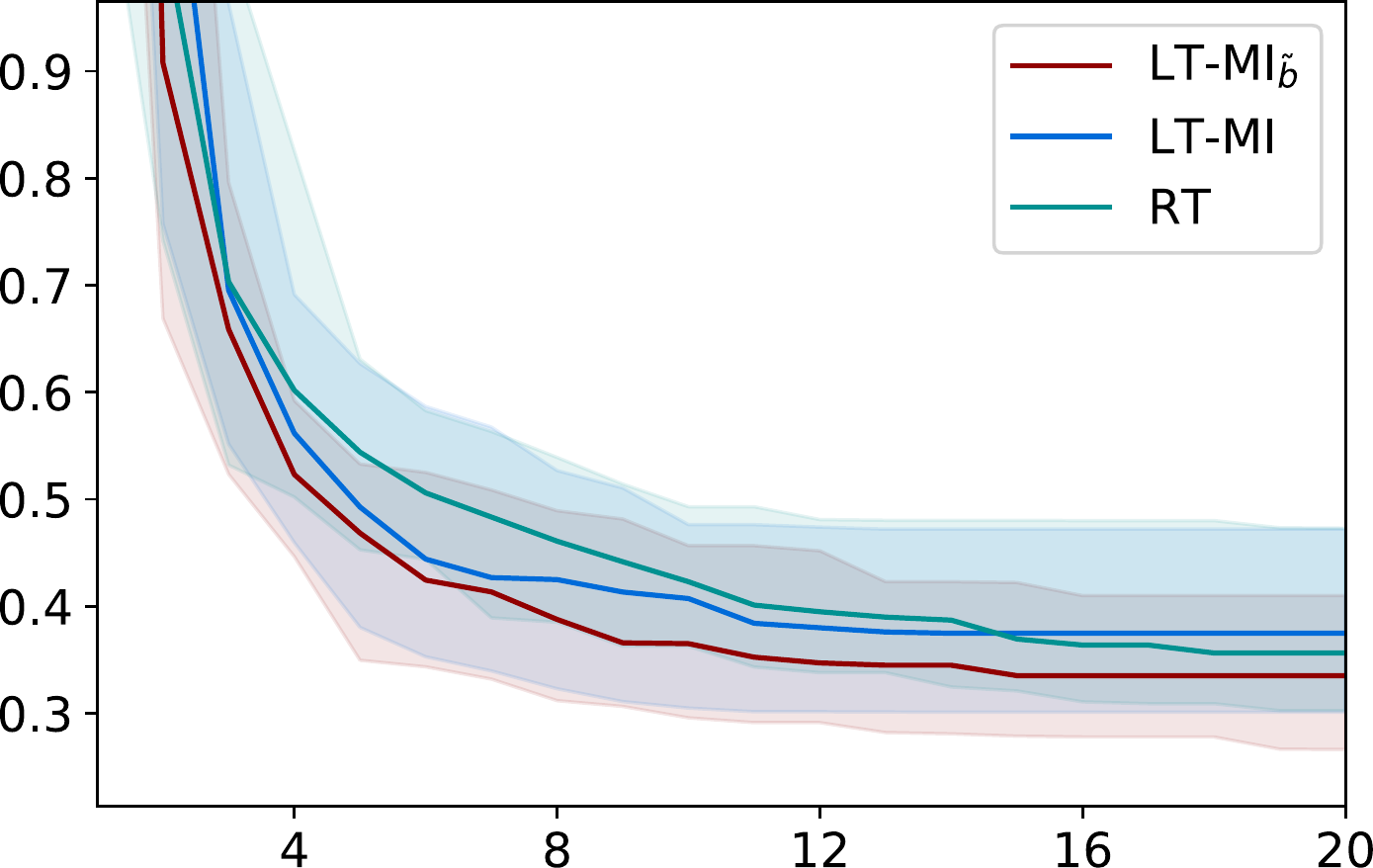}
\\
& \hspace{0.5cm}Generations \\
\end{tabular}
\vspace{-0.2cm}
\caption{Median fitness of the best solution of 30 runs  on Yac, for LT--$\text{MI}_{\tilde{b}}$, LT--MI, and RT ($10^\text{th}$ and $90^\text{th}$ percentiles in shaded area).}\label{fig:exp1-yacht}
\end{figure}

The result of this experiment is that the use of the new $\text{MI}_{\tilde{b}}$ to build the LT simply enables GP-GOMEA to perform a more competent variation than the use of MI. Also, using the LT this way leads to better results than when making random changes with the RT. Figure~\ref{fig:exp1-yacht} shows the evolution of the training NMSE for the dataset Yac. It can be seen that the LT--$\text{MI}_{\tilde{b}}$ allows to quickly reach smaller errors than the other two FOS types. We observed similar training patterns for the other datasets (not shown here). 

In the remainder, when we write ``LT'', we refer to LT--$\text{MI}_{\tilde{b}}$.

\subsection{Experiment: assessing propagation of node patterns}\label{sec:understanding-linkage}
The previous experiment showed that using linkage-driven variation (LT) can be favorable compared to random variation (RT). This seems to confirm the hypothesis that, in certain SR problems, salient underlying patterns of nodes exist in the genotype that can be exploited.
Another aspect that can be considered w.r.t. such hypothesis is how final solutions look: if linkage learning identifies specific node patterns, it can be expected that their propagation will lead to the discovery of similar solutions over different runs.

Therefore, we now want to assess whether the use of the LT has a bigger chance to lead to the discovery of a particular best-of-run solution, compared to the use of the RT. 
We use the same parameter setting as described in Sec.~\ref{sec:exp-normalized-mi}, but perform 100 repetitions. While each run uses a different random seed (e.g., for population initialization), we fix the dataset split, as changing the training set results in changing the fitness function. We repeat the 100 runs on 5 random dataset splits, on the smallest dataset Yac.
Together with $n^\text{pop}=2000$ as in the previous experiment, we also consider a doubled $n^\text{pop}=4000$.

\begin{figure}
\centering
\setlength{\tabcolsep}{0.15em}
\def\arraystretch{1.0}
\begin{tabular}{ccc}
\begin{sideways}\hspace{0.5cm} \small \# Duplicates\end{sideways}&
\includegraphics[width=.35\linewidth]{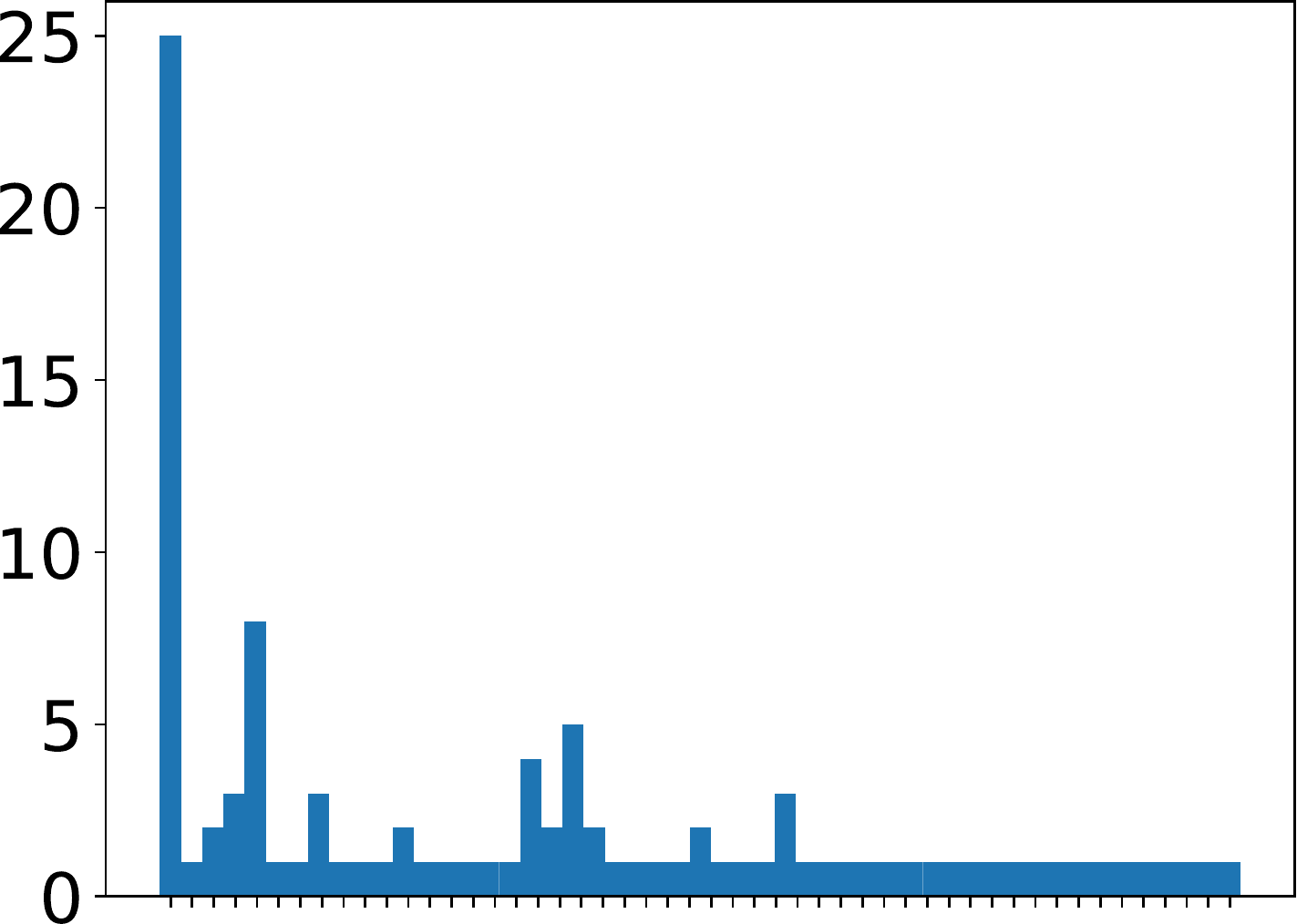} & 
\includegraphics[width=.35\linewidth]{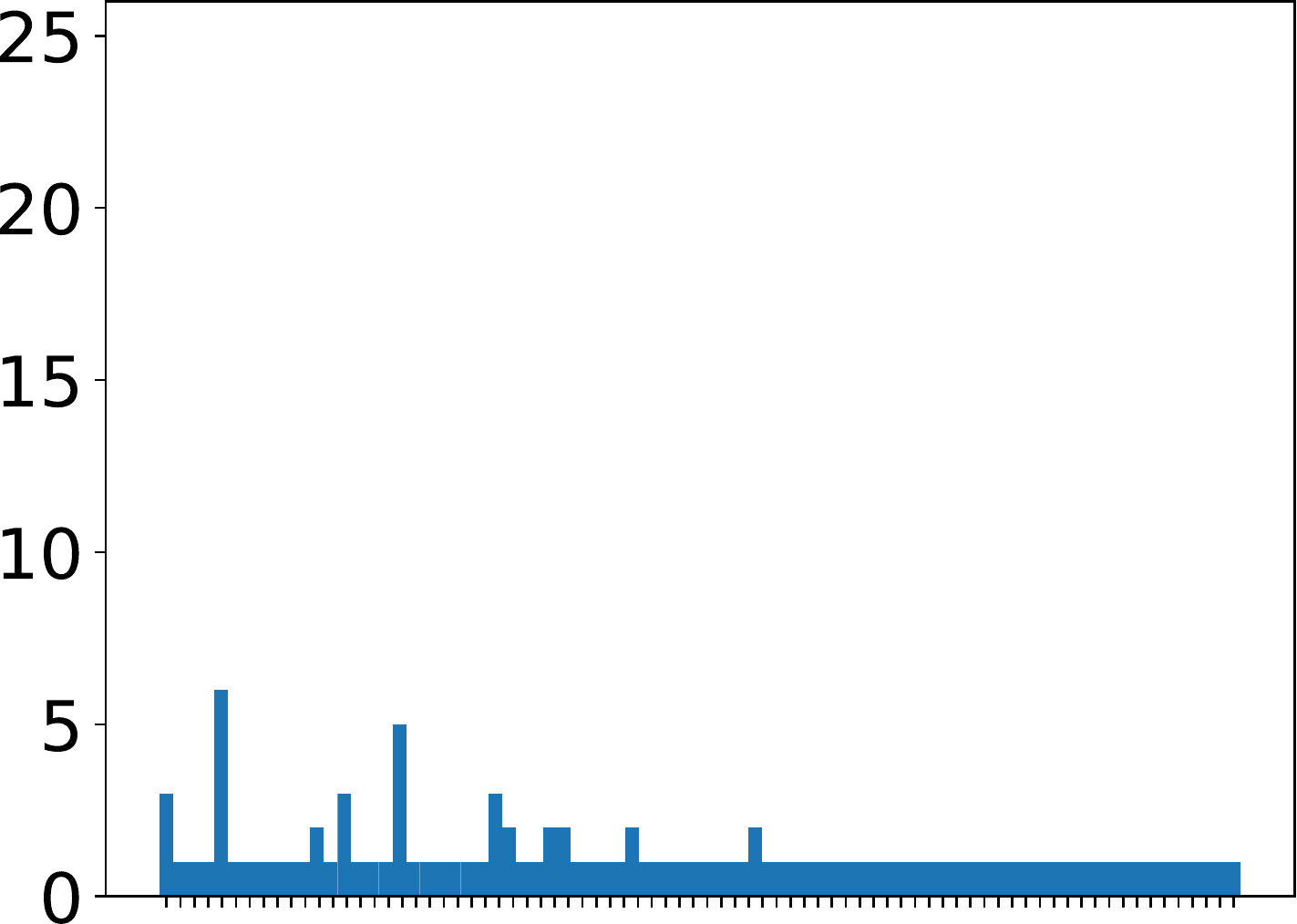}\\ [-1mm]
& \ \ \ \ \small Solutions & \ \ \ \ \small Solutions\\
\end{tabular}
\vspace{-0.3cm}
\caption{Distribution of best found solutions for 100 runs by using the LT (left) and the RT (right) with $n^\text{pop}=4000$ on the second dataset split of Yac.}\label{fig:distribution-solutions}
\end{figure}

Table~\ref{tab:non-unique-solutions} reports the number of best found solutions that have at least one duplicate, i.e. their genotype is semantically equivalent (e.g., $x_1 + x_2 = x_2 + x_1$), along different runs for 5 random splits of Yac (semantic equivalence was determined by automatic tests\footnote{Including the use of symbolic simplification with \url{https://andrewclausen.net/computing/deriv.html}.} followed by manual inspection). It can be seen that the LT finds more duplicate solutions than the RT, by a margin of around 30\% (difference between averages).  
Figure~\ref{fig:distribution-solutions} shows the distribution of solutions found for the second dataset split with $n^\text{pop}=4000$, i.e. where both the LT and the RT found a large number of duplicates. The LT has a marked chance of leading to the discovery of a particular solution, up to one-fourth of the times. When the RT is used, a same solution is found only up to 6 times out of 100.

This confirms the hypothesis that linkage-based variation can propagate salient node patterns more than random variation should such patterns exist, enhancing the likelihood of discovering particular solutions.

\begin{table}
\centering
\caption{Percentage of best solutions with duplicates found by GP-GOMEA with LT and RT for different splits of Yac.}\label{tab:non-unique-solutions}
\scriptsize
\scalebox{1.0}{
\begin{tabular}{cccccc}
\toprule
 & \multicolumn{2}{c}{$n^\text{pop}=2000$} & \ssW & \multicolumn{2}{c}{$n^\text{pop}=4000$} \\
Split & LT & RT & & LT & RT\\
\midrule
1 & 36 & 18 & & 44 & 15\\
2 & 42 & 12 & & 49 & 21 \\
3 & 40 & 7 & & 43 & 8\\
4 & 43 & 8 & & 45 & 25\\
5 & 36 & 16 & & 49 & 16\\
Avg. & 39 & 12 & & 46 & 17\\
\bottomrule
\end{tabular}
}
\end{table}

\section{Ephemeral random constants \& linkage}\label{sec:linkage-erc}
In many GP problems, and in particular in SR, the use of ERCs can be very beneficial~\citep{poli2008field}. An ERC is a terminal which is set to a constant only when instantiated in a solution. In SR, this constant is commonly sampled uniformly at random from a user-defined interval.

Because every node instance of ERC is a different constant, linkage learning needs to deal with a large number of different symbols. This can lead to two shortcomings. First, a very large population size may be needed for salient node patterns to emerge. Second, data structures used to store the frequencies of symbols grow really big and become slow (e.g., hash maps).

We explore three strategies to deal with this:
\begin{inparaenum}
\item \underline{all-const}: Ignore the shortcomings, and consider all different constants as different symbols during linkage learning;
\item \underline{no-const}: Skip all constants during linkage learning, i.e. set their frequency to zero. This approximation is reasonable since all constants are unique at initialization, and the respective frequency is almost zero. However, during evolution some constants will be propagated while others will be discarded, making this approximation less and less accurate over time;
\item \underline{bin-const}: Perform on-line binning. We set a maximum number $\gamma$ of constants to consider. After $\gamma$ different constants have been encountered in frequency counting, any further constant is considered to fall into the same bin as the closest constant among the first $\gamma$. The closest constant can be determined with binary search in $\log_2(\gamma)$ steps. Contrary to strategy no-const, we expect the error of this approximation to lower over time, because selection lowers diversity, meaning that the total number of different constants will be reduced as generations pass.
\end{inparaenum}

\begin{table}
\caption{Median training NMSE and median test NMSE of 30 runs for GP-GOMEA with the LT using the three strategies all-const, no-const, bin-const, and with the RT.}\label{tab:linkage-erc}
\centering
\scriptsize
\scalebox{1.0}{
\setlength{\tabcolsep}{0.1em}
\begin{tabular}{l S c S c S c S c c S c S c S c S c}
\toprule 
 & \multicolumn{8}{c}{Training NMSE} & \ssW\ssW & \multicolumn{8}{c}{Test NMSE} \\
Dataset & \multicolumn{2}{c}{\color{cbScarlet}all-const} & \multicolumn{2}{c}{\color{cbBlue}no-const} & \multicolumn{2}{c}{\color{cbPink}bin-const} & \multicolumn{2}{c}{\color{cbTeal}RT} & & \multicolumn{2}{c}{\color{cbScarlet}all-const} & \multicolumn{2}{c}{\color{cbBlue}no-const} & \multicolumn{2}{c}{\color{cbPink}bin-const} & \multicolumn{2}{c}{\color{cbTeal}RT} 
\\
\midrule

Air 
&\ssW{} 27.7 & \ssW{}\wtC{cbPink}{}\btC{cbTeal}{}
&\ssW{} 28.0 & \ssW{}\wtC{cbPink}{}\btC{cbTeal}{}
&\ssW{} 27.5 & \btC{cbScarlet}{}\btC{cbBlue}{}\btC{cbTeal}{}
&\ssW{} 31.4 & \wtC{cbScarlet}{}\wtC{cbBlue}{}\wtC{cbPink} 
&
&\ssW{} 28.7 & \btC{cbBlue}{}\wtC{cbPink}{}\btC{cbTeal}{}
&\ssW{} 29.6 & \wtC{cbScarlet}{}\wtC{cbPink}{}\btC{cbTeal}{}
&\ssW{} 27.8 & \btC{cbScarlet}{}\btC{cbBlue}{}\btC{cbTeal}{}
&\ssW{} 32.5 & \wtC{cbScarlet}{}\wtC{cbBlue}{}\wtC{cbPink}{} 
\\

Bos 
&\ssW{} 15.2 & \ssW{}\wtC{cbPink}{}\btC{cbTeal}{}
&\ssW{} 15.3 & \ssW{}\ssW{}\btC{cbTeal} 
&\ssW{} 15.0 & \btC{cbScarlet}{}\ssW{}\btC{cbTeal}{}
&\ssW{} 17.6 & \wtC{cbScarlet}{}\wtC{cbBlue}{}\wtC{cbPink}
& 
&\ssW{} 24.2 & \ssW{}\wtC{cbPink}{}\wtC{cbTeal}{}
&\ssW{} 23.2 & \ssW{}\wtC{cbPink}{}\btC{cbTeal} 
&\ssW{} 21.8 & \btC{cbScarlet}{}\btC{cbBlue}{}\btC{cbTeal} 
&\ssW{} 24.2 & \btC{cbScarlet}{}\btC{cbBlue}{}\wtC{cbPink}
\\

Con 
&\ssW{} 17.2 &\btC{cbBlue}{}\wtC{cbPink}{}\btC{cbTeal} 
&\ssW{} 17.2 & \wtC{cbScarlet}{}\wtC{cbPink}{}\btC{cbTeal} 
&\ssW{} 17.0 & \btC{cbScarlet}{}\btC{cbBlue}{}\btC{cbTeal} 
&\ssW{} 18.5 & \wtC{cbScarlet}{}\wtC{cbBlue}{}\wtC{cbPink}{}
& 
&\ssW{} 18.5 &\ssW{}\ssW{}\btC{cbTeal} 
&\ssW{} 18.7 & \ssW{}\ssW{}\btC{cbTeal} 
&\ssW{} 18.8 & \ssW{}\ssW{}\btC{cbTeal} 
&\ssW{} 19.8 & \wtC{cbScarlet}{}\wtC{cbBlue}{}\wtC{cbPink}{}

\\

Dow 
&\ssW{} 21.4 &\ssW{}\wtC{cbPink}{}\btC{cbTeal}
&\ssW{} 21.1 &\ssW{}\ssW{}\btC{cbTeal}
&\ssW{} 20.7 &\btC{cbScarlet}{}\ssW{}\btC{cbTeal}{}
&\ssW{} 24.5 &\wtC{cbScarlet}{}\wtC{cbBlue}{}\wtC{cbPink}{}
& 
&\ssW{} 22.8 &\wtC{cbBlue}{}\ssW{}\btC{cbTeal}{}
&\ssW{} 21.9 &\btC{cbScarlet}{}\btC{cbPink}{}\btC{cbTeal}
&\ssW{} 22.5 &\ssW{}\wtC{cbBlue}{}\btC{cbTeal}{}
&\ssW{} 25.5 &\wtC{cbScarlet}{}\wtC{cbBlue}{}\wtC{cbPink}{}
\\

EnC 
&\ssW{} 5.51 &\btC{cbBlue}{}\btC{cbPink}{}\btC{cbTeal}
&\ssW{} 5.72 &\wtC{cbScarlet}{}\ssW{}\btC{cbTeal}
&\ssW{} 5.76 &\wtC{cbScarlet}{}\ssW{}\btC{cbTeal} 
&\ssW{} 6.44 &\wtC{cbScarlet}{}\wtC{cbBlue}{}\wtC{cbPink}{}
& 
&\ssW{} 6.18 &\btC{cbBlue}{}\ssW{}\btC{cbTeal}
&\ssW{} 6.34 &\wtC{cbScarlet}{}\wtC{cbPink}{}\btC{cbTeal}
&\ssW{} 6.00 &\ssW{}\btC{cbBlue}{}\btC{cbTeal}{}
&\ssW{} 6.77 &\wtC{cbScarlet}{}\wtC{cbBlue}{}\wtC{cbPink}{}
\\

EnH 
&\ssW{} 3.00 &\btC{cbBlue}{}\wtC{cbPink}{}\btC{cbTeal}
&\ssW{} 3.14 &\wtC{cbScarlet}{}\wtC{cbPink}{}\btC{cbTeal}
&\ssW{} 2.80 &\btC{cbScarlet}{}\btC{cbBlue}{}\btC{cbTeal}
&\ssW{} 4.10 &\wtC{cbScarlet}{}\wtC{cbBlue}{}\wtC{cbPink}{}
& 
&\ssW{} 3.28 &\btC{cbBlue}{}\wtC{cbPink}{}\btC{cbTeal}
&\ssW{} 3.33 &\wtC{cbScarlet}{}\wtC{cbPink}{}\btC{cbTeal}
&\ssW{} 3.11 &\btC{cbScarlet}{}\btC{cbBlue}{}\btC{cbTeal}
&\ssW{} 4.67 &\wtC{cbScarlet}{}\wtC{cbBlue}{}\wtC{cbPink}{}
\\

Tow 
&\ssW{} 12.3 &\ssW{}\wtC{cbPink}{}\btC{cbTeal} 
&\ssW{} 12.2 &\ssW{}\ssW{}\btC{cbTeal} 
&\ssW{} 12.3 &\btC{cbScarlet}{}\ssW{}\btC{cbTeal} 
&\ssW{} 13.2 &\wtC{cbScarlet}{}\wtC{cbBlue}{}\wtC{cbPink}{}
& 
&\ssW{} 12.9 &\ssW{}\ssW{}\btC{cbTeal} 
&\ssW{} 12.8 &\ssW{}\ssW{}\btC{cbTeal} 
&\ssW{} 12.8 &\ssW{}\ssW{}\btC{cbTeal} 
&\ssW{} 13.5 &\wtC{cbScarlet}{}\wtC{cbBlue}{}\wtC{cbPink}{}
\\

WiR 
&\ssW{} 60.3 &\ssW{}\btC{cbPink}{}\btC{cbTeal}
&\ssW{} 60.2 &\ssW{}\ssW{}\btC{cbTeal}{}
&\ssW{} 60.2 &\wtC{cbScarlet}{}\ssW{}\btC{cbTeal}{}
&\ssW{} 61.2 &\wtC{cbScarlet}{}\wtC{cbBlue}{}\wtC{cbPink}{}
& 
&\ssW{} 63.6 &\ssW{}\ssW{}\btC{cbTeal}
&\ssW{} 62.9 &\ssW{}\ssW{}\ssW{}
&\ssW{} 62.9 &\ssW{}\ssW{}\btC{cbTeal}{}
&\ssW{} 63.2 &\wtC{cbScarlet}{}\ssW{}\wtC{cbPink}{}
\\

WiW 
&\ssW{} 67.6 &\btC{cbBlue}{}\btC{cbPink}{}\btC{cbTeal} 
&\ssW{} 68.1 &\wtC{cbScarlet}{}\wtC{cbPink}{}\btC{cbTeal} 
&\ssW{} 68.0 &\wtC{cbScarlet}{}\btC{cbBlue}{}\btC{cbTeal}
&\ssW{} 68.5 &\wtC{cbScarlet}{}\wtC{cbBlue}{}\wtC{cbPink}{}
& 
&\ssW{} 68.9 &\ssW{}\ssW{}\btC{cbTeal} 
&\ssW{} 69.0 &\ssW{}\ssW{}\btC{cbTeal} 
&\ssW{} 69.4 &\ssW{}\ssW{}\btC{cbTeal}
&\ssW{} 69.9 &\wtC{cbScarlet}{}\wtC{cbBlue}{}\wtC{cbPink}{}
\\

Yac 
&\ssW{} 0.32 &\btC{cbBlue}{}\btC{cbPink}{}\btC{cbTeal}  
&\ssW{} 0.35 &\wtC{cbScarlet}{}\wtC{cbPink}{}\btC{cbTeal} 
&\ssW{} 0.34 &\wtC{cbScarlet}{}\btC{cbBlue}{}\btC{cbTeal} 
&\ssW{} 0.38 &\wtC{cbScarlet}{}\wtC{cbBlue}{}\wtC{cbPink}{}
& 
&\ssW{} 0.55 &\ssW{}\ssW{}\btC{cbTeal}  
&\ssW{} 0.61 &\ssW{}\wtC{cbPink}{}\btC{cbTeal} 
&\ssW{} 0.52 &\ssW{}\btC{cbBlue}{}\btC{cbTeal} 
&\ssW{} 0.63 &\wtC{cbScarlet}{}\wtC{cbBlue}{}\wtC{cbPink}{}
\\

\bottomrule
\end{tabular}
}
\vspace{-3mm}
\end{table}

\begin{table}
\caption{Median time of 30 runs for GP-GOMEA with the LT using the three strategies all-const, no-const, bin-const, and with the RT.}\label{tab:linkage-erc-time}
\centering
\scriptsize
\scalebox{1.0}{
\setlength{\tabcolsep}{0.1em}
\begin{tabular}{l S c S c S c S c}
\toprule 

& \multicolumn{8}{c}{Time (s)} \\

Dataset & \multicolumn{2}{c}{\color{cbScarlet}all-const} & \multicolumn{2}{c}{\color{cbBlue}no-const} & \multicolumn{2}{c}{\color{cbPink}bin-const} & \multicolumn{2}{c}{\color{cbTeal}RT}
\\
\midrule

Air 
&\ssW{} 355.4 &\wtC{cbBlue}{}\wtC{cbPink}{}\wtC{cbTeal}{}
&\ssW{} 71.4 &\btC{cbScarlet}{}\btC{cbPink}{}\btC{cbTeal}{}
&\ssW{} 80.0 &\btC{cbScarlet}{}\wtC{cbBlue}{}\btC{cbTeal}{}
&\ssW{} 80.1 &\btC{cbScarlet}{}\wtC{cbBlue}{}\wtC{cbPink}{}
\\

Bos 
&\ssW{} 63.4 & \wtC{cbBlue}{}\wtC{cbPink}{}\wtC{cbTeal}{}
&\ssW{} 29.4 & \btC{cbScarlet}{}\btC{cbPink}{}\wtC{cbTeal}{}
&\ssW{} 30.9 & \btC{cbScarlet}{}\wtC{cbBlue}{}\wtC{cbTeal}{}
&\ssW{} 24.5 & \btC{cbScarlet}{}\btC{cbBlue}{}\btC{cbPink} 
\\

Con 
&\ssW{} 154.9 & \wtC{cbBlue}{}\wtC{cbPink}{}\wtC{cbTeal}{}
&\ssW{} 56.7 &\btC{cbScarlet}{}\btC{cbPink}{}\ssW{} 
&\ssW{} 59.8 & \btC{cbScarlet}{}\wtC{cbBlue}{}\ssW{}
&\ssW{} 58.4 & \btC{cbScarlet}{}\ssW{}\ssW{} 
\\

Dow 
&\ssW{} 53.8 &\wtC{cbBlue}{}\btC{cbPink}{}\wtC{cbTeal}{}
&\ssW{} 51.7 &\ssW{}{}\btC{cbPink}{}\wtC{cbTeal}{}
&\ssW{} 54.9 &\wtC{cbScarlet}{}\wtC{cbBlue}{}\wtC{cbTeal}{}
&\ssW{} 37.7 &\btC{cbScarlet}{}\btC{cbBlue}{}\btC{cbPink}{} 
\\

EnC 
&\ssW{} 147.2 &\wtC{cbBlue}{}\wtC{cbPink}{}\wtC{cbTeal}{} 
&\ssW{} 40.5 &\btC{cbScarlet}{}\btC{cbPink}{}\btC{cbTeal}
&\ssW{} 43.5 &\btC{cbScarlet}{}\wtC{cbBlue}{}\btC{cbTeal}
&\ssW{} 45.6 &\btC{cbScarlet}{}\wtC{cbBlue}{}\wtC{cbPink}{}  
\\

EnH 
&\ssW{} 145.0 &\wtC{cbBlue}{}\wtC{cbPink}{}\wtC{cbTeal}
&\ssW{} 45.8 &\btC{cbScarlet}{}\btC{cbPink}{}\ssW{} 
&\ssW{} 49.4 &\btC{cbScarlet}{}\wtC{cbBlue}{}\wtC{cbTeal}
&\ssW{} 45.7 &\btC{cbScarlet}{}\ssW{}\btC{cbPink}
\\

Tow 
&\ssW{} 255.9 &\wtC{cbBlue}{}\wtC{cbPink}{}\wtC{cbTeal}{} 
&\ssW{} 246.6 &\btC{cbScarlet}{}\ssW{}\wtC{cbTeal}
&\ssW{} 245.6 &\btC{cbScarlet}{}\ssW{}\wtC{cbTeal}{}
&\ssW{} 233.9 &\btC{cbScarlet}{}\btC{cbBlue}{}\btC{cbPink}{}
\\

WiR 
&\ssW{} 126.1 & \wtC{cbBlue}{}\wtC{cbPink}{}\wtC{cbTeal}
&\ssW{} 67.7 & \btC{cbScarlet}{}\btC{cbPink}{}\ssW{} 
&\ssW{} 80.2 & \btC{cbScarlet}{}\wtC{cbBlue}{}\wtC{cbTeal}
&\ssW{} 70.1 & \btC{cbScarlet}{}\ssW{}\btC{cbPink}{}  
\\

WiW 
&\ssW{} 285.0 &\wtC{cbBlue}{}\wtC{cbPink}{}\wtC{cbTeal} 
&\ssW{} 213.3 &\btC{cbScarlet}{}\btC{cbPink}{}\btC{cbTeal} 
&\ssW{} 237.2 &\btC{cbScarlet}{}\wtC{cbBlue}{}\wtC{cbTeal}
&\ssW{} 224.1 &\btC{cbScarlet}{}\wtC{cbBlue}{}\btC{cbPink}{}
\\

Yac 
&\ssW{} 236.5 &\wtC{cbBlue}{}\wtC{cbPink}{}\wtC{cbTeal} 
&\ssW{} 23.9 &\btC{cbScarlet}{}\btC{cbPink}{}\wtC{cbTeal} 
&\ssW{} 24.8 &\btC{cbScarlet}{}\wtC{cbBlue}{}\wtC{cbTeal} 
&\ssW{} 22.8 &\btC{cbScarlet}{}\btC{cbBlue}{}\btC{cbPink}{}
\\

\bottomrule
\end{tabular}
}
\vspace{-3mm}
\end{table}

\subsection{Experiment: linkage learning with ERCs}\label{sec:exp-ll-with-erc}
We use the same parameter setup of the experiment in Sec.~\ref{sec:exp-normalized-mi}, this time adding an ERC terminal to the terminal set. We compare the three strategies to handle ERCs when learning the LT. For this experiment and in the rest of the article, we use $\gamma = 100$ in bin-const. We observed that for problems with a small number of features (e.g., Air and Yac), i.e., where ERC sampling is more likely and thus more constants are produced, this choice reduces the number of constant symbols to be considered by linkage learning in the first generations by a factor of $\sim 50$.
We also report the results obtained with the RT as a baseline, under the hypothesis that using ERCs compromises linkage learning to the point that random variation becomes equally good or better. 

The results of this experiment are shown in Table~\ref{tab:linkage-erc} (training and test NMSE) and Table~\ref{tab:linkage-erc-time} (running time). The Friedman test reveals significant differences among the configurations for train, test, and time performance. Note that the use of ERCs leads to lower errors compared to not using them (compare with Table~\ref{tab:exp-1-results}).

In terms of training error, the RT is always outperformed by the use of the LT, no matter the strategy. The all-const strategy is significantly better than no-const in half of the problems, and never worse. Overall, bin-const performs best, with 6 out of 10 significantly better results than all-const. The fact that all-const can be outperformed by bin-const supports the hypothesis that linkage learning can be compromised by the presence of too many constants to consider, which hide the true salient patterns. 
Test results are overall similar to the training ones, but less comparisons are significant.

In terms of time, all-const is almost always significantly worse than the other methods, and often by a consistent margin. This is particularly marked for problems with a small number of features (i.e., Air, Yac). There, more random constants are present in the initial population, since the probability of sampling the ERC from the terminal set is inversely proportional to the number of features. 

Interestingly, despite the lack of a linkage-learning overhead, using the RT is not always the fastest option. This is because random variation leads to a slower convergence of the population compared to the linkage-based one, where salient patterns are quickly propagated, and less variation attempts result in changes of the genotype that require a fitness evaluation (see Sec.~\ref{sec:gom}). The slower convergence caused by the RT can also be seen in Figure~\ref{fig:exp1-yacht} (for the previous experiment), and was also observed in other work, in terms of diversity preservation~\citep{medvet2018unveiling}. 

Between the LT-based strategies, the fastest is no-const, at the cost of a bigger training error. Although consistently slower than no-const, bin-const is still quite fast, and achieves the lowest training errors. We found bin-const to be preferable in test NMSE as well. In the following, we always use bin-const, with $\gamma=100$.

\section{Interleaved Multistart Scheme}\label{sec:ims}
The Interleaved Multistart Scheme (IMS) is a wrapper for evolutionary runs largely inspired by the work of~\citep{harik1999parameterless} on genetic algorithms. It works by interleaving the execution of several runs of increasing resources (e.g., population size). The main motivation for using the IMS is to make an EA much more robust to parameter settings, and alleviate the need for practitioners to tinker with parameters. In fact, the whole design of GP-GOMEA attempts to promote the aspects of ease-of-use and robustness: the EA has no need for parameters that specify how to conduct variation (e.g., crossover or mutation rates), nor how to conduct selection (e.g., tournament size).
The IMS or similar schemes are often used with MBEAs~\citep{lin2018investigation,goldman2014parameter}, where population size plays a crucial role in determining the quality of model building. 
Note that although the IMS has potential to be parallelized, here it is used in a sequential manner.

An IMS for GP-GOMEA was first proposed in~\citep{virgolin2017scalable}, and its outline is as follows. A collection of parameter settings $\sigma_\text{base}$ is given as input, which will be used in the first run $R_1$. The IMS runs until a termination criterion is met (e.g., number of generations, time budget). The run $R_i$ performs one generation if no run that precedes it exists (e.g., because it is the first run or because all previous runs have been terminated), or if the previous run $R_{i-1}$ has executed $g$ generations. The first time $R_i$ is about to execute a generation, it is initialized using the parameter settings $\sigma_\text{base}$ scaled by the index $i$. For example, the population size can be set to $2^{i-1} n^\text{pop}_\text{base}$ (i.e., doubling the population size of the previous run). Finally, when a run completes a generation, a check is done to determine if the run should be terminated (explained below).

\subsection{An IMS for supervised learning tasks}
The first implementation of the IMS for GP-GOMEA was designed to deal with GP benchmark problems of pure optimization. That implementation therefore scaled both the population size and the height of trees in an attempt to find the optimal solution (of unknown size)~\citep{virgolin2017scalable}.

In this work, we use the IMS as follows.
\begin{inparaenum}
\item \emph{Scaling of parameter settings}: We scale only the population size. For run $R_i$, the population size is set to $n^\text{pop}_i = 2^{i-1}n^\text{pop}_\text{base}$.
\item \emph{Run termination}: A run $R_i$ is terminated if the fitness of its best solution is worse than the one of a run $R_j$ initialized later, i.e., with $j>i$, or if it converges to all identical solutions.
\end{inparaenum}

Differently from \citep{virgolin2017scalable} we no longer scale the tree height $h$ because in SR, and in supervised learning tasks in general, no optimum is known beforehand, and it is rather desired to find a solution that generalizes well to unseen examples. Moreover, $h$ bounds the maximum solution size, which influences interpretability. Hence $h$ is left as a parameter for the user to set, and we recommend $h \leq 4$ to increase the chance that solutions will be interpretable (see Sec.~\ref{sec:discussion}).

We set the run termination criteria to be based upon the fitness of best solutions instead of mean population fitness as done by~\citep{harik1999parameterless} and~\citep{virgolin2017scalable}, because in SR it can happen that the error of a few solutions becomes so large that it compromises the mean population fitness. This can trigger the termination criteria even if solutions exist that are competitive with the ones of other runs. Also differently from the other versions of the IMS, when terminating a run, we do not automatically terminate all previous runs. Indeed, some runs with smaller parameter settings may still be very competitive (e.g., due to the fortunate sampling of particular constants when using ERCs).

We lastly propose to exploit the fact that many runs are performed within the IMS to tackle a central problem of learning tasks: generalization. Instead of discarding the best solutions of terminating runs, we store them in an archive. When the IMS terminates, we re-compute the fitness of each solution in the archive using a set of examples different from the training set, i.e. the validation set, and return the new best performing, i.e., the solution that generalized best. The final test performance is measured on a third, separate set of examples (test set).

\section{Benchmarking GP-GOMEA}\label{sec:benchmarking}
We compare GP-GOMEA (using the new LT) with tree-based GP with traditional subtree crossover and subtree mutation (GP-Trad), tree-based GP using the state-of-the-art, semantic-aware operator Random Desired Operator (GP-RDO)~\citep{pawlak2015semantic}, and Decision Tree for Regression (DTR)~\citep{breiman1984classification}. 

We consider RDO because, as mentioned in the introduction, semantic-aware operators have been studied with interest in the last years. Several works either built upon RDO, or used RDO as a baseline for comparison (see, e.g.,~\citep{chen2018improving,pawlak2018competent,virgolin2019linear}). Yet, consistently large solutions were found. It is interesting to assess how RDO fares when rather strict solution size limits are enforced.
Because of such limits, we remark we cannot consider another popular set of semantic-aware operators, i.e., the operators used by Geometric Semantic Genetic Programming (GSGP)~\citep{moraglio2012geometric}. These operators work by stacking entire solutions together, necessarily causing extremely large solution growth (even if smart simplifications are attempted~\citep{martins2018solving}).

We consider DTR because it is considered among the state-of-the-art algorithms to learn interpretable models~\citep{doshi2017towards,guidotti2018survey}. We remark that DTR ensembles (e.g.,~\citep{breiman2001random,chen2016xgboost}) are typically markedly more accurate than single DTRs, but are considered not interpretable.

\subsection{Experimental setup}
For the EAs, we use a fixed time limit of $1,000$ seconds\footnote{Experiments were run on an Intel\textsuperscript{\textregistered} Xeon\textsuperscript{\textregistered} Processor E5-2650 v2.}. We choose a time-based comparison because GP-GOMEA performs more evaluations per generation than other GP algorithms (up to $2\ell - 2$ evaluations per generation with the LT), and so that the overhead of learning the LT (which does not involve evaluations) is taken into account. 

We consider maximum solution sizes $\ell = 15, 31, 63$ (tree nodes), i.e. corresponding to $h=3,4,5$ respectively, for full $r$-ary trees. The EAs are run with a typical fixed population size $n^\text{pop}=1000$ and also with the IMS, considering three values for the number of generations in between runs $g$: 4, 6, and 8. For the fixed population size, if the population of GP-GOMEA converges before the time limit, since there is no mutation, it is randomly re-started. Choices of $g$ between 4 and 8 are standards from literature~\citep{bouter2017exploiting,
virgolin2017scalable}. 

Our implementation of GP-Trad and GP-RDO mostly follows the one of~\citep{pawlak2015semantic}. The population is initialized with the \emph{Ramped Half-and-Half} method, with tree height between 2 and $h$. Selection is performed with tournament of size 7. GP-Trad uses a rate of $0.9$ for subtree crossover, and of $0.1$ for subtree mutation.
GP-RDO uses the population-based library of subtrees, a rate of $0.9$ for RDO, and of $0.1$ for subtree mutation. Subtree roots to be variated are chosen with the \emph{uniform depth mutation} method, which makes nodes of all depths equally likely to be selected~\citep{pawlak2015semantic}. Elitism is ensured by cloning the best solution into the next generation. All EAs are implemented in \CC{}, and the code is available at: \url{https://goo.gl/15tMV7}.

For GP-Trad we consider two versions, to account for different types of solution size limitation. In the first version, called GP-Trad\textsuperscript{$h$}, we force trees to be constrained within a maximum height ($h=3,4$), as done for GP-GOMEA. This way, we can see which algorithm searches better in the same representation space. In the second version, GP-Trad\textsuperscript{$\ell$}, we allow more freedom in tree shape, by only bounding the number of tree nodes. This limit is set to the maximum number of nodes obtainable in a full $r$-ary tree of height $h$ ($\ell=15$ for $h=3$, $\ell=31$ for $h=4$). 
As indicated by previous literature \citep{gathercole1996adverse,langdon1997analysis}, and as will be shown later in the results, GP-Trad\textsuperscript{$\ell$} outperforms GP-Trad\textsuperscript{$h$}. We found that the same holds also for GP-RDO, and present here only its best configuration, i.e., a version where the number of tree nodes is limited like for GP-Trad\textsuperscript{$\ell$}.

We use the Python Scikit-learn implementation of DTR~\citep{pedregosa2011scikit}, with 5-fold cross-validation grid-search over the training set to tune the following hyper-parameters: \emph{splitter} $\in$ \{`\emph{best}',`\emph{random}'\}; \emph{max\_features} $\in$ \{$\frac{1}{2}, \frac{3}{4}, 1$\}; \emph{max\_depth} $\in$ \{3,4,5,6\} (documentation available at \url{http://goo.gl/hbyFq2}). We do not allow larger depth values because, like for GP solutions, excessively large decision trees are uninterpretable. The best generalizing model found by cross-validation is then used on the test set.

\begin{table*}
\caption{Median validation and test NMSE of 30 runs with $\ell = 15$ for GP-GOMEA (G), GP-Trad\textsuperscript{$h$} (T\textsuperscript{$\ell$}), GP-Trad\textsuperscript{$h$} (T\textsuperscript{$\ell$}), GP-RDO (R) with $n^\text{pop}=1000$ and IMS with $g \in \{ 4, 6, 8 \}$, and DTR. Significance is assessed within each population scheme w.r.t. GP-GOMEA. The last row reports the number of times the EA performs significantly better (B) and worse (W) than GP-GOMEA.}\label{tab:validation3}
\centering
\setlength{\tabcolsep}{0.001em}
\tiny
\scalebox{1.0}{
\begin{tabular}{lScScScSccScScScSccScScScSccScScScScSc}
\toprule
\multicolumn{38}{c}{\small \textbf{Validation $\ell = 15$}} \\ 
 & \multicolumn{8}{c}{$n^\text{pop}=1000$} & \ssW{}\ssW{} & \multicolumn{8}{c}{IMS $g=4$} & \ssW{}\ssW{} & \multicolumn{8}{c}{IMS $g=6$} & \ssW{}\ssW{} & \multicolumn{9}{c}{IMS $g=8$} \\
 \ssW{}& 
\multicolumn{2}{c}{\color{cbScarlet}G} & \multicolumn{2}{c}{\color{black}T\textsuperscript{$h$}} & \multicolumn{2}{c}{\color{black}T\textsuperscript{$\ell$}} & \multicolumn{2}{c}{\color{black}R} & & 
\multicolumn{2}{c}{\color{cbScarlet}G} & \multicolumn{2}{c}{\color{black}T\textsuperscript{$h$}} & \multicolumn{2}{c}{\color{black}T\textsuperscript{$\ell$}} & \multicolumn{2}{c}{\color{black}R} & & 
\multicolumn{2}{c}{\color{cbScarlet}G} & \multicolumn{2}{c}{\color{black}T\textsuperscript{$h$}} & \multicolumn{2}{c}{\color{black}T\textsuperscript{$\ell$}} & \multicolumn{2}{c}{\color{black}R} & & 
\multicolumn{2}{c}{\color{cbScarlet}G} & \multicolumn{2}{c}{\color{black}T\textsuperscript{$h$}} & \multicolumn{2}{c}{\color{black}T\textsuperscript{$\ell$}} & \multicolumn{2}{c}{\color{black}R} & \multicolumn{2}{c}{D}\\
\midrule

Air &
\multicolumn{2}{c}{\ssW{}39.2\ssW{}}&
\ssW{}40.6&\wtC{cbScarlet}&
\ssW{}35.0&\btC{cbScarlet}&
\ssW{}44.0&\wtC{cbScarlet}&
&
\multicolumn{2}{c}{\ssW{}34.7\ssW{}}&
\ssW{}38.3&\wtC{cbScarlet}&
\ssW{}31.4&\btC{cbScarlet}&
\ssW{}42.5&\wtC{cbScarlet}&
&
\multicolumn{2}{c}{\ssW{}34.9\ssW{}}&
\ssW{}39.7&\wtC{cbScarlet}&
\ssW{}33.6&\btC{cbScarlet}&
\ssW{}42.0&\wtC{cbScarlet}&
&
\multicolumn{2}{c}{\ssW{}34.4\ssW{}}&
\ssW{}39.4&\wtC{cbScarlet}&
\ssW{}32.0&\btC{cbScarlet}&
\ssW{}42.3&\wtC{cbScarlet}&
\ssW{}31.1&\btC{cbScarlet}
\\ 
Bos &
\multicolumn{2}{c}{\ssW{}21.1\ssW{}}&
\ssW{}23.4&\ssW{}&
\ssW{}25.3&\wtC{cbScarlet}&
\ssW{}25.7&\wtC{cbScarlet}&
&
\multicolumn{2}{c}{\ssW{}18.2\ssW{}}&
\ssW{}21.2&\wtC{cbScarlet}&
\ssW{}19.0&\wtC{cbScarlet}&
\ssW{}20.8&\wtC{cbScarlet}&
&
\multicolumn{2}{c}{\ssW{}19.2\ssW{}}&
\ssW{}21.2&\wtC{cbScarlet}&
\ssW{}20.4&\ssW{}&
\ssW{}20.9&\wtC{cbScarlet}&
&
\multicolumn{2}{c}{\ssW{}19.4\ssW{}}&
\ssW{}21.6&\wtC{cbScarlet}&
\ssW{}19.9&\wtC{cbScarlet}&
\ssW{}22.5&\wtC{cbScarlet}&
\ssW{}22.9&\wtC{cbScarlet}
\\ 
Con &
\multicolumn{2}{c}{\ssW{}23.2\ssW{}}&
\ssW{}25.3&\wtC{cbScarlet}&
\ssW{}23.4&\ssW{}&
\ssW{}27.0&\wtC{cbScarlet}&
&
\multicolumn{2}{c}{\ssW{}20.3\ssW{}}&
\ssW{}23.1&\wtC{cbScarlet}&
\ssW{}19.4&\btC{cbScarlet}&
\ssW{}26.4&\wtC{cbScarlet}&
&
\multicolumn{2}{c}{\ssW{}20.2\ssW{}}&
\ssW{}23.3&\wtC{cbScarlet}&
\ssW{}19.9&\btC{cbScarlet}&
\ssW{}26.2&\wtC{cbScarlet}&
&
\multicolumn{2}{c}{\ssW{}19.4\ssW{}}&
\ssW{}23.2&\wtC{cbScarlet}&
\ssW{}19.3&\ssW{}&
\ssW{}26.9&\wtC{cbScarlet}&
\ssW{}22.7&\wtC{cbScarlet}
\\ 
Dow &
\multicolumn{2}{c}{\ssW{}26.7\ssW{}}&
\ssW{}28.5&\wtC{cbScarlet}&
\ssW{}27.5&\ssW{}&
\ssW{}30.6&\wtC{cbScarlet}&
&
\multicolumn{2}{c}{\ssW{}24.2\ssW{}}&
\ssW{}26.8&\wtC{cbScarlet}&
\ssW{}24.2&\ssW{}&
\ssW{}32.3&\wtC{cbScarlet}&
&
\multicolumn{2}{c}{\ssW{}24.6\ssW{}}&
\ssW{}26.4&\wtC{cbScarlet}&
\ssW{}24.8&\wtC{cbScarlet}&
\ssW{}31.0&\wtC{cbScarlet}&
&
\multicolumn{2}{c}{\ssW{}24.5\ssW{}}&
\ssW{}26.3&\wtC{cbScarlet}&
\ssW{}25.2&\wtC{cbScarlet}&
\ssW{}31.0&\wtC{cbScarlet}&
\ssW{}30.6&\wtC{cbScarlet}
\\ 
EnC &
\multicolumn{2}{c}{\ssW{}8.72\ssW{}}&
\ssW{}10.6&\wtC{cbScarlet}&
\ssW{}7.34&\ssW{}&
\ssW{}11.0&\wtC{cbScarlet}&
&
\multicolumn{2}{c}{\ssW{}5.86\ssW{}}&
\ssW{}10.2&\wtC{cbScarlet}&
\ssW{}6.49&\wtC{cbScarlet}&
\ssW{}10.7&\wtC{cbScarlet}&
&
\multicolumn{2}{c}{\ssW{}6.01\ssW{}}&
\ssW{}10.3&\wtC{cbScarlet}&
\ssW{}6.24&\ssW{}&
\ssW{}10.5&\wtC{cbScarlet}&
&
\multicolumn{2}{c}{\ssW{}5.87\ssW{}}&
\ssW{}10.2&\wtC{cbScarlet}&
\ssW{}6.10&\wtC{cbScarlet}&
\ssW{}10.8&\wtC{cbScarlet}&
\ssW{}4.23&\btC{cbScarlet}
\\ 
EnH &
\multicolumn{2}{c}{\ssW{}4.95\ssW{}}&
\ssW{}7.45&\wtC{cbScarlet}&
\ssW{}3.83&\btC{cbScarlet}&
\ssW{}7.65&\wtC{cbScarlet}&
&
\multicolumn{2}{c}{\ssW{}3.33\ssW{}}&
\ssW{}7.19&\wtC{cbScarlet}&
\ssW{}3.74&\wtC{cbScarlet}&
\ssW{}7.34&\wtC{cbScarlet}&
&
\multicolumn{2}{c}{\ssW{}3.28\ssW{}}&
\ssW{}7.30&\wtC{cbScarlet}&
\ssW{}3.76&\wtC{cbScarlet}&
\ssW{}7.42&\wtC{cbScarlet}&
&
\multicolumn{2}{c}{\ssW{}3.23\ssW{}}&
\ssW{}7.24&\wtC{cbScarlet}&
\ssW{}3.72&\wtC{cbScarlet}&
\ssW{}7.54&\wtC{cbScarlet}&
\ssW{}0.43&\btC{cbScarlet}
\\ 
Tow &
\multicolumn{2}{c}{\ssW{}12.9\ssW{}}&
\ssW{}14.4&\wtC{cbScarlet}&
\ssW{}13.9&\wtC{cbScarlet}&
\ssW{}20.1&\wtC{cbScarlet}&
&
\multicolumn{2}{c}{\ssW{}12.8\ssW{}}&
\ssW{}13.6&\wtC{cbScarlet}&
\ssW{}13.6&\wtC{cbScarlet}&
\ssW{}20.5&\wtC{cbScarlet}&
&
\multicolumn{2}{c}{\ssW{}12.7\ssW{}}&
\ssW{}14.0&\wtC{cbScarlet}&
\ssW{}13.5&\wtC{cbScarlet}&
\ssW{}20.4&\wtC{cbScarlet}&
&
\multicolumn{2}{c}{\ssW{}13.0\ssW{}}&
\ssW{}14.0&\wtC{cbScarlet}&
\ssW{}13.4&\wtC{cbScarlet}&
\ssW{}20.1&\wtC{cbScarlet}&
\ssW{}11.2&\btC{cbScarlet}
\\ 
WiR &
\multicolumn{2}{c}{\ssW{}65.3\ssW{}}&
\ssW{}64.8&\ssW{}&
\ssW{}64.9&\ssW{}&
\ssW{}66.5&\wtC{cbScarlet}&
&
\multicolumn{2}{c}{\ssW{}63.9\ssW{}}&
\ssW{}64.7&\wtC{cbScarlet}&
\ssW{}64.4&\ssW{}&
\ssW{}65.1&\wtC{cbScarlet}&
&
\multicolumn{2}{c}{\ssW{}63.6\ssW{}}&
\ssW{}63.9&\wtC{cbScarlet}&
\ssW{}64.4&\wtC{cbScarlet}&
\ssW{}64.9&\wtC{cbScarlet}&
&
\multicolumn{2}{c}{\ssW{}63.9\ssW{}}&
\ssW{}63.9&\wtC{cbScarlet}&
\ssW{}64.2&\ssW{}&
\ssW{}65.7&\wtC{cbScarlet}&
\ssW{}71.7&\wtC{cbScarlet}
\\ 
WiW &
\multicolumn{2}{c}{\ssW{}71.4\ssW{}}&
\ssW{}71.3&\ssW{}&
\ssW{}70.9&\ssW{}&
\ssW{}72.6&\wtC{cbScarlet}&
&
\multicolumn{2}{c}{\ssW{}70.8\ssW{}}&
\ssW{}71.2&\wtC{cbScarlet}&
\ssW{}70.7&\wtC{cbScarlet}&
\ssW{}72.3&\wtC{cbScarlet}&
&
\multicolumn{2}{c}{\ssW{}70.7\ssW{}}&
\ssW{}71.5&\wtC{cbScarlet}&
\ssW{}70.8&\wtC{cbScarlet}&
\ssW{}72.6&\wtC{cbScarlet}&
&
\multicolumn{2}{c}{\ssW{}71.2\ssW{}}&
\ssW{}71.4&\wtC{cbScarlet}&
\ssW{}71.2&\wtC{cbScarlet}&
\ssW{}72.6&\wtC{cbScarlet}&
\ssW{}72.2&\wtC{cbScarlet}
\\ 
Yac &
\multicolumn{2}{c}{\ssW{}1.25\ssW{}}&
\ssW{}1.22&\ssW{}&
\ssW{}0.70&\btC{cbScarlet}&
\ssW{}0.96&\btC{cbScarlet}&
&
\multicolumn{2}{c}{\ssW{}0.89\ssW{}}&
\ssW{}1.04&\wtC{cbScarlet}&
\ssW{}0.61&\btC{cbScarlet}&
\ssW{}0.67&\btC{cbScarlet}&
&
\multicolumn{2}{c}{\ssW{}0.92\ssW{}}&
\ssW{}1.01&\wtC{cbScarlet}&
\ssW{}0.61&\btC{cbScarlet}&
\ssW{}0.73&\btC{cbScarlet}&
&
\multicolumn{2}{c}{\ssW{}0.95\ssW{}}&
\ssW{}1.03&\wtC{cbScarlet}&
\ssW{}0.62&\btC{cbScarlet}&
\ssW{}0.76&\btC{cbScarlet}&
\ssW{}0.88&\btC{cbScarlet}
\\ 
\midrule 
B/W &
\multicolumn{2}{c}{---}&
\multicolumn{2}{c}{0/6}&
\multicolumn{2}{c}{3/2}&
\multicolumn{2}{c}{1/9}&
&
\multicolumn{2}{c}{---}&
\multicolumn{2}{c}{0/10}&
\multicolumn{2}{c}{3/5}&
\multicolumn{2}{c}{1/9}&
&
\multicolumn{2}{c}{---}&
\multicolumn{2}{c}{0/10}&
\multicolumn{2}{c}{3/5}&
\multicolumn{2}{c}{1/9}&
&
\multicolumn{2}{c}{---}&
\multicolumn{2}{c}{0/10}&
\multicolumn{2}{c}{2/6}&
\multicolumn{2}{c}{1/9}&
\multicolumn{2}{c}{5/5}\\

\midrule 

\multicolumn{38}{c}{\small \textbf{Test $\ell = 15$}} \\ 
 & \multicolumn{8}{c}{$n^\text{pop}=1000$} & \ssW{}\ssW{} & \multicolumn{8}{c}{IMS $g=4$} & \ssW{}\ssW{} & \multicolumn{8}{c}{IMS $g=6$} & \ssW{}\ssW{} & \multicolumn{9}{c}{IMS $g=8$} \\
 \ssW{}& 
\multicolumn{2}{c}{\color{cbScarlet}G} & \multicolumn{2}{c}{\color{black}T\textsuperscript{$h$}} & \multicolumn{2}{c}{\color{black}T\textsuperscript{$\ell$}} & \multicolumn{2}{c}{\color{black}R} & & 
\multicolumn{2}{c}{\color{cbScarlet}G} & \multicolumn{2}{c}{\color{black}T\textsuperscript{$h$}} & \multicolumn{2}{c}{\color{black}T\textsuperscript{$\ell$}} & \multicolumn{2}{c}{\color{black}R} & & 
\multicolumn{2}{c}{\color{cbScarlet}G} & \multicolumn{2}{c}{\color{black}T\textsuperscript{$h$}} & \multicolumn{2}{c}{\color{black}T\textsuperscript{$\ell$}} & \multicolumn{2}{c}{\color{black}R} & & 
\multicolumn{2}{c}{\color{cbScarlet}G} & \multicolumn{2}{c}{\color{black}T\textsuperscript{$h$}} & \multicolumn{2}{c}{\color{black}T\textsuperscript{$\ell$}} & \multicolumn{2}{c}{\color{black}R} & \multicolumn{2}{c}{D}\\
\midrule

Air &
\multicolumn{2}{c}{\ssW{}38.5\ssW{}}&
\ssW{}40.7&\wtC{cbScarlet}&
\ssW{}35.3&\btC{cbScarlet}&
\ssW{}44.1&\wtC{cbScarlet}&
&
\multicolumn{2}{c}{\ssW{}35.8\ssW{}}&
\ssW{}39.5&\wtC{cbScarlet}&
\ssW{}32.5&\btC{cbScarlet}&
\ssW{}43.3&\wtC{cbScarlet}&
&
\multicolumn{2}{c}{\ssW{}35.2\ssW{}}&
\ssW{}39.3&\wtC{cbScarlet}&
\ssW{}33.2&\btC{cbScarlet}&
\ssW{}42.4&\wtC{cbScarlet}&
&
\multicolumn{2}{c}{\ssW{}35.4\ssW{}}&
\ssW{}39.7&\wtC{cbScarlet}&
\ssW{}32.8&\btC{cbScarlet}&
\ssW{}42.8&\wtC{cbScarlet}&
\ssW{}30.8&\btC{cbScarlet}
\\ 
Bos &
\multicolumn{2}{c}{\ssW{}22.7\ssW{}}&
\ssW{}23.3&\wtC{cbScarlet}&
\ssW{}24.3&\wtC{cbScarlet}&
\ssW{}26.7&\wtC{cbScarlet}&
&
\multicolumn{2}{c}{\ssW{}22.5\ssW{}}&
\ssW{}23.1&\ssW{}&
\ssW{}23.3&\wtC{cbScarlet}&
\ssW{}22.9&\wtC{cbScarlet}&
&
\multicolumn{2}{c}{\ssW{}21.7\ssW{}}&
\ssW{}23.6&\wtC{cbScarlet}&
\ssW{}22.6&\wtC{cbScarlet}&
\ssW{}25.0&\wtC{cbScarlet}&
&
\multicolumn{2}{c}{\ssW{}22.1\ssW{}}&
\ssW{}22.5&\wtC{cbScarlet}&
\ssW{}23.6&\wtC{cbScarlet}&
\ssW{}23.3&\wtC{cbScarlet}&
\ssW{}26.1&\wtC{cbScarlet}
\\ 
Con &
\multicolumn{2}{c}{\ssW{}23.1\ssW{}}&
\ssW{}26.1&\wtC{cbScarlet}&
\ssW{}23.9&\ssW{}&
\ssW{}27.0&\wtC{cbScarlet}&
&
\multicolumn{2}{c}{\ssW{}20.8\ssW{}}&
\ssW{}23.9&\wtC{cbScarlet}&
\ssW{}19.3&\btC{cbScarlet}&
\ssW{}27.7&\wtC{cbScarlet}&
&
\multicolumn{2}{c}{\ssW{}21.2\ssW{}}&
\ssW{}23.9&\wtC{cbScarlet}&
\ssW{}19.9&\btC{cbScarlet}&
\ssW{}26.4&\wtC{cbScarlet}&
&
\multicolumn{2}{c}{\ssW{}20.4\ssW{}}&
\ssW{}24.3&\wtC{cbScarlet}&
\ssW{}19.3&\btC{cbScarlet}&
\ssW{}27.8&\wtC{cbScarlet}&
\ssW{}21.3&\wtC{cbScarlet}
\\ 
Dow &
\multicolumn{2}{c}{\ssW{}26.3\ssW{}}&
\ssW{}27.5&\wtC{cbScarlet}&
\ssW{}26.4&\ssW{}&
\ssW{}31.0&\wtC{cbScarlet}&
&
\multicolumn{2}{c}{\ssW{}24.8\ssW{}}&
\ssW{}26.1&\wtC{cbScarlet}&
\ssW{}24.7&\btC{cbScarlet}&
\ssW{}30.7&\wtC{cbScarlet}&
&
\multicolumn{2}{c}{\ssW{}24.5\ssW{}}&
\ssW{}26.6&\wtC{cbScarlet}&
\ssW{}24.5&\ssW{}&
\ssW{}30.1&\wtC{cbScarlet}&
&
\multicolumn{2}{c}{\ssW{}24.3\ssW{}}&
\ssW{}26.8&\wtC{cbScarlet}&
\ssW{}25.1&\ssW{}&
\ssW{}31.6&\wtC{cbScarlet}&
\ssW{}28.0&\wtC{cbScarlet}
\\ 
EnC &
\multicolumn{2}{c}{\ssW{}9.72\ssW{}}&
\ssW{}11.2&\wtC{cbScarlet}&
\ssW{}7.86&\btC{cbScarlet}&
\ssW{}11.8&\wtC{cbScarlet}&
&
\multicolumn{2}{c}{\ssW{}6.36\ssW{}}&
\ssW{}10.6&\wtC{cbScarlet}&
\ssW{}6.80&\wtC{cbScarlet}&
\ssW{}11.5&\wtC{cbScarlet}&
&
\multicolumn{2}{c}{\ssW{}6.37\ssW{}}&
\ssW{}10.5&\wtC{cbScarlet}&
\ssW{}6.18&\ssW{}&
\ssW{}10.9&\wtC{cbScarlet}&
&
\multicolumn{2}{c}{\ssW{}6.02\ssW{}}&
\ssW{}10.5&\wtC{cbScarlet}&
\ssW{}6.33&\wtC{cbScarlet}&
\ssW{}11.7&\wtC{cbScarlet}&
\ssW{}4.47&\btC{cbScarlet}
\\ 
EnH &
\multicolumn{2}{c}{\ssW{}5.03\ssW{}}&
\ssW{}7.19&\wtC{cbScarlet}&
\ssW{}4.04&\btC{cbScarlet}&
\ssW{}7.85&\wtC{cbScarlet}&
&
\multicolumn{2}{c}{\ssW{}3.45\ssW{}}&
\ssW{}7.57&\wtC{cbScarlet}&
\ssW{}3.88&\wtC{cbScarlet}&
\ssW{}7.62&\wtC{cbScarlet}&
&
\multicolumn{2}{c}{\ssW{}3.28\ssW{}}&
\ssW{}7.64&\wtC{cbScarlet}&
\ssW{}3.88&\wtC{cbScarlet}&
\ssW{}7.59&\wtC{cbScarlet}&
&
\multicolumn{2}{c}{\ssW{}3.51\ssW{}}&
\ssW{}7.56&\wtC{cbScarlet}&
\ssW{}3.86&\wtC{cbScarlet}&
\ssW{}7.65&\wtC{cbScarlet}&
\ssW{}0.33&\btC{cbScarlet}
\\ 
Tow &
\multicolumn{2}{c}{\ssW{}13.4\ssW{}}&
\ssW{}14.4&\wtC{cbScarlet}&
\ssW{}13.9&\wtC{cbScarlet}&
\ssW{}20.3&\wtC{cbScarlet}&
&
\multicolumn{2}{c}{\ssW{}13.0\ssW{}}&
\ssW{}14.1&\wtC{cbScarlet}&
\ssW{}14.0&\wtC{cbScarlet}&
\ssW{}20.8&\wtC{cbScarlet}&
&
\multicolumn{2}{c}{\ssW{}12.9\ssW{}}&
\ssW{}14.1&\wtC{cbScarlet}&
\ssW{}13.7&\wtC{cbScarlet}&
\ssW{}20.7&\wtC{cbScarlet}&
&
\multicolumn{2}{c}{\ssW{}13.0\ssW{}}&
\ssW{}14.3&\wtC{cbScarlet}&
\ssW{}13.3&\wtC{cbScarlet}&
\ssW{}20.5&\wtC{cbScarlet}&
\ssW{}11.2&\btC{cbScarlet}
\\ 
WiR &
\multicolumn{2}{c}{\ssW{}63.1\ssW{}}&
\ssW{}63.7&\wtC{cbScarlet}&
\ssW{}62.4&\ssW{}&
\ssW{}64.6&\wtC{cbScarlet}&
&
\multicolumn{2}{c}{\ssW{}63.3\ssW{}}&
\ssW{}63.4&\ssW{}&
\ssW{}63.2&\ssW{}&
\ssW{}64.4&\wtC{cbScarlet}&
&
\multicolumn{2}{c}{\ssW{}63.6\ssW{}}&
\ssW{}63.5&\ssW{}&
\ssW{}63.2&\btC{cbScarlet}&
\ssW{}64.3&\wtC{cbScarlet}&
&
\multicolumn{2}{c}{\ssW{}63.4\ssW{}}&
\ssW{}63.8&\ssW{}&
\ssW{}63.3&\ssW{}&
\ssW{}63.7&\wtC{cbScarlet}&
\ssW{}72.6&\wtC{cbScarlet}
\\ 
WiW &
\multicolumn{2}{c}{\ssW{}70.5\ssW{}}&
\ssW{}70.5&\ssW{}&
\ssW{}70.1&\ssW{}&
\ssW{}71.3&\wtC{cbScarlet}&
&
\multicolumn{2}{c}{\ssW{}70.4\ssW{}}&
\ssW{}70.0&\wtC{cbScarlet}&
\ssW{}70.3&\wtC{cbScarlet}&
\ssW{}71.6&\wtC{cbScarlet}&
&
\multicolumn{2}{c}{\ssW{}69.7\ssW{}}&
\ssW{}70.5&\wtC{cbScarlet}&
\ssW{}70.1&\wtC{cbScarlet}&
\ssW{}71.0&\wtC{cbScarlet}&
&
\multicolumn{2}{c}{\ssW{}70.2\ssW{}}&
\ssW{}70.5&\wtC{cbScarlet}&
\ssW{}70.3&\wtC{cbScarlet}&
\ssW{}71.8&\wtC{cbScarlet}&
\ssW{}72.2&\wtC{cbScarlet}
\\ 
Yac &
\multicolumn{2}{c}{\ssW{}1.23\ssW{}}&
\ssW{}1.23&\ssW{}&
\ssW{}0.78&\btC{cbScarlet}&
\ssW{}0.95&\btC{cbScarlet}&
&
\multicolumn{2}{c}{\ssW{}1.16\ssW{}}&
\ssW{}1.24&\wtC{cbScarlet}&
\ssW{}0.73&\btC{cbScarlet}&
\ssW{}0.77&\btC{cbScarlet}&
&
\multicolumn{2}{c}{\ssW{}1.17\ssW{}}&
\ssW{}1.24&\ssW{}&
\ssW{}0.73&\btC{cbScarlet}&
\ssW{}0.77&\btC{cbScarlet}&
&
\multicolumn{2}{c}{\ssW{}1.17\ssW{}}&
\ssW{}1.23&\wtC{cbScarlet}&
\ssW{}0.71&\btC{cbScarlet}&
\ssW{}0.86&\btC{cbScarlet}&
\ssW{}0.91&\btC{cbScarlet}
\\ 
\midrule 
B/W &
\multicolumn{2}{c}{---}&
\multicolumn{2}{c}{0/8}&
\multicolumn{2}{c}{4/2}&
\multicolumn{2}{c}{1/9}&
&
\multicolumn{2}{c}{---}&
\multicolumn{2}{c}{0/8}&
\multicolumn{2}{c}{4/5}&
\multicolumn{2}{c}{1/9}&
&
\multicolumn{2}{c}{---}&
\multicolumn{2}{c}{0/8}&
\multicolumn{2}{c}{4/4}&
\multicolumn{2}{c}{1/9}&
&
\multicolumn{2}{c}{---}&
\multicolumn{2}{c}{0/9}&
\multicolumn{2}{c}{3/5}&
\multicolumn{2}{c}{1/9}&
\multicolumn{2}{c}{5/5}\\

\bottomrule

\end{tabular}
}
\vspace{-3mm}
\end{table*}

\begin{table*}
\caption{Median validation and test NMSE of 30 runs with $\ell = 31$. Details as in Table~\ref{tab:validation3}.}\label{tab:validation4}
\centering
\setlength{\tabcolsep}{0.001em}
\tiny
\scalebox{1.0}{
\begin{tabular}{lScScScSccScScScSccScScScSccScScScScSc}
\toprule
\multicolumn{38}{c}{\small \textbf{Validation $\ell = 31$}} \\ 
 & \multicolumn{8}{c}{$n^\text{pop}=1000$} & \ssW{}\ssW{} & \multicolumn{8}{c}{IMS $g=4$} & \ssW{}\ssW{} & \multicolumn{8}{c}{IMS $g=6$} & \ssW{}\ssW{} & \multicolumn{9}{c}{IMS $g=8$} \\
 \ssW{}& 
\multicolumn{2}{c}{\color{cbScarlet}G} & \multicolumn{2}{c}{\color{black}T\textsuperscript{$h$}} & \multicolumn{2}{c}{\color{black}T\textsuperscript{$\ell$}} & \multicolumn{2}{c}{\color{black}R} & & 
\multicolumn{2}{c}{\color{cbScarlet}G} & \multicolumn{2}{c}{\color{black}T\textsuperscript{$h$}} & \multicolumn{2}{c}{\color{black}T\textsuperscript{$\ell$}} & \multicolumn{2}{c}{\color{black}R} & & 
\multicolumn{2}{c}{\color{cbScarlet}G} & \multicolumn{2}{c}{\color{black}T\textsuperscript{$h$}} & \multicolumn{2}{c}{\color{black}T\textsuperscript{$\ell$}} & \multicolumn{2}{c}{\color{black}R} & & 
\multicolumn{2}{c}{\color{cbScarlet}G} & \multicolumn{2}{c}{\color{black}T\textsuperscript{$h$}} & \multicolumn{2}{c}{\color{black}T\textsuperscript{$\ell$}} & \multicolumn{2}{c}{\color{black}R} & \multicolumn{2}{c}{D}\\
\midrule

Air &
\multicolumn{2}{c}{\ssW{}26.4\ssW{}}&
\ssW{}33.8&\wtC{cbScarlet}&
\ssW{}32.1&\wtC{cbScarlet}&
\ssW{}36.0&\wtC{cbScarlet}&
&
\multicolumn{2}{c}{\ssW{}24.9\ssW{}}&
\ssW{}25.9&\wtC{cbScarlet}&
\ssW{}23.2&\btC{cbScarlet}&
\ssW{}37.0&\wtC{cbScarlet}&
&
\multicolumn{2}{c}{\ssW{}25.0\ssW{}}&
\ssW{}27.1&\wtC{cbScarlet}&
\ssW{}24.9&\ssW{}&
\ssW{}37.6&\wtC{cbScarlet}&
&
\multicolumn{2}{c}{\ssW{}24.8\ssW{}}&
\ssW{}28.4&\wtC{cbScarlet}&
\ssW{}24.8&\ssW{}&
\ssW{}37.1&\wtC{cbScarlet}&
\ssW{}31.1&\wtC{cbScarlet}
\\ 
Bos &
\multicolumn{2}{c}{\ssW{}22.3\ssW{}}&
\ssW{}21.1&\ssW{}&
\ssW{}19.2&\ssW{}&
\ssW{}24.8&\wtC{cbScarlet}&
&
\multicolumn{2}{c}{\ssW{}16.7\ssW{}}&
\ssW{}18.8&\wtC{cbScarlet}&
\ssW{}16.2&\wtC{cbScarlet}&
\ssW{}20.4&\wtC{cbScarlet}&
&
\multicolumn{2}{c}{\ssW{}17.4\ssW{}}&
\ssW{}18.3&\wtC{cbScarlet}&
\ssW{}17.3&\wtC{cbScarlet}&
\ssW{}20.6&\wtC{cbScarlet}&
&
\multicolumn{2}{c}{\ssW{}17.3\ssW{}}&
\ssW{}18.4&\wtC{cbScarlet}&
\ssW{}17.6&\wtC{cbScarlet}&
\ssW{}20.4&\wtC{cbScarlet}&
\ssW{}22.9&\wtC{cbScarlet}
\\ 
Con &
\multicolumn{2}{c}{\ssW{}17.3\ssW{}}&
\ssW{}18.6&\wtC{cbScarlet}&
\ssW{}17.9&\wtC{cbScarlet}&
\ssW{}20.8&\wtC{cbScarlet}&
&
\multicolumn{2}{c}{\ssW{}16.0\ssW{}}&
\ssW{}17.6&\wtC{cbScarlet}&
\ssW{}16.7&\wtC{cbScarlet}&
\ssW{}20.5&\wtC{cbScarlet}&
&
\multicolumn{2}{c}{\ssW{}16.6\ssW{}}&
\ssW{}18.1&\wtC{cbScarlet}&
\ssW{}16.4&\wtC{cbScarlet}&
\ssW{}20.1&\wtC{cbScarlet}&
&
\multicolumn{2}{c}{\ssW{}16.1\ssW{}}&
\ssW{}18.3&\wtC{cbScarlet}&
\ssW{}17.2&\wtC{cbScarlet}&
\ssW{}20.2&\wtC{cbScarlet}&
\ssW{}22.7&\wtC{cbScarlet}
\\ 
Dow &
\multicolumn{2}{c}{\ssW{}21.3\ssW{}}&
\ssW{}22.6&\wtC{cbScarlet}&
\ssW{}22.6&\ssW{}&
\ssW{}24.3&\wtC{cbScarlet}&
&
\multicolumn{2}{c}{\ssW{}19.4\ssW{}}&
\ssW{}21.6&\wtC{cbScarlet}&
\ssW{}19.2&\ssW{}&
\ssW{}25.6&\wtC{cbScarlet}&
&
\multicolumn{2}{c}{\ssW{}19.4\ssW{}}&
\ssW{}21.2&\wtC{cbScarlet}&
\ssW{}19.4&\wtC{cbScarlet}&
\ssW{}25.4&\wtC{cbScarlet}&
&
\multicolumn{2}{c}{\ssW{}19.2\ssW{}}&
\ssW{}21.9&\wtC{cbScarlet}&
\ssW{}20.1&\wtC{cbScarlet}&
\ssW{}25.8&\wtC{cbScarlet}&
\ssW{}30.6&\wtC{cbScarlet}
\\ 
EnC &
\multicolumn{2}{c}{\ssW{}5.14\ssW{}}&
\ssW{}5.60&\wtC{cbScarlet}&
\ssW{}4.99&\btC{cbScarlet}&
\ssW{}7.62&\wtC{cbScarlet}&
&
\multicolumn{2}{c}{\ssW{}4.62\ssW{}}&
\ssW{}5.51&\wtC{cbScarlet}&
\ssW{}4.82&\wtC{cbScarlet}&
\ssW{}8.04&\wtC{cbScarlet}&
&
\multicolumn{2}{c}{\ssW{}4.35\ssW{}}&
\ssW{}6.04&\wtC{cbScarlet}&
\ssW{}4.56&\wtC{cbScarlet}&
\ssW{}8.48&\wtC{cbScarlet}&
&
\multicolumn{2}{c}{\ssW{}4.37\ssW{}}&
\ssW{}5.65&\wtC{cbScarlet}&
\ssW{}4.73&\wtC{cbScarlet}&
\ssW{}7.81&\wtC{cbScarlet}&
\ssW{}4.23&\btC{cbScarlet}
\\ 
EnH &
\multicolumn{2}{c}{\ssW{}2.29\ssW{}}&
\ssW{}2.54&\wtC{cbScarlet}&
\ssW{}1.75&\btC{cbScarlet}&
\ssW{}6.21&\wtC{cbScarlet}&
&
\multicolumn{2}{c}{\ssW{}1.95\ssW{}}&
\ssW{}3.05&\wtC{cbScarlet}&
\ssW{}1.72&\btC{cbScarlet}&
\ssW{}4.97&\wtC{cbScarlet}&
&
\multicolumn{2}{c}{\ssW{}2.00\ssW{}}&
\ssW{}2.84&\wtC{cbScarlet}&
\ssW{}1.65&\btC{cbScarlet}&
\ssW{}5.93&\wtC{cbScarlet}&
&
\multicolumn{2}{c}{\ssW{}1.88\ssW{}}&
\ssW{}3.10&\wtC{cbScarlet}&
\ssW{}1.62&\ssW{}&
\ssW{}6.11&\wtC{cbScarlet}&
\ssW{}0.43&\btC{cbScarlet}
\\ 
Tow &
\multicolumn{2}{c}{\ssW{}12.0\ssW{}}&
\ssW{}13.0&\wtC{cbScarlet}&
\ssW{}12.6&\wtC{cbScarlet}&
\ssW{}17.5&\wtC{cbScarlet}&
&
\multicolumn{2}{c}{\ssW{}11.8\ssW{}}&
\ssW{}12.3&\wtC{cbScarlet}&
\ssW{}11.9&\ssW{}&
\ssW{}17.8&\wtC{cbScarlet}&
&
\multicolumn{2}{c}{\ssW{}11.7\ssW{}}&
\ssW{}12.2&\wtC{cbScarlet}&
\ssW{}12.2&\wtC{cbScarlet}&
\ssW{}16.6&\wtC{cbScarlet}&
&
\multicolumn{2}{c}{\ssW{}12.0\ssW{}}&
\ssW{}12.4&\wtC{cbScarlet}&
\ssW{}11.8&\ssW{}&
\ssW{}17.6&\wtC{cbScarlet}&
\ssW{}11.2&\btC{cbScarlet}
\\ 
WiR &
\multicolumn{2}{c}{\ssW{}64.2\ssW{}}&
\ssW{}64.7&\wtC{cbScarlet}&
\ssW{}64.7&\wtC{cbScarlet}&
\ssW{}65.9&\wtC{cbScarlet}&
&
\multicolumn{2}{c}{\ssW{}62.8\ssW{}}&
\ssW{}62.4&\wtC{cbScarlet}&
\ssW{}62.6&\ssW{}&
\ssW{}64.5&\wtC{cbScarlet}&
&
\multicolumn{2}{c}{\ssW{}62.3\ssW{}}&
\ssW{}63.6&\wtC{cbScarlet}&
\ssW{}62.1&\ssW{}&
\ssW{}64.1&\wtC{cbScarlet}&
&
\multicolumn{2}{c}{\ssW{}62.6\ssW{}}&
\ssW{}62.9&\wtC{cbScarlet}&
\ssW{}61.8&\ssW{}&
\ssW{}64.6&\wtC{cbScarlet}&
\ssW{}71.7&\wtC{cbScarlet}
\\ 
WiW &
\multicolumn{2}{c}{\ssW{}70.2\ssW{}}&
\ssW{}70.4&\wtC{cbScarlet}&
\ssW{}70.9&\ssW{}&
\ssW{}71.4&\wtC{cbScarlet}&
&
\multicolumn{2}{c}{\ssW{}69.6\ssW{}}&
\ssW{}69.7&\ssW{}&
\ssW{}69.7&\ssW{}&
\ssW{}71.1&\wtC{cbScarlet}&
&
\multicolumn{2}{c}{\ssW{}70.0\ssW{}}&
\ssW{}70.2&\wtC{cbScarlet}&
\ssW{}70.0&\ssW{}&
\ssW{}71.0&\wtC{cbScarlet}&
&
\multicolumn{2}{c}{\ssW{}70.0\ssW{}}&
\ssW{}70.1&\wtC{cbScarlet}&
\ssW{}69.6&\ssW{}&
\ssW{}71.2&\wtC{cbScarlet}&
\ssW{}72.2&\wtC{cbScarlet}
\\ 
Yac &
\multicolumn{2}{c}{\ssW{}0.46\ssW{}}&
\ssW{}0.59&\wtC{cbScarlet}&
\ssW{}0.42&\btC{cbScarlet}&
\ssW{}0.57&\wtC{cbScarlet}&
&
\multicolumn{2}{c}{\ssW{}0.37\ssW{}}&
\ssW{}0.51&\wtC{cbScarlet}&
\ssW{}0.40&\wtC{cbScarlet}&
\ssW{}0.59&\wtC{cbScarlet}&
&
\multicolumn{2}{c}{\ssW{}0.38\ssW{}}&
\ssW{}0.56&\wtC{cbScarlet}&
\ssW{}0.38&\ssW{}&
\ssW{}0.54&\wtC{cbScarlet}&
&
\multicolumn{2}{c}{\ssW{}0.40\ssW{}}&
\ssW{}0.54&\wtC{cbScarlet}&
\ssW{}0.42&\wtC{cbScarlet}&
\ssW{}0.52&\wtC{cbScarlet}&
\ssW{}0.88&\wtC{cbScarlet}
\\ 
\midrule 
B/W &
\multicolumn{2}{c}{---}&
\multicolumn{2}{c}{0/9}&
\multicolumn{2}{c}{3/4}&
\multicolumn{2}{c}{0/10}&
&
\multicolumn{2}{c}{---}&
\multicolumn{2}{c}{0/9}&
\multicolumn{2}{c}{2/4}&
\multicolumn{2}{c}{0/10}&
&
\multicolumn{2}{c}{---}&
\multicolumn{2}{c}{0/10}&
\multicolumn{2}{c}{1/5}&
\multicolumn{2}{c}{0/10}&
&
\multicolumn{2}{c}{---}&
\multicolumn{2}{c}{0/10}&
\multicolumn{2}{c}{0/5}&
\multicolumn{2}{c}{0/10}&
\multicolumn{2}{c}{3/7}\\

\midrule 

\multicolumn{38}{c}{\small \textbf{Test $\ell = 31$}} \\ 
 & \multicolumn{8}{c}{$n^\text{pop}=1000$} & \ssW{}\ssW{} & \multicolumn{8}{c}{IMS $g=4$} & \ssW{}\ssW{} & \multicolumn{8}{c}{IMS $g=6$} & \ssW{}\ssW{} & \multicolumn{9}{c}{IMS $g=8$} \\
 \ssW{}& 
\multicolumn{2}{c}{\color{cbScarlet}G} & \multicolumn{2}{c}{\color{black}T\textsuperscript{$h$}} & \multicolumn{2}{c}{\color{black}T\textsuperscript{$\ell$}} & \multicolumn{2}{c}{\color{black}R} & & 
\multicolumn{2}{c}{\color{cbScarlet}G} & \multicolumn{2}{c}{\color{black}T\textsuperscript{$h$}} & \multicolumn{2}{c}{\color{black}T\textsuperscript{$\ell$}} & \multicolumn{2}{c}{\color{black}R} & & 
\multicolumn{2}{c}{\color{cbScarlet}G} & \multicolumn{2}{c}{\color{black}T\textsuperscript{$h$}} & \multicolumn{2}{c}{\color{black}T\textsuperscript{$\ell$}} & \multicolumn{2}{c}{\color{black}R} & & 
\multicolumn{2}{c}{\color{cbScarlet}G} & \multicolumn{2}{c}{\color{black}T\textsuperscript{$h$}} & \multicolumn{2}{c}{\color{black}T\textsuperscript{$\ell$}} & \multicolumn{2}{c}{\color{black}R} & \multicolumn{2}{c}{D}\\
\midrule

Air &
\multicolumn{2}{c}{\ssW{}26.4\ssW{}}&
\ssW{}33.5&\wtC{cbScarlet}&
\ssW{}30.8&\wtC{cbScarlet}&
\ssW{}37.1&\wtC{cbScarlet}&
&
\multicolumn{2}{c}{\ssW{}25.9\ssW{}}&
\ssW{}26.5&\wtC{cbScarlet}&
\ssW{}23.3&\btC{cbScarlet}&
\ssW{}37.6&\wtC{cbScarlet}&
&
\multicolumn{2}{c}{\ssW{}24.9\ssW{}}&
\ssW{}27.1&\wtC{cbScarlet}&
\ssW{}24.7&\ssW{}&
\ssW{}39.2&\wtC{cbScarlet}&
&
\multicolumn{2}{c}{\ssW{}24.9\ssW{}}&
\ssW{}28.8&\wtC{cbScarlet}&
\ssW{}26.1&\ssW{}&
\ssW{}38.2&\wtC{cbScarlet}&
\ssW{}30.8&\wtC{cbScarlet}
\\ 
Bos &
\multicolumn{2}{c}{\ssW{}21.4\ssW{}}&
\ssW{}22.8&\ssW{}&
\ssW{}21.6&\wtC{cbScarlet}&
\ssW{}26.2&\wtC{cbScarlet}&
&
\multicolumn{2}{c}{\ssW{}20.1\ssW{}}&
\ssW{}21.3&\ssW{}&
\ssW{}21.8&\wtC{cbScarlet}&
\ssW{}23.4&\wtC{cbScarlet}&
&
\multicolumn{2}{c}{\ssW{}20.9\ssW{}}&
\ssW{}21.2&\wtC{cbScarlet}&
\ssW{}22.2&\wtC{cbScarlet}&
\ssW{}23.2&\wtC{cbScarlet}&
&
\multicolumn{2}{c}{\ssW{}20.2\ssW{}}&
\ssW{}22.3&\wtC{cbScarlet}&
\ssW{}22.6&\wtC{cbScarlet}&
\ssW{}26.0&\wtC{cbScarlet}&
\ssW{}26.1&\wtC{cbScarlet}
\\ 
Con &
\multicolumn{2}{c}{\ssW{}17.6\ssW{}}&
\ssW{}18.7&\wtC{cbScarlet}&
\ssW{}17.8&\wtC{cbScarlet}&
\ssW{}21.5&\wtC{cbScarlet}&
&
\multicolumn{2}{c}{\ssW{}16.9\ssW{}}&
\ssW{}18.1&\wtC{cbScarlet}&
\ssW{}17.1&\wtC{cbScarlet}&
\ssW{}21.2&\wtC{cbScarlet}&
&
\multicolumn{2}{c}{\ssW{}16.7\ssW{}}&
\ssW{}18.8&\wtC{cbScarlet}&
\ssW{}16.9&\ssW{}&
\ssW{}21.1&\wtC{cbScarlet}&
&
\multicolumn{2}{c}{\ssW{}17.2\ssW{}}&
\ssW{}18.3&\wtC{cbScarlet}&
\ssW{}17.0&\ssW{}&
\ssW{}21.5&\wtC{cbScarlet}&
\ssW{}21.3&\wtC{cbScarlet}
\\ 
Dow &
\multicolumn{2}{c}{\ssW{}20.3\ssW{}}&
\ssW{}21.9&\wtC{cbScarlet}&
\ssW{}22.2&\wtC{cbScarlet}&
\ssW{}24.4&\wtC{cbScarlet}&
&
\multicolumn{2}{c}{\ssW{}19.2\ssW{}}&
\ssW{}20.7&\wtC{cbScarlet}&
\ssW{}19.1&\ssW{}&
\ssW{}24.4&\wtC{cbScarlet}&
&
\multicolumn{2}{c}{\ssW{}18.9\ssW{}}&
\ssW{}21.4&\wtC{cbScarlet}&
\ssW{}18.6&\ssW{}&
\ssW{}24.4&\wtC{cbScarlet}&
&
\multicolumn{2}{c}{\ssW{}18.7\ssW{}}&
\ssW{}22.2&\wtC{cbScarlet}&
\ssW{}20.2&\wtC{cbScarlet}&
\ssW{}25.5&\wtC{cbScarlet}&
\ssW{}28.0&\wtC{cbScarlet}
\\ 
EnC &
\multicolumn{2}{c}{\ssW{}5.28\ssW{}}&
\ssW{}5.91&\wtC{cbScarlet}&
\ssW{}4.76&\btC{cbScarlet}&
\ssW{}7.00&\wtC{cbScarlet}&
&
\multicolumn{2}{c}{\ssW{}4.43\ssW{}}&
\ssW{}5.76&\wtC{cbScarlet}&
\ssW{}4.79&\wtC{cbScarlet}&
\ssW{}7.69&\wtC{cbScarlet}&
&
\multicolumn{2}{c}{\ssW{}4.44\ssW{}}&
\ssW{}6.05&\wtC{cbScarlet}&
\ssW{}4.71&\ssW{}&
\ssW{}8.73&\wtC{cbScarlet}&
&
\multicolumn{2}{c}{\ssW{}4.60\ssW{}}&
\ssW{}5.62&\wtC{cbScarlet}&
\ssW{}4.77&\wtC{cbScarlet}&
\ssW{}7.94&\wtC{cbScarlet}&
\ssW{}4.47&\btC{cbScarlet}
\\ 
EnH &
\multicolumn{2}{c}{\ssW{}2.29\ssW{}}&
\ssW{}2.49&\wtC{cbScarlet}&
\ssW{}1.83&\btC{cbScarlet}&
\ssW{}5.98&\wtC{cbScarlet}&
&
\multicolumn{2}{c}{\ssW{}2.05\ssW{}}&
\ssW{}3.20&\wtC{cbScarlet}&
\ssW{}1.58&\btC{cbScarlet}&
\ssW{}5.12&\wtC{cbScarlet}&
&
\multicolumn{2}{c}{\ssW{}2.10\ssW{}}&
\ssW{}3.07&\wtC{cbScarlet}&
\ssW{}1.75&\btC{cbScarlet}&
\ssW{}5.77&\wtC{cbScarlet}&
&
\multicolumn{2}{c}{\ssW{}2.00\ssW{}}&
\ssW{}2.91&\wtC{cbScarlet}&
\ssW{}1.55&\btC{cbScarlet}&
\ssW{}6.51&\wtC{cbScarlet}&
\ssW{}0.33&\btC{cbScarlet}
\\ 
Tow &
\multicolumn{2}{c}{\ssW{}12.2\ssW{}}&
\ssW{}13.2&\wtC{cbScarlet}&
\ssW{}13.1&\wtC{cbScarlet}&
\ssW{}18.7&\wtC{cbScarlet}&
&
\multicolumn{2}{c}{\ssW{}12.1\ssW{}}&
\ssW{}12.6&\wtC{cbScarlet}&
\ssW{}12.0&\btC{cbScarlet}&
\ssW{}18.2&\wtC{cbScarlet}&
&
\multicolumn{2}{c}{\ssW{}12.1\ssW{}}&
\ssW{}12.4&\wtC{cbScarlet}&
\ssW{}12.3&\ssW{}&
\ssW{}16.8&\wtC{cbScarlet}&
&
\multicolumn{2}{c}{\ssW{}12.2\ssW{}}&
\ssW{}12.7&\wtC{cbScarlet}&
\ssW{}12.0&\ssW{}&
\ssW{}17.2&\wtC{cbScarlet}&
\ssW{}11.2&\btC{cbScarlet}
\\ 
WiR &
\multicolumn{2}{c}{\ssW{}62.1\ssW{}}&
\ssW{}63.1&\ssW{}&
\ssW{}62.1&\ssW{}&
\ssW{}63.5&\wtC{cbScarlet}&
&
\multicolumn{2}{c}{\ssW{}62.7\ssW{}}&
\ssW{}63.1&\wtC{cbScarlet}&
\ssW{}61.9&\ssW{}&
\ssW{}63.9&\wtC{cbScarlet}&
&
\multicolumn{2}{c}{\ssW{}62.4\ssW{}}&
\ssW{}62.9&\wtC{cbScarlet}&
\ssW{}63.3&\wtC{cbScarlet}&
\ssW{}64.2&\wtC{cbScarlet}&
&
\multicolumn{2}{c}{\ssW{}61.9\ssW{}}&
\ssW{}63.0&\wtC{cbScarlet}&
\ssW{}62.9&\wtC{cbScarlet}&
\ssW{}63.4&\wtC{cbScarlet}&
\ssW{}72.6&\wtC{cbScarlet}
\\ 
WiW &
\multicolumn{2}{c}{\ssW{}69.0\ssW{}}&
\ssW{}69.7&\wtC{cbScarlet}&
\ssW{}69.8&\wtC{cbScarlet}&
\ssW{}70.2&\wtC{cbScarlet}&
&
\multicolumn{2}{c}{\ssW{}69.4\ssW{}}&
\ssW{}69.3&\ssW{}&
\ssW{}69.2&\ssW{}&
\ssW{}70.6&\wtC{cbScarlet}&
&
\multicolumn{2}{c}{\ssW{}69.1\ssW{}}&
\ssW{}69.4&\wtC{cbScarlet}&
\ssW{}69.2&\wtC{cbScarlet}&
\ssW{}70.7&\wtC{cbScarlet}&
&
\multicolumn{2}{c}{\ssW{}69.1\ssW{}}&
\ssW{}69.6&\wtC{cbScarlet}&
\ssW{}69.3&\btC{cbScarlet}&
\ssW{}70.5&\wtC{cbScarlet}&
\ssW{}72.2&\wtC{cbScarlet}
\\ 
Yac &
\multicolumn{2}{c}{\ssW{}0.52\ssW{}}&
\ssW{}0.66&\wtC{cbScarlet}&
\ssW{}0.49&\btC{cbScarlet}&
\ssW{}0.66&\wtC{cbScarlet}&
&
\multicolumn{2}{c}{\ssW{}0.50\ssW{}}&
\ssW{}0.58&\wtC{cbScarlet}&
\ssW{}0.47&\ssW{}&
\ssW{}0.67&\wtC{cbScarlet}&
&
\multicolumn{2}{c}{\ssW{}0.50\ssW{}}&
\ssW{}0.64&\wtC{cbScarlet}&
\ssW{}0.48&\btC{cbScarlet}&
\ssW{}0.63&\wtC{cbScarlet}&
&
\multicolumn{2}{c}{\ssW{}0.53\ssW{}}&
\ssW{}0.63&\wtC{cbScarlet}&
\ssW{}0.48&\ssW{}&
\ssW{}0.70&\wtC{cbScarlet}&
\ssW{}0.91&\wtC{cbScarlet}
\\ 
\midrule 
B/W &
\multicolumn{2}{c}{---}&
\multicolumn{2}{c}{0/8}&
\multicolumn{2}{c}{3/6}&
\multicolumn{2}{c}{0/10}&
&
\multicolumn{2}{c}{---}&
\multicolumn{2}{c}{0/8}&
\multicolumn{2}{c}{3/3}&
\multicolumn{2}{c}{0/10}&
&
\multicolumn{2}{c}{---}&
\multicolumn{2}{c}{0/10}&
\multicolumn{2}{c}{2/3}&
\multicolumn{2}{c}{0/10}&
&
\multicolumn{2}{c}{---}&
\multicolumn{2}{c}{0/10}&
\multicolumn{2}{c}{2/4}&
\multicolumn{2}{c}{0/10}&
\multicolumn{2}{c}{3/7}\\

\bottomrule

\end{tabular}
}
\vspace{-3mm}
\end{table*}

\subsection{Results: benchmarking GP-GOMEA}
We consider validation and test NMSE. We now show validation rather than training error because the IMS returns the solution which better generalizes to the validation set among the ones found by different runs (same for DTR due to cross-validation). 
Tables~\ref{tab:validation3},~\ref{tab:validation4}, and~\ref{tab:validation5} show the results for maximum sizes $\ell=15,31,63$ ($h=3,4,5$) respectively.
On each set of results, the Friedman test reveals significant differences among the algorithms. As we are only interested in benchmarking GP-GOMEA, we test whether significant performance differences exist only between GP-GOMEA and the other algorithms (with Bonferroni-corrected Wilcoxon signed-rank test).

We begin with some general results. Overall, error magnitudes are lower for larger values of $\ell$. This is not surprising: limiting solution size limits the complexity of relationships that can be modeled. Another general result is that errors on validation and test set are generally close. Likely, the validation data is a sufficiently accurate surrogate of the test data in these datasets, and solution size limitations make over-fitting unlikely. Finally, note that the results for DTR are the same in all tables.

We now compare GP-GOMEA with GP-Trad\textsuperscript{$h$}, focusing on statistical significance tests (see rows ``B/W'' of the tables), over all size limit configurations. Recall that these two algorithms work with the same type of limitation, i.e., based on maximum tree height. No matter the population sizing method, GP-GOMEA is almost always significantly better than GP-Trad\textsuperscript{$h$}. GP-GOMEA relies on the LT with improved linkage learning, which we showed to be superior to using the RT, i.e., blind variation, in the previous series of experiments (Sec.~\ref{sec:exp-normalized-mi}, \ref{sec:exp-ll-with-erc}). Subtree crossover and subtree mutation are blind as well, and can only swap subtrees, which may be a limitation.

GP-GOMEA and GP-Trad\textsuperscript{$\ell$} are compared next. Recall that GP-Trad\textsuperscript{$\ell$} is allowed to evolve any tree shape, as long as the limit in number of nodes is respected. Having this extra freedom, GP-Trad\textsuperscript{$\ell$} performs better than GP-Trad\textsuperscript{$h$} (not explicitly reported in the tables), which confirms previous literature results~\citep{gathercole1996adverse,langdon1997analysis}. No marked difference exists between GP-GOMEA and GP-Trad\textsuperscript{$\ell$} along different configurations. By counting the number of times one EA is found to be significantly better than the other along \emph{all} 240 comparisons, GP-GOMEA beats GP-Trad\textsuperscript{$\ell$} by a small margin: 87 significantly lower error distributions vs. 65 (88 draws).

For the traditional use of a single population ($n^\text{pop}=1000$), GP-Trad\textsuperscript{$\ell$} is slightly better than GP-GOMEA for $\ell=15$ (Table~\ref{tab:validation3}), slightly worse for $\ell=31$ (Table~\ref{tab:validation4}), and similar for $\ell=63$ (Table~\ref{tab:validation5}), on both validation and test errors. 
The performance of the two (and also of the other EAs) improves when using the IMS. Although not explicitly shown in the tables, using the IMS is typically significantly better than not using it. When using a single fixed population size and a single run, only a single best-found solution is found. Depending on the configuration of that run, in particular the size of the population, that final solution may be underfitted or overfitted. When using a scheme such as the IMS, multiple solutions are marked best in the different interleaved runs. These solutions can subsequently be compared more in terms of generalization merits, i.e., by observing the associated performance on the validation set. The best performing solution can then ultimately be returned. Essentially, this thus provides a means to mitigate to some extent the problem of underfitting or overfitting. It should be noted, however, that the extent to which the setup of the IMS, particularly in terms of growing population sizes, contributes to this is not immediately clear. This could be studied by comparing with a scheme in which multiple runs are also performed, but all with a single population size. The final results of these runs can then also first be tested against the validation set. Likely, the use of a scheme like the IMS has an advantage because multiple population sizes will be tried. Therefore, likely a larger variety of results will be produced to test against the validation set, but a closer examination of this impact is left for future work.

The comparisons between GP-Trad\textsuperscript{$\ell$} and GP-GOMEA tend to shift in favor of the latter when using the IMS, particularly for larger values of $g$. For $g=4$, outcomes are still overall mixed along different $\ell$ limits. For $g=8$, GP-GOMEA is preferable, with moderately more significant wins for $\ell=15$, several more wins for $\ell=31$, and slightly more wins for $\ell=63$.

\begin{figure}
\centering
\setlength{\tabcolsep}{0.1em}
\def\arraystretch{0.5}
\begin{tabular}{cc}
\begin{sideways}\ \ \ \ \ \ \small Max population size\end{sideways}
& \includegraphics[width=0.6\linewidth]{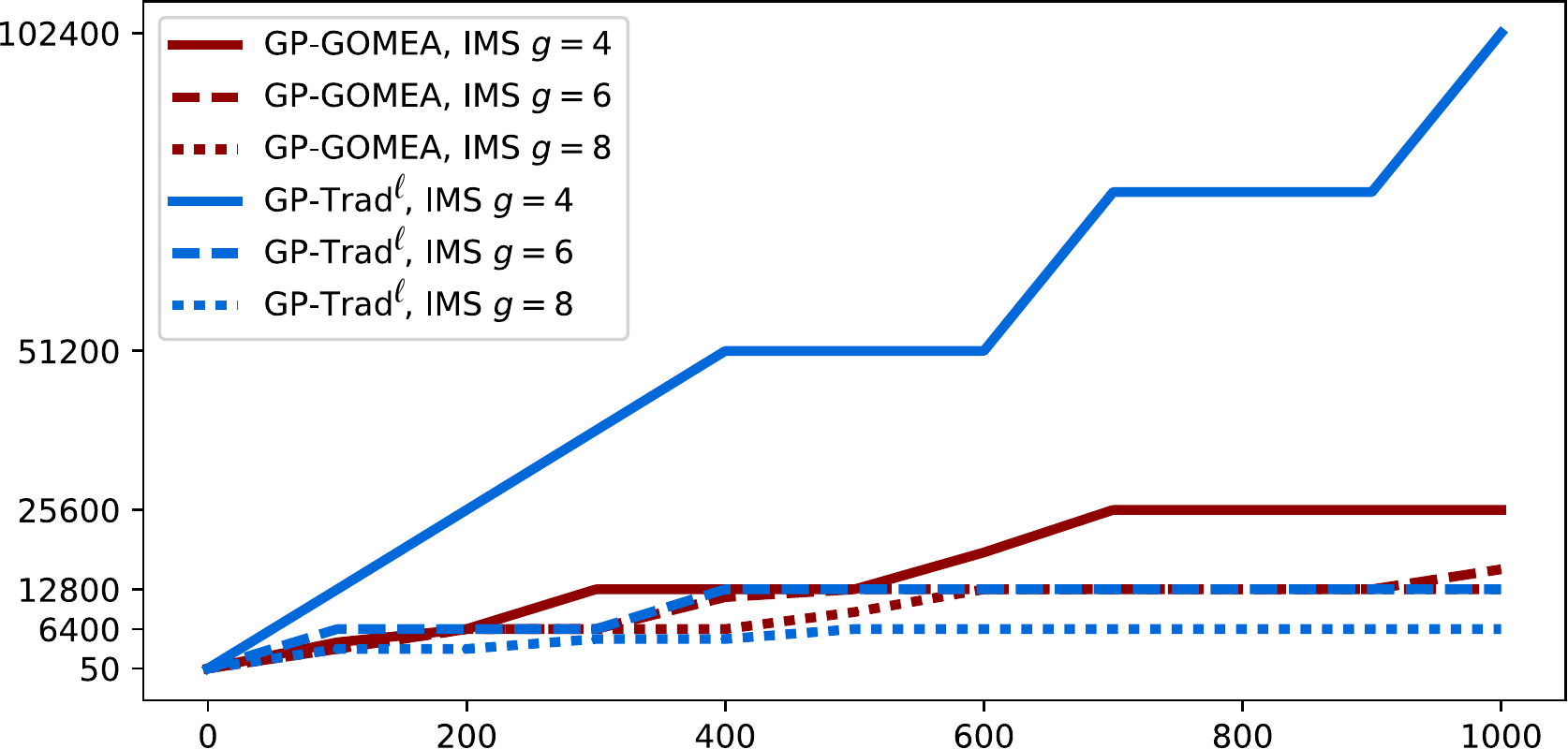}\\
& \ \ \ \ \ \ \small Time \\
\end{tabular}
\vspace{-3mm}
\caption{Maximum population size reached (vertical axis) in time (seconds, horizontal axis) with the IMS for GP-GOMEA ($h=4$ limit) and GP-Trad\textsuperscript{$\ell$} ($\ell=31$ limit), for $g \in \{4,6,8\}$. The median among problems and repetitions is shown.}\label{fig:ims-populations}\vspace{-3mm}
\end{figure}

To investigate further the comparison between GP-GOMEA and GP-Trad\textsuperscript{$\ell$}, we consider the effect of $g$ of the IMS for $\ell=31$ (similar results are found for the other size limits). Figure~\ref{fig:ims-populations} shows the median maximum population size reached by the IMS for different values of $g$ in GP-GOMEA and GP-Trad\textsuperscript{$\ell$}. As can be expected, the bigger $g$, the less runs and the smaller populations at play. GP-Trad\textsuperscript{$\ell$} reaches much bigger population sizes than GP-GOMEA when $g=4$ (on average 3 times bigger). This is because GP-Trad\textsuperscript{$\ell$} executes generations much faster than GP-GOMEA: it does not learn a linkage model, and performs $n^\text{pop}$ evaluations per generation. GP-GOMEA performs $(2\ell-2)n^\text{pop}$ variation steps (size of LT excluding its root times the population size) and up to $(2\ell-2)n^\text{pop}$ evaluations per generation (only meaningful variation steps are evaluated).

GP-Trad\textsuperscript{$\ell$} performs well for small values of $g$ due to huge populations being instantiated with trees of various shape, i.e., expensive random search. Note that this behavior may be problematic when limited memory is available, especially if caching mechanisms are desirable to reduce the number of expensive evaluations (e.g., caching the output of each node as in~\citep{pawlak2015semantic,virgolin2017scalable}).
On the other hand, GP-GOMEA works fairly well with much smaller populations, as long as they are big enough to enable effective linkage learning (the fixed $n^\text{pop}=1000$ is smaller than the population sizes reached with the IMS). Despite the disadvantage of adhering to a specific tree shape, GP-GOMEA is typically preferable than GP-Trad\textsuperscript{$\ell$} for larger values of $g$. Furthermore, Figure~\ref{fig:ims-populations} shows that GP-GOMEA population scaling behaves sensibly w.r.t. $g$, i.e., it does not grow abruptly when $g$ becomes small, nor shrink excessively when $g$ becomes larger. This latter aspect is because in GP-GOMEA populations ultimately converge to a same solution, and are terminated, allowing for bigger runs to start. In GP-Trad\textsuperscript{$\ell$} this is unlikely to happen, because of the use of mutation and stochastic (tournament) selection, stalling the IMS. For the larger $g=8$, GP-GOMEA reaches on average 1.6 times bigger populations than GP-Trad\textsuperscript{$\ell$}.

GP-RDO, although allowed to evolve trees of different shape like GP-Trad\textsuperscript{$\ell$}, performs poorly on all problems, with all settings. It performs significantly worse than GP-GOMEA almost everywhere (it is also worse than GP-Trad\textsuperscript{$\ell$}). It is known that GP-RDO normally finds big solutions, and it is also reasonable to expect that it needs big solutions to work well, e.g., to build a large set of diverse subtrees for the internal library queried by RDO~\citep{virgolin2019linear}. The strict size limitation basically breaks GP-RDO. However, we remark that this EA was never designed to work under these circumstances. In fact, when solution size is not strictly limited, GP-RDO achieves excellent performance~\citep{pawlak2015semantic}.

DTR is compared with GP-GOMEA using the IMS with $g=8$. Although GP-GOMEA is not optimized (e.g., by tuning the function set), it performs on par with tuned DTR for $\ell=15$, and better for $\ell=31,63$, on both validation and test sets. Where one algorithm outperforms the other, the magnitude of difference in errors are relatively large compared to the ones between EAs. This is because GP and DTR synthesize models of completely different nature (decision trees only use if-then-else statements).

\begin{table*}
\caption{Median validation and test NMSE of 30 runs with $\ell = 63$. Details as in Table~\ref{tab:validation3}.}\label{tab:validation5}
\centering
\setlength{\tabcolsep}{0.001em}
\tiny
\scalebox{1.0}{
\begin{tabular}{lScScScSccScScScSccScScScSccScScScScSc}
\toprule
\multicolumn{38}{c}{\small \textbf{Validation $\ell = 63$}} \\ 
 & \multicolumn{8}{c}{$n^\text{pop}=1000$} & \ssW{}\ssW{} & \multicolumn{8}{c}{IMS $g=4$} & \ssW{}\ssW{} & \multicolumn{8}{c}{IMS $g=6$} & \ssW{}\ssW{} & \multicolumn{9}{c}{IMS $g=8$} \\
 \ssW{}& 
\multicolumn{2}{c}{\color{cbScarlet}G} & \multicolumn{2}{c}{\color{black}T\textsuperscript{$h$}} & \multicolumn{2}{c}{\color{black}T\textsuperscript{$\ell$}} & \multicolumn{2}{c}{\color{black}R} & & 
\multicolumn{2}{c}{\color{cbScarlet}G} & \multicolumn{2}{c}{\color{black}T\textsuperscript{$h$}} & \multicolumn{2}{c}{\color{black}T\textsuperscript{$\ell$}} & \multicolumn{2}{c}{\color{black}R} & & 
\multicolumn{2}{c}{\color{cbScarlet}G} & \multicolumn{2}{c}{\color{black}T\textsuperscript{$h$}} & \multicolumn{2}{c}{\color{black}T\textsuperscript{$\ell$}} & \multicolumn{2}{c}{\color{black}R} & & 
\multicolumn{2}{c}{\color{cbScarlet}G} & \multicolumn{2}{c}{\color{black}T\textsuperscript{$h$}} & \multicolumn{2}{c}{\color{black}T\textsuperscript{$\ell$}} & \multicolumn{2}{c}{\color{black}R} & \multicolumn{2}{c}{D}\\
\midrule

Air &
\multicolumn{2}{c}{\ssW{}22.6\ssW{}}&
\ssW{}25.3&\wtC{cbScarlet}&
\ssW{}25.0&\wtC{cbScarlet}&
\ssW{}33.0&\wtC{cbScarlet}&
&
\multicolumn{2}{c}{\ssW{}20.6\ssW{}}&
\ssW{}22.4&\wtC{cbScarlet}&
\ssW{}20.8&\ssW{}&
\ssW{}35.1&\wtC{cbScarlet}&
&
\multicolumn{2}{c}{\ssW{}20.7\ssW{}}&
\ssW{}23.3&\wtC{cbScarlet}&
\ssW{}21.3&\wtC{cbScarlet}&
\ssW{}34.3&\wtC{cbScarlet}&
&
\multicolumn{2}{c}{\ssW{}20.8\ssW{}}&
\ssW{}24.5&\wtC{cbScarlet}&
\ssW{}20.1&\ssW{}&
\ssW{}34.2&\wtC{cbScarlet}&
\ssW{}31.1&\wtC{cbScarlet}
\\ 
Bos &
\multicolumn{2}{c}{\ssW{}21.1\ssW{}}&
\ssW{}19.5&\ssW{}&
\ssW{}21.9&\ssW{}&
\ssW{}22.1&\ssW{}&
&
\multicolumn{2}{c}{\ssW{}16.5\ssW{}}&
\ssW{}16.8&\wtC{cbScarlet}&
\ssW{}15.7&\ssW{}&
\ssW{}19.7&\wtC{cbScarlet}&
&
\multicolumn{2}{c}{\ssW{}16.3\ssW{}}&
\ssW{}17.6&\wtC{cbScarlet}&
\ssW{}15.4&\btC{cbScarlet}&
\ssW{}18.9&\wtC{cbScarlet}&
&
\multicolumn{2}{c}{\ssW{}16.2\ssW{}}&
\ssW{}18.5&\wtC{cbScarlet}&
\ssW{}16.7&\wtC{cbScarlet}&
\ssW{}21.2&\wtC{cbScarlet}&
\ssW{}22.9&\wtC{cbScarlet}
\\ 
Con &
\multicolumn{2}{c}{\ssW{}16.6\ssW{}}&
\ssW{}17.4&\wtC{cbScarlet}&
\ssW{}16.6&\ssW{}&
\ssW{}18.5&\wtC{cbScarlet}&
&
\multicolumn{2}{c}{\ssW{}15.2\ssW{}}&
\ssW{}16.1&\wtC{cbScarlet}&
\ssW{}15.7&\wtC{cbScarlet}&
\ssW{}18.5&\wtC{cbScarlet}&
&
\multicolumn{2}{c}{\ssW{}15.5\ssW{}}&
\ssW{}16.5&\wtC{cbScarlet}&
\ssW{}15.6&\wtC{cbScarlet}&
\ssW{}19.6&\wtC{cbScarlet}&
&
\multicolumn{2}{c}{\ssW{}15.3\ssW{}}&
\ssW{}16.3&\wtC{cbScarlet}&
\ssW{}15.9&\wtC{cbScarlet}&
\ssW{}19.0&\wtC{cbScarlet}&
\ssW{}22.7&\wtC{cbScarlet}
\\ 
Dow &
\multicolumn{2}{c}{\ssW{}18.6\ssW{}}&
\ssW{}19.0&\ssW{}&
\ssW{}18.8&\wtC{cbScarlet}&
\ssW{}21.7&\wtC{cbScarlet}&
&
\multicolumn{2}{c}{\ssW{}17.4\ssW{}}&
\ssW{}17.8&\wtC{cbScarlet}&
\ssW{}16.7&\btC{cbScarlet}&
\ssW{}24.1&\wtC{cbScarlet}&
&
\multicolumn{2}{c}{\ssW{}17.7\ssW{}}&
\ssW{}18.2&\wtC{cbScarlet}&
\ssW{}17.0&\btC{cbScarlet}&
\ssW{}24.3&\wtC{cbScarlet}&
&
\multicolumn{2}{c}{\ssW{}17.8\ssW{}}&
\ssW{}19.8&\ssW{}&
\ssW{}17.6&\btC{cbScarlet}&
\ssW{}22.4&\wtC{cbScarlet}&
\ssW{}30.6&\wtC{cbScarlet}
\\ 
EnC &
\multicolumn{2}{c}{\ssW{}4.66\ssW{}}&
\ssW{}5.15&\wtC{cbScarlet}&
\ssW{}4.26&\btC{cbScarlet}&
\ssW{}5.55&\wtC{cbScarlet}&
&
\multicolumn{2}{c}{\ssW{}3.67\ssW{}}&
\ssW{}4.37&\wtC{cbScarlet}&
\ssW{}4.14&\wtC{cbScarlet}&
\ssW{}6.92&\wtC{cbScarlet}&
&
\multicolumn{2}{c}{\ssW{}3.85\ssW{}}&
\ssW{}4.50&\wtC{cbScarlet}&
\ssW{}4.02&\wtC{cbScarlet}&
\ssW{}7.08&\wtC{cbScarlet}&
&
\multicolumn{2}{c}{\ssW{}3.76\ssW{}}&
\ssW{}4.88&\wtC{cbScarlet}&
\ssW{}3.99&\wtC{cbScarlet}&
\ssW{}6.78&\wtC{cbScarlet}&
\ssW{}4.23&\wtC{cbScarlet}
\\ 
EnH &
\multicolumn{2}{c}{\ssW{}1.65\ssW{}}&
\ssW{}1.52&\wtC{cbScarlet}&
\ssW{}1.13&\btC{cbScarlet}&
\ssW{}2.63&\wtC{cbScarlet}&
&
\multicolumn{2}{c}{\ssW{}0.69\ssW{}}&
\ssW{}1.54&\wtC{cbScarlet}&
\ssW{}0.84&\wtC{cbScarlet}&
\ssW{}4.02&\wtC{cbScarlet}&
&
\multicolumn{2}{c}{\ssW{}0.92\ssW{}}&
\ssW{}1.78&\wtC{cbScarlet}&
\ssW{}1.02&\wtC{cbScarlet}&
\ssW{}3.81&\wtC{cbScarlet}&
&
\multicolumn{2}{c}{\ssW{}0.87\ssW{}}&
\ssW{}1.78&\wtC{cbScarlet}&
\ssW{}0.80&\ssW{}&
\ssW{}3.68&\wtC{cbScarlet}&
\ssW{}0.43&\btC{cbScarlet}
\\ 
Tow &
\multicolumn{2}{c}{\ssW{}11.5\ssW{}}&
\ssW{}11.7&\wtC{cbScarlet}&
\ssW{}11.7&\wtC{cbScarlet}&
\ssW{}15.7&\wtC{cbScarlet}&
&
\multicolumn{2}{c}{\ssW{}11.3\ssW{}}&
\ssW{}10.9&\ssW{}&
\ssW{}11.1&\ssW{}&
\ssW{}16.1&\wtC{cbScarlet}&
&
\multicolumn{2}{c}{\ssW{}11.4\ssW{}}&
\ssW{}11.3&\ssW{}&
\ssW{}10.9&\btC{cbScarlet}&
\ssW{}17.0&\wtC{cbScarlet}&
&
\multicolumn{2}{c}{\ssW{}11.3\ssW{}}&
\ssW{}11.9&\wtC{cbScarlet}&
\ssW{}11.2&\btC{cbScarlet}&
\ssW{}16.6&\wtC{cbScarlet}&
\ssW{}11.2&\btC{cbScarlet}
\\ 
WiR &
\multicolumn{2}{c}{\ssW{}64.4\ssW{}}&
\ssW{}64.6&\btC{cbScarlet}&
\ssW{}65.2&\wtC{cbScarlet}&
\ssW{}64.3&\btC{cbScarlet}&
&
\multicolumn{2}{c}{\ssW{}63.0\ssW{}}&
\ssW{}62.4&\ssW{}&
\ssW{}62.5&\ssW{}&
\ssW{}63.8&\wtC{cbScarlet}&
&
\multicolumn{2}{c}{\ssW{}62.3\ssW{}}&
\ssW{}62.9&\ssW{}&
\ssW{}62.8&\ssW{}&
\ssW{}64.6&\wtC{cbScarlet}&
&
\multicolumn{2}{c}{\ssW{}62.7\ssW{}}&
\ssW{}62.9&\wtC{cbScarlet}&
\ssW{}62.5&\wtC{cbScarlet}&
\ssW{}64.5&\wtC{cbScarlet}&
\ssW{}71.7&\wtC{cbScarlet}
\\ 
WiW &
\multicolumn{2}{c}{\ssW{}70.1\ssW{}}&
\ssW{}70.1&\ssW{}&
\ssW{}68.8&\btC{cbScarlet}&
\ssW{}70.9&\wtC{cbScarlet}&
&
\multicolumn{2}{c}{\ssW{}69.2\ssW{}}&
\ssW{}69.2&\ssW{}&
\ssW{}68.7&\btC{cbScarlet}&
\ssW{}71.1&\wtC{cbScarlet}&
&
\multicolumn{2}{c}{\ssW{}68.9\ssW{}}&
\ssW{}69.3&\wtC{cbScarlet}&
\ssW{}69.4&\ssW{}&
\ssW{}71.6&\wtC{cbScarlet}&
&
\multicolumn{2}{c}{\ssW{}69.1\ssW{}}&
\ssW{}69.7&\wtC{cbScarlet}&
\ssW{}69.6&\ssW{}&
\ssW{}71.4&\wtC{cbScarlet}&
\ssW{}72.2&\wtC{cbScarlet}
\\ 
Yac &
\multicolumn{2}{c}{\ssW{}0.46\ssW{}}&
\ssW{}0.45&\ssW{}&
\ssW{}0.37&\btC{cbScarlet}&
\ssW{}0.46&\ssW{}&
&
\multicolumn{2}{c}{\ssW{}0.32\ssW{}}&
\ssW{}0.38&\wtC{cbScarlet}&
\ssW{}0.33&\ssW{}&
\ssW{}0.40&\wtC{cbScarlet}&
&
\multicolumn{2}{c}{\ssW{}0.32\ssW{}}&
\ssW{}0.39&\wtC{cbScarlet}&
\ssW{}0.33&\ssW{}&
\ssW{}0.44&\wtC{cbScarlet}&
&
\multicolumn{2}{c}{\ssW{}0.33\ssW{}}&
\ssW{}0.40&\wtC{cbScarlet}&
\ssW{}0.33&\btC{cbScarlet}&
\ssW{}0.48&\wtC{cbScarlet}&
\ssW{}0.88&\wtC{cbScarlet}
\\ 
\midrule 
B/W &
\multicolumn{2}{c}{---}&
\multicolumn{2}{c}{1/5}&
\multicolumn{2}{c}{4/4}&
\multicolumn{2}{c}{1/7}&
&
\multicolumn{2}{c}{---}&
\multicolumn{2}{c}{0/7}&
\multicolumn{2}{c}{2/3}&
\multicolumn{2}{c}{0/10}&
&
\multicolumn{2}{c}{---}&
\multicolumn{2}{c}{0/8}&
\multicolumn{2}{c}{3/4}&
\multicolumn{2}{c}{0/10}&
&
\multicolumn{2}{c}{---}&
\multicolumn{2}{c}{0/9}&
\multicolumn{2}{c}{3/4}&
\multicolumn{2}{c}{0/10}&
\multicolumn{2}{c}{2/8}\\

\midrule 

\multicolumn{38}{c}{\small \textbf{Test $\ell = 63$}} \\ 
 & \multicolumn{8}{c}{$n^\text{pop}=1000$} & \ssW{}\ssW{} & \multicolumn{8}{c}{IMS $g=4$} & \ssW{}\ssW{} & \multicolumn{8}{c}{IMS $g=6$} & \ssW{}\ssW{} & \multicolumn{9}{c}{IMS $g=8$} \\
 \ssW{}& 
\multicolumn{2}{c}{\color{cbScarlet}G} & \multicolumn{2}{c}{\color{black}T\textsuperscript{$h$}} & \multicolumn{2}{c}{\color{black}T\textsuperscript{$\ell$}} & \multicolumn{2}{c}{\color{black}R} & & 
\multicolumn{2}{c}{\color{cbScarlet}G} & \multicolumn{2}{c}{\color{black}T\textsuperscript{$h$}} & \multicolumn{2}{c}{\color{black}T\textsuperscript{$\ell$}} & \multicolumn{2}{c}{\color{black}R} & & 
\multicolumn{2}{c}{\color{cbScarlet}G} & \multicolumn{2}{c}{\color{black}T\textsuperscript{$h$}} & \multicolumn{2}{c}{\color{black}T\textsuperscript{$\ell$}} & \multicolumn{2}{c}{\color{black}R} & & 
\multicolumn{2}{c}{\color{cbScarlet}G} & \multicolumn{2}{c}{\color{black}T\textsuperscript{$h$}} & \multicolumn{2}{c}{\color{black}T\textsuperscript{$\ell$}} & \multicolumn{2}{c}{\color{black}R} & \multicolumn{2}{c}{D}\\
\midrule

Air &
\multicolumn{2}{c}{\ssW{}23.0\ssW{}}&
\ssW{}25.5&\wtC{cbScarlet}&
\ssW{}25.9&\wtC{cbScarlet}&
\ssW{}31.5&\wtC{cbScarlet}&
&
\multicolumn{2}{c}{\ssW{}21.1\ssW{}}&
\ssW{}22.5&\wtC{cbScarlet}&
\ssW{}19.6&\ssW{}&
\ssW{}34.9&\wtC{cbScarlet}&
&
\multicolumn{2}{c}{\ssW{}21.7\ssW{}}&
\ssW{}22.4&\wtC{cbScarlet}&
\ssW{}21.9&\wtC{cbScarlet}&
\ssW{}33.8&\wtC{cbScarlet}&
&
\multicolumn{2}{c}{\ssW{}21.2\ssW{}}&
\ssW{}23.4&\wtC{cbScarlet}&
\ssW{}21.6&\ssW{}&
\ssW{}34.1&\wtC{cbScarlet}&
\ssW{}30.8&\wtC{cbScarlet}
\\ 
Bos &
\multicolumn{2}{c}{\ssW{}22.0\ssW{}}&
\ssW{}20.0&\ssW{}&
\ssW{}21.2&\ssW{}&
\ssW{}21.9&\ssW{}&
&
\multicolumn{2}{c}{\ssW{}19.2\ssW{}}&
\ssW{}20.7&\wtC{cbScarlet}&
\ssW{}20.4&\wtC{cbScarlet}&
\ssW{}24.1&\wtC{cbScarlet}&
&
\multicolumn{2}{c}{\ssW{}21.5\ssW{}}&
\ssW{}20.3&\ssW{}&
\ssW{}20.3&\btC{cbScarlet}&
\ssW{}25.0&\wtC{cbScarlet}&
&
\multicolumn{2}{c}{\ssW{}19.8\ssW{}}&
\ssW{}19.7&\wtC{cbScarlet}&
\ssW{}21.4&\wtC{cbScarlet}&
\ssW{}25.8&\wtC{cbScarlet}&
\ssW{}26.1&\wtC{cbScarlet}
\\ 
Con &
\multicolumn{2}{c}{\ssW{}15.9\ssW{}}&
\ssW{}17.1&\wtC{cbScarlet}&
\ssW{}16.5&\wtC{cbScarlet}&
\ssW{}18.3&\wtC{cbScarlet}&
&
\multicolumn{2}{c}{\ssW{}15.3\ssW{}}&
\ssW{}16.2&\wtC{cbScarlet}&
\ssW{}15.5&\ssW{}&
\ssW{}19.1&\wtC{cbScarlet}&
&
\multicolumn{2}{c}{\ssW{}15.3\ssW{}}&
\ssW{}16.3&\wtC{cbScarlet}&
\ssW{}15.8&\wtC{cbScarlet}&
\ssW{}19.9&\wtC{cbScarlet}&
&
\multicolumn{2}{c}{\ssW{}15.3\ssW{}}&
\ssW{}16.6&\wtC{cbScarlet}&
\ssW{}16.1&\wtC{cbScarlet}&
\ssW{}18.9&\wtC{cbScarlet}&
\ssW{}21.3&\wtC{cbScarlet}
\\ 
Dow &
\multicolumn{2}{c}{\ssW{}18.3\ssW{}}&
\ssW{}18.6&\wtC{cbScarlet}&
\ssW{}17.4&\btC{cbScarlet}&
\ssW{}22.3&\wtC{cbScarlet}&
&
\multicolumn{2}{c}{\ssW{}17.5\ssW{}}&
\ssW{}17.9&\ssW{}&
\ssW{}17.0&\btC{cbScarlet}&
\ssW{}23.7&\wtC{cbScarlet}&
&
\multicolumn{2}{c}{\ssW{}17.6\ssW{}}&
\ssW{}18.2&\wtC{cbScarlet}&
\ssW{}17.2&\btC{cbScarlet}&
\ssW{}24.6&\wtC{cbScarlet}&
&
\multicolumn{2}{c}{\ssW{}17.7\ssW{}}&
\ssW{}18.2&\ssW{}&
\ssW{}17.9&\ssW{}&
\ssW{}22.6&\wtC{cbScarlet}&
\ssW{}28.0&\wtC{cbScarlet}
\\ 
EnC &
\multicolumn{2}{c}{\ssW{}4.49\ssW{}}&
\ssW{}4.70&\wtC{cbScarlet}&
\ssW{}4.24&\btC{cbScarlet}&
\ssW{}5.63&\wtC{cbScarlet}&
&
\multicolumn{2}{c}{\ssW{}3.77\ssW{}}&
\ssW{}4.37&\wtC{cbScarlet}&
\ssW{}3.99&\wtC{cbScarlet}&
\ssW{}6.94&\wtC{cbScarlet}&
&
\multicolumn{2}{c}{\ssW{}3.93\ssW{}}&
\ssW{}4.42&\wtC{cbScarlet}&
\ssW{}3.95&\wtC{cbScarlet}&
\ssW{}7.42&\wtC{cbScarlet}&
&
\multicolumn{2}{c}{\ssW{}3.95\ssW{}}&
\ssW{}4.85&\wtC{cbScarlet}&
\ssW{}4.20&\wtC{cbScarlet}&
\ssW{}7.37&\wtC{cbScarlet}&
\ssW{}4.47&\wtC{cbScarlet}
\\ 
EnH &
\multicolumn{2}{c}{\ssW{}1.60\ssW{}}&
\ssW{}1.59&\wtC{cbScarlet}&
\ssW{}1.12&\btC{cbScarlet}&
\ssW{}2.74&\wtC{cbScarlet}&
&
\multicolumn{2}{c}{\ssW{}0.80\ssW{}}&
\ssW{}1.52&\wtC{cbScarlet}&
\ssW{}0.89&\wtC{cbScarlet}&
\ssW{}3.73&\wtC{cbScarlet}&
&
\multicolumn{2}{c}{\ssW{}0.88\ssW{}}&
\ssW{}1.67&\wtC{cbScarlet}&
\ssW{}0.94&\ssW{}&
\ssW{}4.12&\wtC{cbScarlet}&
&
\multicolumn{2}{c}{\ssW{}0.89\ssW{}}&
\ssW{}1.92&\wtC{cbScarlet}&
\ssW{}0.93&\wtC{cbScarlet}&
\ssW{}3.71&\wtC{cbScarlet}&
\ssW{}0.33&\btC{cbScarlet}
\\ 
Tow &
\multicolumn{2}{c}{\ssW{}11.6\ssW{}}&
\ssW{}12.2&\wtC{cbScarlet}&
\ssW{}12.1&\wtC{cbScarlet}&
\ssW{}15.9&\wtC{cbScarlet}&
&
\multicolumn{2}{c}{\ssW{}11.5\ssW{}}&
\ssW{}11.4&\ssW{}&
\ssW{}11.4&\ssW{}&
\ssW{}16.8&\wtC{cbScarlet}&
&
\multicolumn{2}{c}{\ssW{}11.6\ssW{}}&
\ssW{}11.5&\ssW{}&
\ssW{}11.2&\btC{cbScarlet}&
\ssW{}16.7&\wtC{cbScarlet}&
&
\multicolumn{2}{c}{\ssW{}11.4\ssW{}}&
\ssW{}12.2&\wtC{cbScarlet}&
\ssW{}11.4&\ssW{}&
\ssW{}17.1&\wtC{cbScarlet}&
\ssW{}11.2&\btC{cbScarlet}
\\ 
WiR &
\multicolumn{2}{c}{\ssW{}63.1\ssW{}}&
\ssW{}63.0&\ssW{}&
\ssW{}64.4&\wtC{cbScarlet}&
\ssW{}62.9&\ssW{}&
&
\multicolumn{2}{c}{\ssW{}62.9\ssW{}}&
\ssW{}62.5&\wtC{cbScarlet}&
\ssW{}61.7&\ssW{}&
\ssW{}62.5&\wtC{cbScarlet}&
&
\multicolumn{2}{c}{\ssW{}62.5\ssW{}}&
\ssW{}63.0&\ssW{}&
\ssW{}62.3&\ssW{}&
\ssW{}63.6&\wtC{cbScarlet}&
&
\multicolumn{2}{c}{\ssW{}62.7\ssW{}}&
\ssW{}63.0&\ssW{}&
\ssW{}61.8&\btC{cbScarlet}&
\ssW{}63.2&\wtC{cbScarlet}&
\ssW{}72.6&\wtC{cbScarlet}
\\ 
WiW &
\multicolumn{2}{c}{\ssW{}68.7\ssW{}}&
\ssW{}69.0&\ssW{}&
\ssW{}68.0&\btC{cbScarlet}&
\ssW{}69.9&\wtC{cbScarlet}&
&
\multicolumn{2}{c}{\ssW{}68.3\ssW{}}&
\ssW{}68.6&\ssW{}&
\ssW{}68.3&\btC{cbScarlet}&
\ssW{}70.2&\wtC{cbScarlet}&
&
\multicolumn{2}{c}{\ssW{}69.1\ssW{}}&
\ssW{}69.4&\ssW{}&
\ssW{}68.2&\btC{cbScarlet}&
\ssW{}70.6&\wtC{cbScarlet}&
&
\multicolumn{2}{c}{\ssW{}68.2\ssW{}}&
\ssW{}69.3&\wtC{cbScarlet}&
\ssW{}69.0&\ssW{}&
\ssW{}70.3&\wtC{cbScarlet}&
\ssW{}72.2&\wtC{cbScarlet}
\\ 
Yac &
\multicolumn{2}{c}{\ssW{}0.44\ssW{}}&
\ssW{}0.46&\ssW{}&
\ssW{}0.40&\btC{cbScarlet}&
\ssW{}0.46&\ssW{}&
&
\multicolumn{2}{c}{\ssW{}0.41\ssW{}}&
\ssW{}0.49&\wtC{cbScarlet}&
\ssW{}0.40&\btC{cbScarlet}&
\ssW{}0.45&\wtC{cbScarlet}&
&
\multicolumn{2}{c}{\ssW{}0.41\ssW{}}&
\ssW{}0.45&\wtC{cbScarlet}&
\ssW{}0.42&\ssW{}&
\ssW{}0.52&\wtC{cbScarlet}&
&
\multicolumn{2}{c}{\ssW{}0.46\ssW{}}&
\ssW{}0.46&\ssW{}&
\ssW{}0.44&\btC{cbScarlet}&
\ssW{}0.50&\wtC{cbScarlet}&
\ssW{}0.91&\wtC{cbScarlet}
\\ 
\midrule 
B/W &
\multicolumn{2}{c}{---}&
\multicolumn{2}{c}{0/6}&
\multicolumn{2}{c}{5/4}&
\multicolumn{2}{c}{0/7}&
&
\multicolumn{2}{c}{---}&
\multicolumn{2}{c}{0/7}&
\multicolumn{2}{c}{3/3}&
\multicolumn{2}{c}{0/10}&
&
\multicolumn{2}{c}{---}&
\multicolumn{2}{c}{0/6}&
\multicolumn{2}{c}{4/3}&
\multicolumn{2}{c}{0/10}&
&
\multicolumn{2}{c}{---}&
\multicolumn{2}{c}{0/7}&
\multicolumn{2}{c}{2/4}&
\multicolumn{2}{c}{0/10}&
\multicolumn{2}{c}{2/8}\\

\bottomrule

\end{tabular}
}
\vspace{-3mm}
\end{table*}

\section{Discussion \& Conclusion}\label{sec:discussion}
We built upon previous work on model-based GP, in particular on GP-GOMEA, to find accurate solutions when a strict limitation on their size is imposed, in the domain of SR. 
We focused on small solutions, in particular much smaller solutions than typically reported in literature, to prevent solutions becoming too large to be (easily) interpretable, a key reason to justify the use of GP in many practical applications.

A first limitation of this work is that to truly \emph{achieve} interpretability may well require different measures. Interpretation is mostly subjective, and many other factors besides solution size are important, including the intuitiveness of the subfunctions composing the solution, potential decompositions into understandable repeating sub-modules, the number of features considered, and the meaning of these features~\citep{lipton2018mythos,doshi2017towards}. Nonetheless, much current research on GP for SR is far from delivering any interpretable results precisely because the size of solutions is far too large (see, e.g., the work of~\citep{martins2018solving}).

We considered solution sizes up to $\ell=63$ (corresponding to $h=5$ for GP-GOMEA with subfunctions of arity $\leq 2$). In our opinion, the limit of $\ell=31$ ($h=4$) is particularly interesting, as interpreting some solutions at this level can already be non-trivial at times. For example, we show the (manually simplified) best test solution found by GP-GOMEA (IMS $g=8$) for Tower and Yacht, i.e. the biggest and smallest dataset respectively, in Figure~\ref{fig:bestsols}. The solution for Tower is arguably easier to understand than the one for Yacht. We found solutions with $\ell=63$ ($h=5$) to be overly long to attempt interpreting, and solutions with $\ell=15$ ($h=3$) to be mostly readable and understandable. We report other example solutions at: \url{http://bit.ly/2IrUFyQ}.

We believe future work should address the aforementioned limitation: effort should be put towards reaching some form of interpretability notions, that go beyond solution size or other custom metrics (e.g.,~\citep{vladislavleva2009order}). User studies involving the end users of the model (e.g., medical doctors for a diagnosis model) could guide the design of notions of interpretability.
If an objective that represents interpretability can be defined, the design of multi-objective (model-based) GP algorithms may lead to very interesting results.

Another limitation of this work lies in the fact that we did not study how linkage learning behaves in GP for SR in depth. In fact, it would be interesting to assess when linkage learning is beneficial, and when it is superfluous or harmful. To this end, a regime of experiments where linkage-related outcomes are predefined, such as emergence of specific patterns, needs to be designed. Simple problems where the true function to regress is known may need to be considered.
Studies of this kind could provide more insights on how to improve linkage learning in GP for SR (and other learning tasks), and are an interesting direction for future work.

Another crucial point to base future research upon is enabling linkage learning and linkage-based mixing in GP with trees of arbitrary shape. In fact, GP-GOMEA was not found to be markedly better than GP-Trad\textsuperscript{$\ell$}, and a large performance gap was found between GP-Trad\textsuperscript{$\ell$} and GP-Trad\textsuperscript{$h$}. This is indicative that there is added value to perform evolution directly on non-templated trees, which, from this perspective, may be considered a limitation of GP-GOMEA. Going beyond the use of a fixed tree template, while still enabling linkage identification and exploitation, is a challenging open problem that could bring very rewarding results. On the other hand, we believe it is interesting to see that when GP-GOMEA and GP-Trad are set to work on the same search space, i.e., when GP-Trad\textsuperscript{$h$} is used, then GP-GOMEA performs markedly better.

In summary and conclusion, we have identified limits and presented ways to improve a key component of a state-of-the-art model-based EA, i.e. \emph{GP-GOMEA}, to competently deal with realistic SR datasets, when small solutions are desired. This key component is linkage learning. We showed that solely and directly relying on mutual information to identify linkage may be undesirable, because the genotype is not uniformly distributed in GP populations, and we provided an approximate biasing method to tackle this problem. We furthermore explored how to incorporate ERCs into linkage learning, and found that on-line binning of constants is an efficient and effective strategy. Lastly, we introduced a new form of the IMS, to relieve practitioners from setting a population size, and from finding a good generalizing solution. Ultimately, our contributions proved successful in improving the performance of GP-GOMEA, leading to the best overall performance against competing EAs, as well as tuned decision trees. We believe our findings set an important first step for the design of better model-based GP algorithms capable of learning interpretable solutions in real-world data.

\begin{figure}
\centering
\scalebox{0.7}{
\parbox{\linewidth}{
\begin{boxedalign*}
& \textbf{Tower: } \\
& 4668.49 -3.56 ( (662.77+x_{21}) x_{12} \div_\text{AQ} x_{16} -x_{1} - x_{15} + x_{5} + 4x_{12} - x_{23} ( x_{6} \div_\text{AQ} x_{1} + 1 ) ) \\
& \textbf{Yacht: } \\
& 0.73 + 33004.40 \left( ( ( x^2_6 \div_\text{AQ} (x_5 x_2) ) \div_\text{AQ} ( x_3 x_2 \div_\text{AQ} ( x_2 \div_\text{AQ} x_1 ) ) ) (x_6+0.30) x^5_6 x_5 \right)
\\
\end{boxedalign*}
}
}
\vspace{-0.5cm}
\caption{Examples of best solution found by GP-GOMEA ($\ell = 31$, IMS $g=8$).}\label{fig:bestsols}
\vspace{-0.5cm}
\end{figure}

\section*{Acknowledgment}
The authors thank the Foundation Kinderen Kankervrij for financial support (project no. 187), and SURFsara for granting access to the Lisa Compute Cluster.

\small

\bibliographystyle{apalike}
\bibliography{gomea_regression}

\begin{thebibliography}{}

\bibitem[Asuncion and Newman, 2007]{asuncion2007uci}
Asuncion, A. and Newman, D. (2007).
\newblock {UCI} machine learning repository.

\bibitem[Bosman and De~Jong, 2004]{bosman2004learning}
Bosman, P. A.~N. and De~Jong, E.~D. (2004).
\newblock Learning probabilistic tree grammars for genetic programming.
\newblock In {\em International Conference on Parallel Problem Solving from
  Nature (PPSN) 2004}, pages 192--201. Springer.

\bibitem[Bouter et~al., 2017]{bouter2017exploiting}
Bouter, A., Alderliesten, T., Witteveen, C., and Bosman, P. A.~N. (2017).
\newblock Exploiting linkage information in real-valued optimization with the
  real-valued gene-pool optimal mixing evolutionary algorithm.
\newblock In {\em Genetic and Evolutionary Computation Conference (GECCO)
  2017}, pages 705--712. ACM.

\bibitem[Breiman, 2001]{breiman2001random}
Breiman, L. (2001).
\newblock Random forests.
\newblock {\em Machine learning}, 45(1):5--32.

\bibitem[Breiman et~al., 1984]{breiman1984classification}
Breiman, L., Friedman, J., Olshen, R.~A., and Stone, C.~J. (1984).
\newblock {\em Classification and regression trees}.
\newblock Wadsworth and Brooks.

\bibitem[Chen et~al., 2015]{chen2015generalisation}
Chen, Q., Xue, B., and Zhang, M. (2015).
\newblock Generalisation and domain adaptation in gp with gradient descent for
  symbolic regression.
\newblock In {\em IEEE Congress on Evolutionary Computation (CEC) 2015}, pages
  1137--1144. IEEE.

\bibitem[Chen et~al., 2018]{chen2018improving}
Chen, Q., Xue, B., and Zhang, M. (2018).
\newblock Improving generalization of genetic programming for symbolic
  regression with angle-driven geometric semantic operators.
\newblock {\em IEEE Transactions on Evolutionary Computation}, 23(3):488--502.

\bibitem[Chen and Guestrin, 2016]{chen2016xgboost}
Chen, T. and Guestrin, C. (2016).
\newblock {XGB}oost: A scalable tree boosting system.
\newblock In {\em Proceedings of the 22nd acm sigkdd international conference
  on knowledge discovery and data mining}, pages 785--794. ACM.

\bibitem[Chen et~al., 2007]{chen2007survey}
Chen, Y., Yu, T.-L., Sastry, K., and Goldberg, D.~E. (2007).
\newblock A survey of linkage learning techniques in genetic and evolutionary
  algorithms.
\newblock {\em IlliGAL report}, 2007014.

\bibitem[De~Melo, 2014]{de2014kaizen}
De~Melo, V.~V. (2014).
\newblock Kaizen programming.
\newblock In {\em Genetic and Evolutionary Computation Conference (GECCO)
  2014}, pages 895--902. ACM.

\bibitem[Dem{\v{s}}ar, 2006]{demvsar2006statistical}
Dem{\v{s}}ar, J. (2006).
\newblock Statistical comparisons of classifiers over multiple data sets.
\newblock {\em Journal of Machine learning research}, 7(Jan):1--30.

\bibitem[Doshi-Velez and Kim, 2017]{doshi2017towards}
Doshi-Velez, F. and Kim, B. (2017).
\newblock Towards a rigorous science of interpretable machine learning.
\newblock {\em arXiv preprint arXiv:1702.08608}.

\bibitem[Ebner et~al., 2001]{ebner2001neutral}
Ebner, M., Shackleton, M., and Shipman, R. (2001).
\newblock How neutral networks influence evolvability.
\newblock {\em Complexity}, 7(2):19--33.

\bibitem[Gathercole and Ross, 1996]{gathercole1996adverse}
Gathercole, C. and Ross, P. (1996).
\newblock An adverse interaction between crossover and restricted tree depth in
  genetic programming.
\newblock In {\em Genetic and Evolutionary Computation Conference (GECCO)
  1996}, pages 291--296. MIT Press.

\bibitem[Goldman and Punch, 2014]{goldman2014parameter}
Goldman, B.~W. and Punch, W.~F. (2014).
\newblock Parameter-less population pyramid.
\newblock In {\em Genetic and Evolutionary Computation Conference (GECCO)
  2014}, pages 785--792. ACM.

\bibitem[Gronau and Moran, 2007]{gronau2007optimal}
Gronau, I. and Moran, S. (2007).
\newblock Optimal implementations of upgma and other common clustering
  algorithms.
\newblock {\em Information Processing Letters}, 104(6):205--210.

\bibitem[Guidotti et~al., 2018]{guidotti2018survey}
Guidotti, R., Monreale, A., Ruggieri, S., Turini, F., Giannotti, F., and
  Pedreschi, D. (2018).
\newblock A survey of methods for explaining black box models.
\newblock {\em ACM computing surveys (CSUR)}, 51(5):93.

\bibitem[Harik et~al., 1999]{harik1999gambler}
Harik, G., Cant{\'u}-Paz, E., Goldberg, D.~E., and Miller, B.~L. (1999).
\newblock The gambler's ruin problem, genetic algorithms, and the sizing of
  populations.
\newblock {\em Evolutionary Computation}, 7(3):231--253.

\bibitem[Harik and Lobo, 1999]{harik1999parameterless}
Harik, G.~R. and Lobo, F.~G. (1999).
\newblock A parameter-less genetic algorithm.
\newblock In {\em Genetic and Evolutionary Computation Conference (GECCO)
  1999}, pages 258--265. Morgan Kaufmann Publishers Inc.

\bibitem[Hasegawa and Iba, 2009]{hasegawa2009latent}
Hasegawa, Y. and Iba, H. (2009).
\newblock Latent variable model for estimation of distribution algorithm based
  on a probabilistic context-free grammar.
\newblock {\em IEEE Transactions on Evolutionary Computation}, 13(4):858--878.

\bibitem[Hauschild and Pelikan, 2011]{hauschild2011introduction}
Hauschild, M. and Pelikan, M. (2011).
\newblock An introduction and survey of estimation of distribution algorithms.
\newblock {\em Swarm and evolutionary computation}, 1(3):111--128.

\bibitem[Hemberg et~al., 2012]{hemberg2012investigation}
Hemberg, E., Veeramachaneni, K., McDermott, J., Berzan, C., and O'Reilly, U.-M.
  (2012).
\newblock An investigation of local patterns for estimation of distribution
  genetic programming.
\newblock In {\em Genetic and Evolutionary Computation Conference (GECCO)
  2012}, pages 767--774. ACM.

\bibitem[Hsu and Yu, 2015]{hsu2015optimization}
Hsu, S.-H. and Yu, T.-L. (2015).
\newblock Optimization by pairwise linkage detection, incremental linkage set,
  and restricted/back mixing: {DSMGA}-{II}.
\newblock In {\em Genetic and Evolutionary Computation Conference (GECCO)
  2015}, pages 519--526. ACM.

\bibitem[Icke and Bongard, 2013]{icke2013improving}
Icke, I. and Bongard, J.~C. (2013).
\newblock Improving genetic programming based symbolic regression using
  deterministic machine learning.
\newblock In {\em IEEE Congress on Evolutionary Computation (CEC) 2013}, pages
  1763--1770. IEEE.

\bibitem[Keijzer, 2003]{keijzer2003improving}
Keijzer, M. (2003).
\newblock Improving symbolic regression with interval arithmetic and linear
  scaling.
\newblock In {\em European Conference on Genetic Programming}, pages 70--82.
  Springer.

\bibitem[Kim et~al., 2014]{kim2014probabilistic}
Kim, K., Shan, Y., Nguyen, X.~H., and McKay, R.~I. (2014).
\newblock Probabilistic model building in genetic programming: A critical
  review.
\newblock {\em Genetic Programming and Evolvable Machines}, 15(2):115--167.

\bibitem[Koza, 1992]{koza1992gp}
Koza, J.~R. (1992).
\newblock {\em Genetic Programming: On the programming of computers by means of
  natural selection}.
\newblock MIT Press, Cambridge, MA, USA.

\bibitem[Krawiec, 2015]{krawiec2016behavioral}
Krawiec, K. (2015).
\newblock {\em Behavioral program synthesis with genetic programming}.
\newblock Springer.

\bibitem[Langdon and Poli, 1997]{langdon1997analysis}
Langdon, W.~B. and Poli, R. (1997).
\newblock An analysis of the max problem in genetic programming.
\newblock {\em Genetic Programming}, 1(997):222--230.

\bibitem[Li et~al., 2010]{li2010genetic}
Li, X., Mabu, S., Zhou, H., Shimada, K., and Hirasawa, K. (2010).
\newblock Genetic network programming with estimation of distribution
  algorithms for class association rule mining in traffic prediction.
\newblock In {\em IEEE Congress on Evolutionary Computation (CEC) 2010}, pages
  1--8. IEEE.

\bibitem[Lin and Yu, 2018]{lin2018investigation}
Lin, Y.-J. and Yu, T.-L. (2018).
\newblock Investigation of the exponential population scheme for genetic
  algorithms.
\newblock In {\em Genetic and Evolutionary Computation Conference (GECCO)
  2018}, pages 975--982. ACM.

\bibitem[Lipton, 2018]{lipton2018mythos}
Lipton, Z.~C. (2018).
\newblock The mythos of model interpretability.
\newblock {\em Queue}, 16(3):30:31--30:57.

\bibitem[Luke and Panait, 2001]{luke2001survey}
Luke, S. and Panait, L. (2001).
\newblock A survey and comparison of tree generation algorithms.
\newblock In {\em Genetic and Evolutionary Computation Conference (GECCO)
  2001}, pages 81--88. Morgan Kaufmann Publishers Inc.

\bibitem[Luong et~al., 2014]{luong2014multi}
Luong, N.~H., La~Poutr{\'e}, H., and Bosman, P. A.~N. (2014).
\newblock Multi-objective gene-pool optimal mixing evolutionary algorithms.
\newblock In {\em Genetic and Evolutionary Computation Conference (GECCO)
  2014}. ACM.

\bibitem[Martins et~al., 2018]{martins2018solving}
Martins, J. F. B.~S., Oliveira, L. O. V.~B., Miranda, L.~F., Casadei, F., and
  Pappa, G.~L. (2018).
\newblock Solving the exponential growth of symbolic regression trees in
  geometric semantic genetic programming.
\newblock In {\em Genetic and Evolutionary Computation Conference (GECCO)
  2018}, pages 1151--1158, New York, NY, USA. ACM.

\bibitem[Medvet et~al., 2018a]{medvet2018gomge}
Medvet, E., Bartoli, A., De~Lorenzo, A., and Tarlao, F. (2018a).
\newblock {GOMGE}: Gene-pool optimal mixing on grammatical evolution.
\newblock In {\em International Conference on Parallel Problem Solving from
  Nature (PPSN) 2018}, pages 223--235. Springer.

\bibitem[Medvet et~al., 2018b]{medvet2018unveiling}
Medvet, E., Virgolin, M., Castelli, M., Bosman, P. A.~N., Gon{\c{c}}alves, I.,
  and Tu{\v{s}}ar, T. (2018b).
\newblock Unveiling evolutionary algorithm representation with {DU} maps.
\newblock {\em Genetic Programming and Evolvable Machines}, 19(3):351--389.

\bibitem[Moraglio et~al., 2012]{moraglio2012geometric}
Moraglio, A., Krawiec, K., and Johnson, C.~G. (2012).
\newblock Geometric semantic genetic programming.
\newblock In {\em International Conference on Parallel Problem Solving from
  Nature (PPSN) 2012}, pages 21--31. Springer.

\bibitem[Ni et~al., 2013]{ni2013use}
Ni, J., Drieberg, R.~H., and Rockett, P.~I. (2013).
\newblock The use of an analytic quotient operator in genetic programming.
\newblock {\em IEEE Trans. Evol. Comput.}, 17(1):146--152.

\bibitem[Orzechowski et~al., 2018]{orzechowski2018where}
Orzechowski, P., La~Cava, W., and Moore, J.~H. (2018).
\newblock Where are we now?: A large benchmark study of recent symbolic
  regression methods.
\newblock In {\em Genetic and Evolutionary Computation Conference (GECCO)
  2018}, pages 1183--1190, New York, NY, USA. ACM.

\bibitem[Pawlak and Krawiec, 2018]{pawlak2018competent}
Pawlak, T.~P. and Krawiec, K. (2018).
\newblock Competent geometric semantic genetic programming for symbolic
  regression and {B}oolean function synthesis.
\newblock {\em Evolutionary Computation}, 26(2):177--212.

\bibitem[Pawlak et~al., 2015]{pawlak2015semantic}
Pawlak, T.~P., Wieloch, B., and Krawiec, K. (2015).
\newblock Semantic backpropagation for designing search operators in genetic
  programming.
\newblock {\em IEEE Trans. Evol. Comput.}, 19(3):326--340.

\bibitem[Pedregosa et~al., 2011]{pedregosa2011scikit}
Pedregosa, F., Varoquaux, G., Gramfort, A., Michel, V., Thirion, B., Grisel,
  O., Blondel, M., Prettenhofer, P., Weiss, R., Dubourg, V., et~al. (2011).
\newblock Scikit-learn: Machine learning in {P}ython.
\newblock {\em Journal of machine learning research}, 12(Oct):2825--2830.

\bibitem[Poli et~al., 2008]{poli2008field}
Poli, R., Langdon, W.~B., McPhee, N.~F., and Koza, J.~R. (2008).
\newblock {\em A field guide to genetic programming}.
\newblock Lulu. com.

\bibitem[Ratle and Sebag, 2001]{ratle2001avoiding}
Ratle, A. and Sebag, M. (2001).
\newblock Avoiding the bloat with stochastic grammar-based genetic programming.
\newblock In {\em International Conference on Artificial Evolution (Evolution
  Artificielle)}, pages 255--266. Springer.

\bibitem[Sadowski et~al., 2013]{sadowski2013usefulness}
Sadowski, K.~L., Bosman, P. A.~N., and Thierens, D. (2013).
\newblock On the usefulness of linkage processing for solving {MAX}-{SAT}.
\newblock In {\em Genetic and Evolutionary Computation Conference (GECCO)
  2013}, pages 853--860. ACM.

\bibitem[Salustowicz and Schmidhuber, 1997]{salustowicz1997probabilistic}
Salustowicz, R. and Schmidhuber, J. (1997).
\newblock Probabilistic incremental program evolution.
\newblock {\em Evolutionary Computation}, 5(2):123--141.

\bibitem[Sastry and Goldberg, 2003]{sastry2003probabilistic}
Sastry, K. and Goldberg, D.~E. (2003).
\newblock Probabilistic model building and competent genetic programming.
\newblock In {\em Genetic Programming Theory and Practice}, pages 205--220.
  Springer.

\bibitem[Shan et~al., 2004]{shan2004grammar}
Shan, Y., McKay, R.~I., Baxter, R., Abbass, H., Essam, D., and Nguyen, H.
  (2004).
\newblock Grammar model-based program evolution.
\newblock In {\em IEEE Congress on Evolutionary Computation (CEC) 2004},
  volume~1, pages 478--485. IEEE.

\bibitem[Sotto and de~Melo, 2017]{sotto2017probabilistic}
Sotto, L. F. D.~P. and de~Melo, V.~V. (2017).
\newblock A probabilistic linear genetic programming with stochastic
  context-free grammar for solving symbolic regression problems.
\newblock In {\em Genetic and Evolutionary Computation Conference (GECCO)
  2017}, pages 1017--1024. ACM.

\bibitem[Tanev, 2007]{tanev2007genetic}
Tanev, I. (2007).
\newblock Genetic programming incorporating biased mutation for evolution and
  adaptation of snakebot.
\newblock {\em Genetic Programming and Evolvable Machines}, 8(1):39--59.

\bibitem[Thierens and Bosman, 2011]{thierens2011optimal}
Thierens, D. and Bosman, P. A.~N. (2011).
\newblock Optimal mixing evolutionary algorithms.
\newblock In {\em Genetic and Evolutionary Computation Conference (GECCO)
  2011}, pages 617--624. ACM.

\bibitem[Thierens and Bosman, 2013]{thierens2013hierarchical}
Thierens, D. and Bosman, P. A.~N. (2013).
\newblock Hierarchical problem solving with the linkage tree genetic algorithm.
\newblock In {\em Genetic and Evolutionary Computation Conference (GECCO)
  2013}, pages 877--884. ACM.

\bibitem[Virgolin et~al., 2018]{virgolin2018symbolic}
Virgolin, M., Alderliesten, T., Bel, A., Witteveen, C., and Bosman, P. A.~N.
  (2018).
\newblock Symbolic regression and feature construction with {GP}-{GOMEA}
  applied to radiotherapy dose reconstruction of childhood cancer survivors.
\newblock In {\em Genetic and Evolutionary Computation Conference (GECCO)
  2018}, pages 1395--1402. ACM.

\bibitem[Virgolin et~al., 2019]{virgolin2019linear}
Virgolin, M., Alderliesten, T., and Bosman, P. A.~N. (2019).
\newblock Linear scaling with and within semantic backpropagation-based genetic
  programming for symbolic regression.
\newblock In {\em Genetic and Evolutionary Computation Conference (GECCO)
  2019}, New York, NY, USA. ACM.

\bibitem[Virgolin et~al., 2017]{virgolin2017scalable}
Virgolin, M., Alderliesten, T., Witteveen, C., and Bosman, P. A.~N. (2017).
\newblock Scalable genetic programming by gene-pool optimal mixing and
  input-space entropy-based building-block learning.
\newblock In {\em Genetic and Evolutionary Computation Conference (GECCO)
  2017}, pages 1041--1048, New York, NY, USA. ACM.

\bibitem[Vladislavleva et~al., 2009]{vladislavleva2009order}
Vladislavleva, E.~J., Smits, G.~F., and Den~Hertog, D. (2009).
\newblock Order of nonlinearity as a complexity measure for models generated by
  symbolic regression via {P}areto genetic programming.
\newblock {\em IEEE Transactions on Evolutionary Computation}, 13(2):333--349.

\bibitem[Wong et~al., 2014]{wong2014grammar}
Wong, P.-K., Lo, L.-Y., Wong, M.-L., and Leung, K.-S. (2014).
\newblock Grammar-based genetic programming with bayesian network.
\newblock In {\em IEEE Congress on Evolutionary Computation (CEC) 2014}, pages
  739--746. IEEE.

\bibitem[Yanai and Iba, 2003]{yanai2003estimation}
Yanai, K. and Iba, H. (2003).
\newblock Estimation of distribution programming based on bayesian network.
\newblock In {\em IEEE Congress on Evolutionary Computation (CEC) 2003},
  volume~3, pages 1618--1625. IEEE.

\bibitem[{\v{Z}}egklitz and Po{\v{s}}{\'\i}k, 2017]{vzegklitz2017symbolic}
{\v{Z}}egklitz, J. and Po{\v{s}}{\'\i}k, P. (2017).
\newblock Symbolic regression algorithms with built-in linear regression.
\newblock {\em arXiv preprint arXiv:1701.03641}.

\bibitem[Zhong et~al., 2018]{zhong2018multifactorial}
Zhong, J., Feng, L., Cai, W., and Ong, Y.-S. (2018).
\newblock Multifactorial genetic programming for symbolic regression problems.
\newblock {\em IEEE Trans. Syst. Man Cybern. Syst.}, (99):1--14.

\end{thebibliography}

\end{document}